\documentclass[a4paper,fleqn]{cas-sc}
\usepackage{setspace}

\usepackage[numbers]{natbib}

\makeatletter
\renewcommand\printorcid[1]{}
\makeatother

\usepackage{placeins}
\usepackage{placeins}
\usepackage{placeins}
\usepackage{graphicx}%
\usepackage{multirow}%
\usepackage{amsmath,amssymb,amsfonts}%
\usepackage{amsthm}%
\usepackage{mathrsfs}%
\usepackage[title]{appendix}%
\usepackage{xcolor}%
\usepackage{textcomp}%
\usepackage{float}
\usepackage{subcaption}
\usepackage{manyfoot}%
\usepackage{booktabs}%
\usepackage{adjustbox}
\usepackage{algorithm}%
\usepackage{algorithmicx}%
\usepackage{algpseudocode}%
\usepackage{listings}%
\usepackage{booktabs}        
\usepackage{multirow}        
\usepackage{threeparttable}  
\usepackage{adjustbox}       
\usepackage{makecell}
\usepackage{graphicx}
\usepackage{array}
\usepackage{longtable}
\usepackage{hyperref}
\usepackage{xcolor}

\newcommand{\up}[1]{\textcolor{blue}{(#1$\uparrow$)}}

\newcommand{\down}[1]{\textcolor{red}{(#1$\downarrow$)}}
\renewcommand{\arraystretch}{1.0}

\def\tsc#1{\csdef{#1}{\textsc{\lowercase{#1}}\xspace}}
\tsc{WGM}
\tsc{QE}

\ExplSyntaxOn
\cs_set:Npn \__first_footerline: {}
\ExplSyntaxOff
\begin{document}
\let\WriteBookmarks\relax

\shorttitle{}    

\shortauthors{}  

\title [mode = title]{Spatio-Semantic Expert Routing Architecture with Mixture-of-Experts for Referring Image Segmentation}

\author[1]{Alaa Dalaq}





\affiliation[1]{organization={King Fahd University of Petroleum and Mineral},
            city={Dhahran},
            postcode={31261}, 
            country={Saudi Arabia}}

\author[1,2]{Muzammil Behzad}

\cormark[1]
\ead{muzammil.behzad@kfupm.edu.sa}


\affiliation[2]{organization={SDAIA-KFUPM Joint Research Center for Artificial Intelligence},
            city={Dhahran},
            postcode={31261}, 
            country={Saudi Arabia}}

\cortext[1]{Corresponding author}

\begin{abstract}
 Referring image segmentation aims to produce a pixel-level mask for the image region described by a natural-language expression. Although pretrained vision-language models have improved semantic grounding, many existing methods still rely on uniform refinement strategies that do not fully match the diverse reasoning requirements of referring expressions. Because of this mismatch, predictions often contain fragmented regions, inaccurate boundaries, or even the wrong object, especially when pretrained backbones are frozen for computational efficiency. To address these limitations, we propose SERA, a Spatio-Semantic Expert Routing Architecture for referring image segmentation. SERA introduces lightweight, expression-aware expert refinement at two complementary stages within a vision-language framework. First, we design SERA-Adapter, which inserts an expression-conditioned adapter into selected backbone blocks to improve spatial coherence and boundary precision through expert-guided refinement and cross-modal attention. We then introduce SERA-Fusion, which strengthens intermediate visual representations by reshaping token features into spatial grids and applying geometry-preserving expert transformations before multimodal interaction. In addition, a lightweight routing mechanism adaptively weights expert contributions while remaining compatible with pretrained representations. To make this routing stable under frozen encoders, SERA uses a parameter-efficient tuning strategy that updates only normalization and bias terms, affecting less than 1\% of the backbone parameters. Experiments on standard referring image segmentation benchmarks show that SERA consistently outperforms strong baselines, with especially clear gains on expressions that require accurate spatial localization and precise boundary delineation.
\end{abstract}



\maketitle

\section{Introduction}\label{intro}

Referring image segmentation (RIS) aims to generate a pixel-accurate mask for the image region specified by a natural-language expression~\cite{yu2016modeling, rethinking, wang2022cris}. Unlike category-based semantic segmentation, RIS requires the model to perform more than merely identifying object categories. It must connect language to visual content while handling spatial relations, fine-grained attributes, contextual cues, and object boundaries. The task becomes especially difficult in cluttered scenes or when the target must be separated from visually similar objects using subtle cues. Recent vision-language models (VLMs) have improved RIS by strengthening alignment between text and image representations~\cite{wang2022cris, yang2022lavt, Pu2024-lc}. Models such as CLIP~\cite{radford2021learning} provide strong semantic text embeddings, while self-supervised visual encoders such as DINO~\cite{caron2021emerging} and DINOv2~\cite{oquab2024dinov2learningrobustvisual} offer transferable spatial features that are useful for dense prediction. Yet current RIS methods still struggle to produce coherent masks with accurate boundaries, especially when the target is small, partially occluded, or visually similar to nearby regions~\cite{wang2022cris, yang2022lavt, zou2023seem}. These challenges become more pronounced when pretrained encoders are kept frozen to reduce training cost and preserve generalization, since this limits how much the visual representation can adapt to the referring task. 

One possible way to address this limitation is through expert specialization. Mixture-of-Experts (MoE) architectures allow different inputs to pass through different specialized transformations rather than forcing all samples through the same processing path~\cite{shazeer2017outrageously, fedus2021switch}. This property is particularly important for RIS, where referring expressions require diverse types of reasoning. Some depend mainly on spatial layout, others on visual appearance, while many rely on contextual or relational cues. However, incorporating expert routing into RIS is not straightforward. Routing can become unstable during training, and poorly designed expert modules may interfere with the pretrained representations that make modern vision-language backbones effective.

Motivated by these challenges, we introduce SERA (Spatio-Semantic Expert Routing Architecture), a vision-language framework for referring image segmentation that incorporates lightweight expert refinement at two complementary stages. The first component, SERA-Adapter, injects expression-conditioned expert corrections into selected transformer blocks of the visual backbone. This enables the model to refine intermediate features using cross-modal context, improving spatial consistency and boundary quality within the backbone. The second component, SERA-Fusion, reshapes visual tokens into spatial feature maps and applies geometry-preserving expert refinement during multimodal interaction. Together, these modules refine visual representations both within the backbone and immediately before fusion, allowing the model to capture complementary cues while avoiding the limitations of a single uniform refinement strategy.

\begin{figure}
    \centering
    \includegraphics[width=\linewidth]{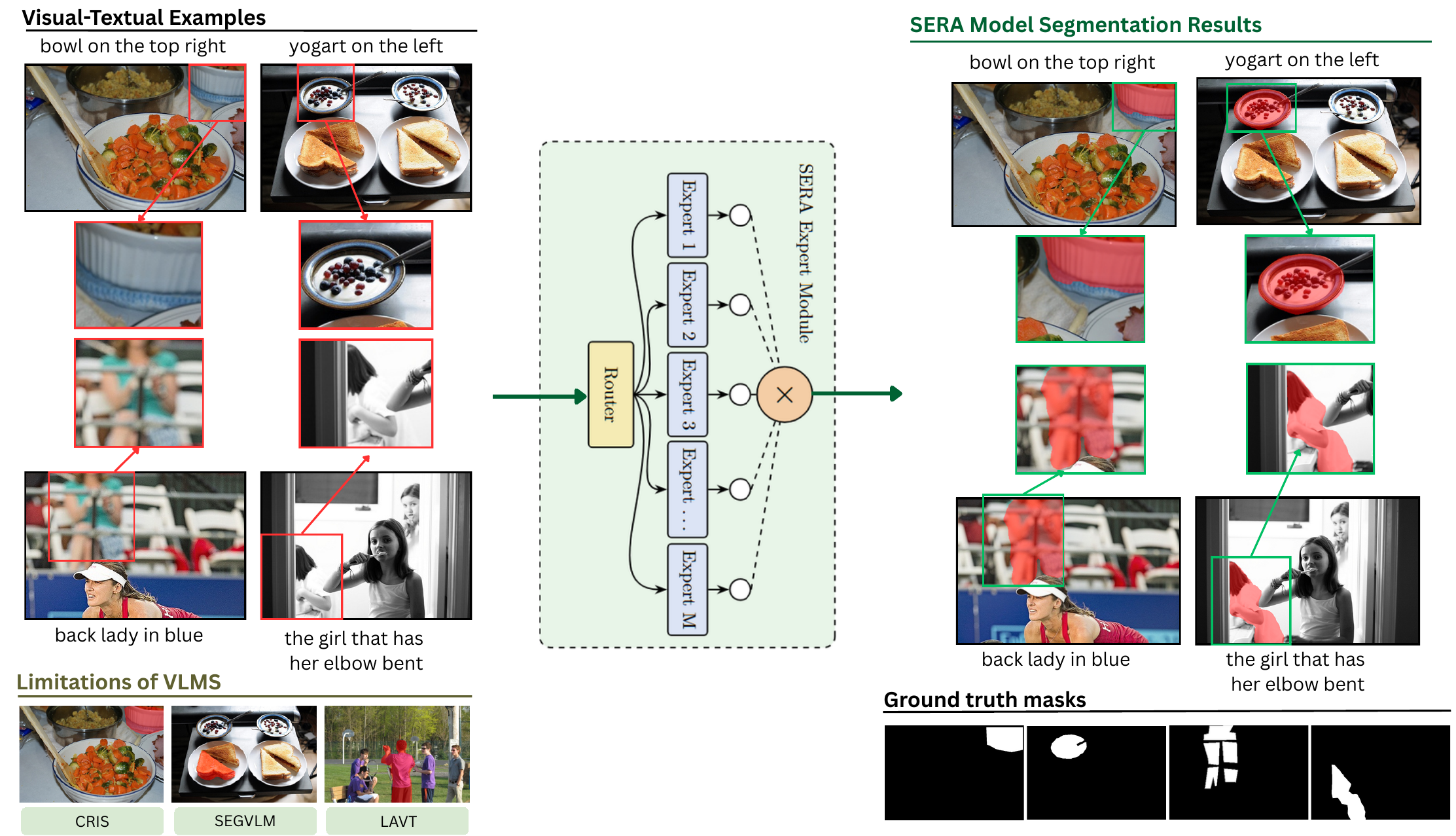}
\caption{  Overview of the proposed SERA. The left side of the figure shows visual-textual examples that illustrate typical challenges in referring image segmentation, along with grammatically challenging prompts that involve fine-grained attribute reasoning, relational grounding, and complex object boundaries. The middle part of the figure shows the SERA architecture, in which a router dynamically assigns routing weights to multiple experts within an expert-guided refinement module. The right side of the figure shows qualitative segmentation results produced by SERA, demonstrating improved alignment between referring expressions and predicted object regions. Ground-truth masks are shown at the bottom for reference.}

\label{fig:abstract}
\end{figure}

Figure~\ref{fig:abstract} highlights typical failure cases in existing RIS models, such as fragmented masks, boundary leakage, and incorrect object selection when expressions are ambiguous. SERA tackles these issues by selectively activating specialized experts conditioned on the input, allowing the refinement process to better match the reasoning requirements of each expression. To maintain stable routing while preserving the strengths of the pretrained backbone, we adopt a parameter-efficient tuning strategy that updates only LayerNorm and bias parameters, affecting less than $1\%$ of the backbone parameters.

\subsection{Salient Features of SERA}
Based on our main contributions, the salient features of our SERA model are summarized as follows:
\begin{itemize}
    \item We introduce SERA, a mixture-of-experts framework for referring image segmentation under pretrained vision-language backbones, enabling expression-conditioned specialization at both backbone and fusion stages.
    \item We propose SERA-Adapter, a backbone-level expert refinement module that injects expression-aware corrections into selected transformer blocks while preserving pretrained representations through a frozen-backbone adaptation regime.
    \item We design SERA-Fusion, a structured expert module that refines intermediate spatial feature maps via conditional Top-$K$ routing, capturing complementary cues including spatial layout, boundary structure, contextual interaction, and global shape consistency.
    \item We incorporate routing stabilization strategies to mitigate expert collapse by combining soft routing in SERA-Adapter for stable backbone adaptation with sparse Top-$K$ routing in SERA-Fusion to encourage expert specialization.
    \item We demonstrate strong zero-shot cross-dataset generalization across the RefCOCO dataset family, indicating that the learned vision-language representations transfer beyond dataset-specific patterns.
\end{itemize}

The remainder of this paper is organized as follows.
Section \ref{section2} reviews related work.
Section \ref{section3} presents the proposed SERA framework.
Section \ref{sec2} describes the experimental setup.
Section \ref{sec2b} reports the experimental results and discussions.
Section \ref{sec4} provides expert routing analysis.
Section \ref{sec5} evaluates zero-shot cross-dataset generalization.
Section \ref{sec6} concludes the paper and discusses limitations and future directions.

\section{Related Work}
\label{section2}
This section reviews the main research areas connected to our method. We begin with referring image segmentation, including early CNN-based approaches, Transformer-based models, and more recent methods built on pretrained vision-language representations. We then discuss mixture-of-experts (MoE) architectures, with an emphasis on conditional computation and expert specialization in vision and multimodal learning. Finally, we review parameter-efficient tuning strategies for adapting large pretrained models, particularly those relevant to dense vision-language tasks.

\subsection{Referring Image Segmentation}
Referring image segmentation (RIS) focuses on localizing and segmenting an object described by a natural-language expression, which requires fine-grained alignment between visual content and linguistic information~\cite{yu2016modeling, LI2026112549, Nguyen-Truong_2025_WACV}. Early RIS methods mainly followed a CNN-RNN design, where convolutional networks were used to extract visual features and recurrent models were used to encode the input expression. Representative examples include the Referring Relationship Network (RRN)~\cite{li2018referring} and Recurrent Multimodal Interaction (RMI)~\cite{liu2017recurrent}. These methods combined visual and language features through sequential interaction and then produced pixel-level predictions using fully convolutional decoders. Although they showed promising results, they were still limited in modeling long-range dependencies and more complex forms of cross-modal interaction.

Transformer-based architectures later changed the direction of RIS research by making multimodal reasoning more flexible through attention mechanisms. In these models, self-attention and cross-attention allow deeper interaction between visual and linguistic features, which improves the modeling of spatial relations and contextual information~\cite{yang2022lavt, dalaq2025deformable}. For example, MDETR~\cite{kamath2021mdetr} and VLT~\cite{ding2021vlt} reported strong results in language-guided grounding and referring image segmentation by combining cross-modal attention with query-based representations. Their success helped establish a strong foundation for multimodal alignment in dense prediction tasks.

More recently, RIS methods have started to benefit from large-scale pretrained vision-language models. Approaches such as CRIS~\cite{wang2022cris}, ETRIS~\cite{xu2023bridging}, and UniLSeg~\cite{liu2023unilseg} improve sentence-to-pixel alignment by transferring multimodal knowledge from pretrained encoders to dense prediction settings. Even with these advances, however, most methods still place the main emphasis on visual-linguistic interaction at the decoding stage, while treating intermediate visual features in a largely uniform way across different referring expressions. This leaves room for more specialized and expression-aware refinement inside pretrained backbones.

\subsection{Mixture-of-Experts}

Mixture-of-Experts (MoE) architectures enable conditional computation by dividing a network into multiple expert submodules and using a learned router to select or weight their outputs for each input. Although the idea of combining specialized experts is not new, recent sparsely gated MoE variants have made this design practical at scale, allowing models to increase capacity without a matching increase in computation~\cite{shazeer2017outrageously}. This line of work was further strengthened by Switch Transformers, which showed that sparse routing can be scaled efficiently to very large models~\cite{fedus2021switch}.

MoE has also been introduced into computer vision. For example, Vision Mixture-of-Experts (V-MoE) incorporates sparse expert layers into Vision Transformers and shows that expert specialization can improve scalability in large-scale image classification~\cite{10.5555/3540261.3540918}. Related ideas have also appeared in multimodal learning. Models such as Uni-MoE use sparse expert routing to process visual and linguistic information through different specialized pathways, making them better suited to heterogeneous reasoning patterns across modalities~\cite{li_unimoe}. 

Despite these advances, most prior MoE studies focus on scalability, efficiency, or high-level reasoning. They do not directly address the preservation of fine-grained spatial structure required for dense prediction. This distinction matters in referring image segmentation (RIS), where success depends on accurate boundary delineation, spatial relation modeling, contextual disambiguation, and coherent object structure. These requirements can vary substantially from one referring expression to another. To the best of our knowledge, sparse expert routing remains underexplored for expression-conditioned dense segmentation within pretrained vision-language frameworks. This limitation motivates our use of structured expert specialization in RIS.

\subsection{Adaptation of Pretrained Vision-Language Models}

Large pretrained vision-language models have improved performance across a wide range of multimodal tasks. Still, adapting them to dense prediction problems such as referring image segmentation (RIS) remains difficult in practice. Fully fine-tuning all backbone parameters is often computationally expensive and can weaken pretrained representations that are important for generalization. For this reason, efficient adaptation strategies have become an attractive alternative.

Parameter-Efficient Tuning (PET) methods adapt pretrained models by updating only a small subset of parameters while leaving most of the backbone frozen. This reduces training cost and memory usage while preserving the benefits of large-scale pretraining. As pretrained models continue to grow, such strategies have become increasingly important for task-specific adaptation.

Existing PET methods are usually grouped into three broad categories. The first adds small trainable components while keeping the original backbone fixed, as in adapter-based methods and prompt tuning~\cite{houlsby2019adapter, li2021prefix, zhou2022prompt}. The second updates only a limited subset of the original parameters, such as bias terms or normalization layers~\cite{guo2020revisiting, zaken2021bitfit}. The third uses low-rank parameterizations to model weight updates efficiently, as in LoRA and related variants~\cite{hu2021lora, karimimahabadi2021compacter}.

Although PET has been studied extensively in natural language processing and more recently in computer vision, many vision-oriented methods still focus on classification or image generation. Dense prediction is harder because pixel-level outputs impose stronger spatial and structural constraints. In RIS, some works have begun to explore efficient adaptation of pretrained vision-language models. For example, ETRIS~\cite{xu2023bridging} and BarleRIa~\cite{wang2023barleria} use adapter-based designs to adapt CLIP~\cite{radford2021learning} for segmentation. However, these methods mainly strengthen cross-modal interaction while still relying on largely uniform refinement across samples.

Our approach follows a frozen-backbone adaptation setting and updates only bias and LayerNorm parameters, while introducing conditional expert routing for structured, expression-aware specialization. 
\section{The Proposed SERA Framework}
\label{section3}
\begin{figure}
    \centering
    \includegraphics[width=\textwidth]{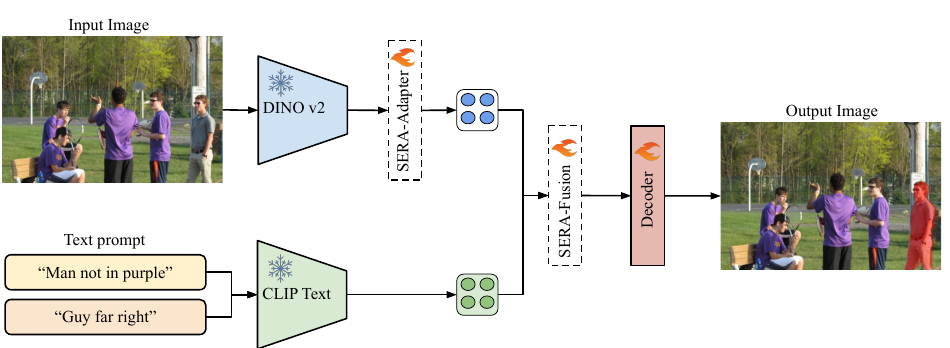}
\caption{ 
Architecture of SERA for referring image segmentation.
Visual features from a frozen DINOv2 backbone are refined using SERA-Adapter and fused with CLIP text embeddings via SERA-Fusion before decoding the final segmentation mask.
}

    \label{fig:overall}
\end{figure}
We illustrate the overall architecture of our proposed SERA framework in Figure~\ref{fig:overall}. As demonstrated, the model builds on a pretrained vision-language referring image segmentation baseline composed of a DINOv2 visual encoder and a CLIP text encoder. Given an input image I and a referring expression Q, the visual encoder produces a sequence of image tokens, while the text encoder extracts a global expression embedding. SERA introduces two targeted modifications to this baseline. First, a SERA-Adapter is inserted into selected transformer blocks of the visual backbone to refine intermediate visual tokens using expression-guided expert modulation. Second, we apply the SERA-Fusion module at the visual-language fusion stage, where spatial visual tokens are reshaped into 2D feature maps and refined using a mixture-of-experts mechanism before mask prediction. Both modules operate on intermediate representations and do not require full fine-tuning of the pretrained encoders.

\subsection{SERA-Adapter: Expert-Driven Backbone Refinement}
\label{sec:sera_adapter}

\begin{figure}
    \centering
    \includegraphics[width=\textwidth]{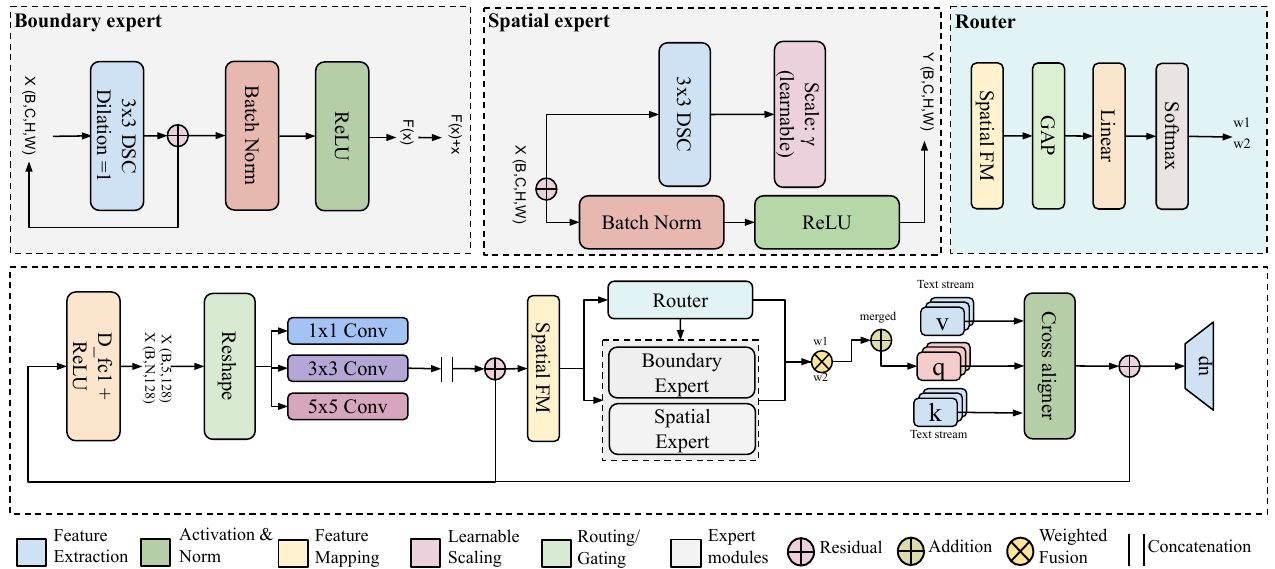}
\caption{ 
Detailed architecture of the proposed SERA-Adapter.
Visual tokens from the backbone are projected into a spatial grid and processed by multi-scale convolutional branches to enrich local context.
The resulting features are refined by two specialized experts, a spatial expert and a boundary expert, whose outputs are adaptively weighted by a lightweight router.
The refined visual features are then aligned with text representations through cross-modal attention and incorporated into the backbone via a residual update.
}

    \label{fig:sera-adapter}
\end{figure}

As illustrated in Figure~\ref{fig:sera-adapter}, the SERA-Adapter processes backbone visual tokens by first mapping them into a 2D spatial feature grid and applying multi-scale convolutional projections to enrich local spatial context.
The model then feeds the resulting feature grid into two specialized refinement experts, a boundary expert and a spatial expert, and adaptively combines their outputs through a routing module, producing a fused representation for subsequent interaction with the referring expression in the cross-modal alignment module.

\subsubsection{Spatio-Semantic Refinement Experts}
SERA-Adapter refines the spatial feature grid using two depthwise experts with complementary behavior: a boundary expert that injects edge responses, and a spatial expert that enhances local feature consistency. Both experts preserve spatial resolution and operate on the same feature map.

\paragraph{Boundary Expert.}
The boundary expert enhances contour-sensitive responses using a learnable depthwise $3\times3$ convolution.
The resulting responses are injected into the input through a scaled residual update as:
\begin{equation}
\mathbf{B} = \mathrm{ReLU}\!\left(
\mathrm{BN}\!\left(
\mathbf{G} + \beta \cdot \mathrm{DWConv}_{3\times3}(\mathbf{G})
\right)
\right),
\end{equation}
where $\beta=0.1$ is a fixed scaling factor.
This design preserves spatial resolution while stabilizing the refinement through normalization and activation.

\paragraph{Spatial Expert.}
The spatial expert applies a depthwise $3\times3$ convolution followed by batch normalization and a ReLU activation, and incorporates a scaled residual connection as:
\begin{equation}
\mathbf{S} = \phi\!\left(\mathrm{DWConv}_{3\times3}(\mathbf{G})\right) + \alpha \mathbf{G},
\end{equation}
where $\phi(\cdot)$ denotes the composition of batch normalization and ReLU, and $\alpha = 0.3$.

\subsubsection{Adaptive Soft Routing}

As shown in Figure~\ref{fig:sera-adapter}, the SERA-Adapter employs a soft routing module to combine the outputs of the spatial and boundary experts adaptively. Given the projected token sequence, we first isolate the spatial tokens and compute a global summary by averaging over the spatial dimension as:
\begin{equation}
\mathbf{z} = \frac{1}{N}\sum_{i=1}^{N}\mathbf{x}_{i},
\qquad 
\mathbf{z}\in\mathbb{R}^{B\times d},
\end{equation}
where $\{\mathbf{x}_{i}\}_{i=1}^{N}$ denote the spatial tokens after projection (excluding prefix tokens), and $d$ is the adapter channel width. The router applies a linear projection followed by softmax normalization to produce routing weights as:
\begin{equation}
[w_s, w_b] = \boldsymbol{\sigma}\!\left(\mathcal{R}(\mathbf{z})\right),
\end{equation}
where $\mathcal{R}(\cdot)$ is a learnable linear layer and $\boldsymbol{\sigma}(\cdot)$ denotes the softmax function. The expert outputs are then combined using the predicted weights and injected residually into the enriched spatial feature grid as:
\begin{equation}
\mathbf{G}_{\mathrm{corr}}
=
\mathbf{G}_{\mathrm{rich}}
+
\alpha\, w_s\, \mathbf{E}_{s}
+
\beta\, w_b\, \mathbf{E}_{b},
\end{equation}
where $\mathbf{E}_{s}$ and $\mathbf{E}_{b}$ denote the spatial and boundary expert outputs, respectively, and $\alpha=0.25$, $\beta=0.15$ are fixed scaling coefficients.

\subsubsection{Adapter Pipeline}
The SERA-Adapter refines intermediate visual tokens through spatial projection, multi-scale context enrichment, expert-based correction, and cross-modal alignment. Let $\mathbf{X}\in\mathbb{R}^{B\times P\times D_v}$ denote the incoming visual token sequence. We first project tokens to the adapter width $d$ using a linear layer as:
\begin{equation}
\mathbf{X}' = \mathrm{ReLU}(\mathbf{X}\mathbf{W}_1),
\qquad 
\mathbf{X}'\in\mathbb{R}^{B\times P\times d}.
\end{equation}
Following the baseline token layout, we split $\mathbf{X}'$ into prefix tokens and spatial tokens, reshape the spatial tokens into a 2D grid, and obtain $\mathbf{G}\in\mathbb{R}^{B\times d\times H'\times W'}$. To enrich local spatial context, $\mathbf{G}$ is processed by parallel convolutional branches with different kernel sizes (e.g., $1\times1$, $3\times3$, $5\times5$) and fused in a residual manner as:
\begin{equation}
\mathbf{G}_{\mathrm{rich}} = \mathbf{G} + \mathcal{M}(\mathbf{G}),
\end{equation}
where $\mathcal{M}(\cdot)$ aggregates the multi-scale branch outputs.

Next, $\mathbf{G}_{\mathrm{rich}}$ is refined by two lightweight experts: a spatial expert and a boundary expert, both implemented using depthwise $3\times3$ convolutions with residual sharpening and normalization. A soft router computes input-dependent weights that combine the two expert outputs to form the corrected grid as:
\begin{equation}
\mathbf{G}_{\mathrm{corr}}
=
\mathbf{G}_{\mathrm{rich}}
+
\alpha\, w_s\, \mathbf{E}_{s}
+
\beta\, w_b\, \mathbf{E}_{b}.
\end{equation}
Finally, we flatten $\mathbf{G}_{\mathrm{corr}}$ back into a spatial token sequence and apply cross-modal attention with the text embedding $\mathbf{T}$ as:
\begin{equation}
\mathbf{X}^{\star}_{\mathrm{spa}} = \mathcal{C}(\mathrm{Flat}(\mathbf{G}_{\mathrm{corr}}), \mathbf{T}),
\end{equation}
We then concatenate the prefix tokens and apply a residual update as:
\begin{equation}
\hat{\mathbf{X}} = \mathrm{Concat}(\mathbf{X}'_{\mathrm{pre}}, \mathbf{X}^{\star}_{\mathrm{spa}}) + \mathbf{X}'.
\end{equation}
The adapter output is projected back to the backbone dimension via a second linear layer, producing $\mathbf{Y}\in\mathbb{R}^{B\times P\times D_v}$ for subsequent backbone blocks.

\subsubsection{Integration with Selected Vision Encoder Blocks}
\label{sec:sera_adapter_insertion}
The SERA-Adapter operates within selected transformer blocks of the DINOv2 visual backbone by augmenting the feed-forward residual pathway, while leaving the self-attention computation unchanged. Given an input token sequence $\mathbf{X}$, the block update can be formulated as:

\begin{equation}
\mathbf{X} \leftarrow \mathbf{X} + \mathcal{F}(\mathbf{X}) + \lambda\,\mathcal{A}(\mathbf{X}, \mathbf{T}),
\end{equation}
where $\mathcal{F}(\cdot)$ denotes the original feed-forward transformation, $\mathcal{A}(\cdot)$ represents the SERA-Adapter pipeline described above, $\mathbf{T}$ is the corresponding text embedding, and $\lambda$ controls the contribution of the adapter.
This formulation preserves the original DINOv2 block structure and dimensionality while introducing expression-conditioned refinement in a residual and parameter-efficient manner.

\subsection{SERA-Fusion: Expert-Guided Feature Aggregation}
\label{sec:sera_fusion}

\begin{figure}
    \centering
    \includegraphics[width=\textwidth]{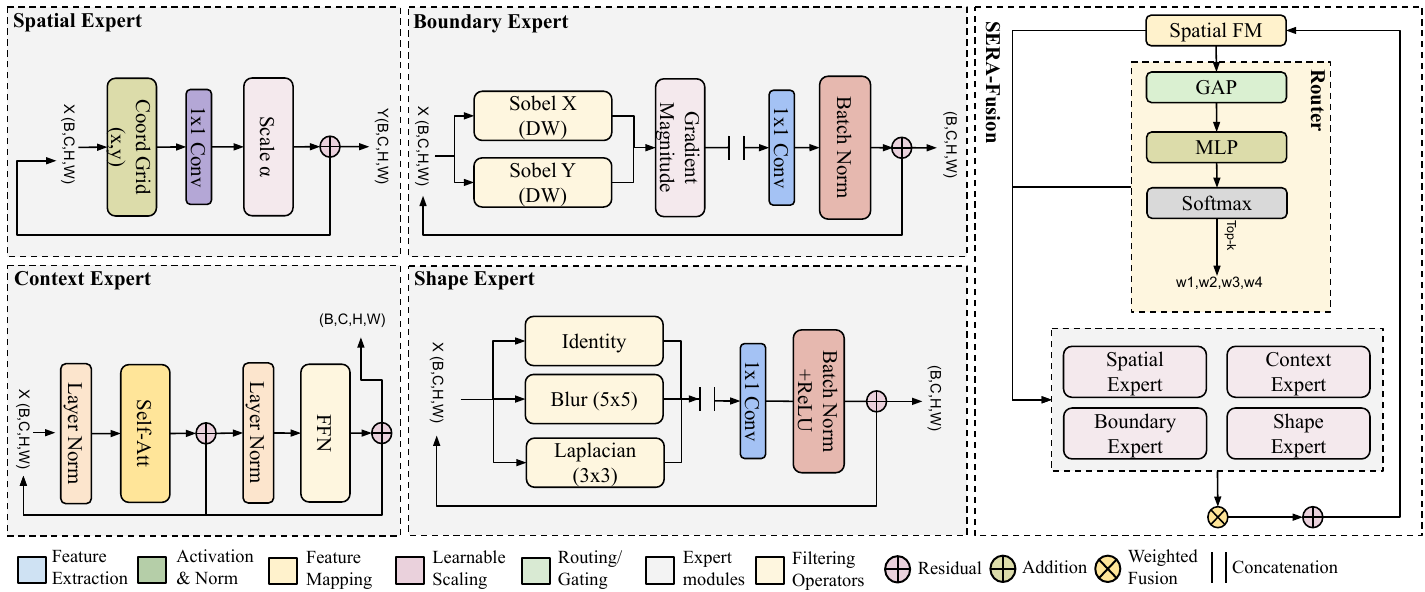}
\caption{ 
Architecture of the proposed SERA-Fusion module.
Intermediate visual feature maps are refined using a set of specialized experts targeting complementary cues, including spatial layout, boundary structure, contextual interaction, and global shape consistency.
A lightweight routing network computes input-dependent expert weights from pooled spatial features and aggregates the selected expert outputs through weighted fusion.
The refined representation preserves spatial resolution and feeds into the segmentation decoder.
}
    \label{fig:sera-fusion}
\end{figure}

As illustrated in Figure~\ref{fig:sera-fusion}, SERA-Fusion performs expert-guided refinement on intermediate spatial feature maps at the visual-language fusion stage to enhance representations prior to mask prediction. Given spatial feature maps extracted from selected backbone stages, the module applies specialized experts that capture complementary cues, including spatial structure, boundary information, contextual dependencies, and shape priors. At each refinement stage, a routing mechanism computes input-dependent weights to adaptively combine expert outputs, producing enhanced spatial features for subsequent cross-modal alignment with the referring expression. This design enables selective emphasis on distinct visual cues while preserving architectural stability and parameter efficiency.

\subsubsection{Experts Design}
SERA-Fusion employs a set of specialized experts that operate on spatial visual feature maps and capture complementary visual cues relevant to referring image segmentation.
Let $\mathbf{X}\in\mathbb{R}^{B\times C\times H\times W}$ denote an intermediate spatial feature representation.
All experts preserve spatial resolution and channel dimensionality while operating in a residual refinement manner.

\paragraph{Spatial Expert.}
The spatial expert injects explicit positional information to facilitate reasoning over relative spatial relationships.
Given an intermediate feature map $\mathbf{X} \in \mathbb{R}^{B \times C \times H \times W}$, a normalized coordinate grid $\mathbf{G} \in \mathbb{R}^{B \times 2 \times H \times W}$ is first constructed.
The grid is projected to the feature dimension using a $1\times1$ convolution and added residually to the input as:
\begin{equation}
E_{\mathrm{spa}}(\mathbf{X})
=
\mathbf{X}
+
\alpha \cdot
\mathrm{Conv}_{1\times1}(\mathbf{G}),
\end{equation}
where $\alpha$ is a fixed scaling factor controlling the strength of the positional injection.
This operation preserves both spatial resolution and channel dimensionality.

\paragraph{Context Expert.}
The context expert captures long-range spatial dependencies through self-attention-based contextual aggregation.
Given an intermediate feature map $\mathbf{X} \in \mathbb{R}^{B \times C \times H \times W}$,
the spatial dimensions are first flattened into a sequence representation
$\mathbf{X}_s \in \mathbb{R}^{B \times HW \times C}$.
Multi-head self-attention is applied, followed by a position-wise feed-forward network with residual connections as:
\begin{align}
\mathbf{Z}_1 &= \mathbf{X}_s + \mathcal{T}(\mathcal{N}(\mathbf{X}_s)), \\
\mathbf{Z}_2 &= \mathbf{Z}_1 + \mathcal{G}(\mathcal{N}(\mathbf{Z}_1)),
\end{align}
where $\mathcal{T}(\cdot)$ denotes the multi-head self-attention operator,
$\mathcal{G}(\cdot)$ denotes the feed-forward transformation,
and $\mathcal{N}(\cdot)$ denotes layer normalization.
The refined representation is reshaped back into its spatial form:
\begin{equation}
E_{\mathrm{ctx}}(\mathbf{X})
=
\mathrm{Reshape}(\mathbf{Z}_2).
\end{equation}
This operation preserves channel dimensionality and spatial resolution.

\paragraph{Boundary Expert.}
The boundary expert enhances contour-sensitive representations using fixed gradient-based filtering.
Given an intermediate feature map $\mathbf{X} \in \mathbb{R}^{B \times C \times H \times W}$,
depthwise Sobel operators are applied to extract horizontal and vertical gradients:
\begin{equation}
\mathbf{G}_x = \nabla_x \mathbf{X}, \quad
\mathbf{G}_y = \nabla_y \mathbf{X},
\end{equation}
where $\nabla_x$ and $\nabla_y$ are depthwise convolutional filters initialized as Sobel operators.
The gradient magnitude is computed as:
\begin{equation}
\mathbf{G}_{\mathrm{mag}} =
\sqrt{\mathbf{G}_x^2 + \mathbf{G}_y^2 + \epsilon},
\end{equation}
where $\epsilon$ is a small constant for numerical stability.
The original features and gradient responses are concatenated along the channel dimension and fused using a $1\times1$ convolution:
\begin{equation}
E_{\mathrm{bnd}}(\mathbf{X})
=
\mathbf{X}
+
\phi
\left(
\mathrm{Conv}_{1\times1}
\big(
[\mathbf{X}, \mathbf{G}_{\mathrm{mag}}, \mathbf{G}_x + \mathbf{G}_y]
\big)
\right),
\end{equation}
where $[\cdot]$ denotes channel-wise concatenation and
$\phi(\cdot)$ represents batch normalization followed by ReLU activation.
This expert preserves both spatial resolution and channel dimensionality.

\paragraph{Shape Expert.}
The shape expert promotes global structural consistency by combining low-frequency smoothing and high-frequency structural cues.
Given an intermediate feature map $\mathbf{X} \in \mathbb{R}^{B \times C \times H \times W}$,
a depthwise blur operator and a depthwise Laplacian filter are applied:
\begin{equation}
\mathbf{X}_{\mathrm{blur}} = \mathrm{Blur}(\mathbf{X}), \quad
\mathbf{X}_{\mathrm{lap}} = \mathrm{Laplace}(\mathbf{X}),
\end{equation}
where the operators are implemented as depthwise convolutions, with the Laplacian kernel initialized to a standard discrete Laplace operator.
The original features and structural responses are concatenated along the channel dimension and fused using a $1\times1$ convolution:
\begin{equation}
E_{\mathrm{shp}}(\mathbf{X})
=
\mathbf{X}
+
\phi
\left(
\mathrm{Conv}_{1\times1}
\big(
[\mathbf{X}, \mathbf{X}_{\mathrm{blur}}, \mathbf{X}_{\mathrm{lap}}]
\big)
\right),
\end{equation}
where $[\cdot]$ denotes channel-wise concatenation and
$\phi(\cdot)$ represents batch normalization followed by ReLU activation.
This operation preserves spatial resolution and channel dimensionality while enhancing structural coherence.

\subsubsection{Conditional Routing}
We employ a Top-$K$ sparse gating mechanism to modulate expert contributions conditioned on the input feature map.
Given an intermediate representation $\mathbf{X} \in \mathbb{R}^{B \times C \times H \times W}$, a global descriptor is obtained via spatial average pooling:
\begin{equation}
\mathbf{z} = \mathrm{GAP}(\mathbf{X}) \in \mathbb{R}^{B \times C}.
\end{equation}
Afterwards, the router logits are predicted using a two-layer projection:
\begin{equation}
\mathbf{r} = \mathbf{W}_2 \sigma(\mathbf{W}_1 \mathbf{z}),
\end{equation}
where $\sigma(\cdot)$ denotes ReLU activation and $\mathbf{r}\in\mathbb{R}^{B\times E}$.
During training, a Gaussian perturbation is added to encourage routing diversity:
\begin{equation}
\tilde{\mathbf{r}} = \frac{\mathbf{r} + \epsilon}{\tau},
\quad
\epsilon \sim \mathcal{N}(0, \sigma_{\text{noise}}^2),
\end{equation}
where $\tau$ denotes temperature. Noise is applied only during training.
Top-$k$ selection retains the $k$ largest logits per sample:
\begin{equation}
\mathcal{I}^{(b)} = \mathrm{TopK}(\tilde{\mathbf{r}}^{(b)}, k).
\end{equation}
Similarly, the sparse routing weights are computed by applying softmax over the selected logits:
\begin{equation}
w_i^{(b)} =
\begin{cases}
\frac{\exp(\tilde{r}_i^{(b)})}
{\sum_{j \in \mathcal{I}^{(b)}} \exp(\tilde{r}_j^{(b)})},
& i \in \mathcal{I}^{(b)}, \\
0, & \text{otherwise}.
\end{cases}
\end{equation}
Finally, the aggregated output is expressed as:
\begin{equation}
\mathbf{Y}^{(b)} = \sum_{i=1}^{E} w_i^{(b)} E_i(\mathbf{X}^{(b)}).
\end{equation}
To regularize router confidence, we include a Z-loss term penalizing the mean squared magnitude of router logits:
\begin{equation}
\mathcal{L}_z = \lambda_z \frac{1}{BE} \|\mathbf{r}\|_2^2.
\end{equation}

\subsection{Mitigating Expert Collapse and Routing Stability}

Mixture-of-Experts models may suffer from expert collapse, where the routing mechanism disproportionately routes samples to only one or a few experts, even when multiple experts are selected (e.g., Top-$k$ routing with $k=4$). We deliberately adopt different routing strategies across architectural stages. SERA-Adapter, inserted inside selected DINOv2 transformer blocks, uses soft routing to ensure stable residual refinement within the pretrained backbone. In contrast, SERA-Fusion operates on intermediate spatial feature maps prior to mask prediction, where sparse Top-$k$ routing effectively encourages specialization without disrupting backbone representations.

To prevent expert collapse under sparse routing, we introduce auxiliary regularization applied only during training. Let $\mathbf{r}\in\mathbb{R}^{B\times E}$ denote the routing logits for a mini-batch of size $B$ and $E$ experts.
First, we apply a squared-logit penalty to prevent over-confident routing as:
\begin{equation}
\mathcal{L}_{\text{logit}}
=
\lambda_{\text{logit}} \, \mathbb{E}\!\left[\|\mathbf{r}\|_2^2\right].
\end{equation}
Second, we encourage balanced expert utilization by penalizing the coefficient of variation (CV) of expert routing mass across the batch.
Let $\mathbf{u}\in\mathbb{R}^{E}$ denote the normalized routing mass per expert as:
\begin{equation}
u_i = \frac{\sum_{b=1}^{B} w_{b,i}}{\sum_{j=1}^{E}\sum_{b=1}^{B} w_{b,j}},
\end{equation}
where $w_{b,i}$ denotes the routing weight assigned to expert $i$ for sample $b$. To prevent the router from overusing a small subset of experts, we introduce a load-balancing regularization term. Specifically, we penalize uneven expert utilization using the coefficient of variation over the expert usage vector $\mathbf{u}$ as:

\begin{equation}
\mathcal{L}_{\text{balance}}
=
\lambda_{\text{bal}} \, \mathrm{CV}(\mathbf{u})^2,
\end{equation}
where $\mathrm{CV}(\cdot)$ denotes the coefficient of variation. 
In addition, we apply a similar regularization on the fraction of samples assigned to each expert in order to stabilize token allocation during training. Combining these terms, the overall mixture-of-experts regularization is defined as:

\begin{equation}
\mathcal{L}_{\text{MoE}}
=
\mathcal{L}_{\text{logit}}
+
\mathcal{L}_{\text{balance}}
+
\mathcal{L}_{\text{token}}.
\end{equation}
These auxiliary losses are used only during training and introduce no additional overhead at inference time.

\section{Experimental Setup}\label{sec2}
This section discusses the datasets, evaluation protocols and the implementation details.

\subsection{Datasets and Evaluation Protocol}
We evaluate SERA on three standard referring image segmentation benchmarks: RefCOCO \cite{kazemzadeh-etal-2014-referitgame}, RefCOCO+ \cite{kazemzadeh-etal-2014-referitgame}, and RefCOCOg (G-Ref) \cite{mao2016generation}.
All datasets are built upon MS-COCO images and provide natural-language referring expressions paired with pixel-level object masks. RefCOCO is the primary RIS benchmark and contains 19,994 images with 142,210 referring expressions describing 50,000 objects. RefCOCO+ includes 19,992 images with 141,564 expressions for 49,856 objects and is more challenging due to the removal of absolute spatial terms (e.g., left, right).
RefCOCOg (G-Ref) consists of 26,711 images with 104,560 expressions for 54,822 objects and features longer, more descriptive referring expressions.
Following prior work, we report results on both the Google (g) and UMD (u) splits of RefCOCOg.

\begin{table*}[!t]
\centering
\small
\setlength{\tabcolsep}{4pt}
\renewcommand{\arraystretch}{1.15}
\caption{  State-of-the-art comparison of RIS methods and parameter-efficient tuning (PET) on RefCOCO/RefCOCO+/G-Ref. Results are reported in mIoU. Best results are in bold.}
\label{tab:sota_refcoco}
\begin{adjustbox}{width=\textwidth}
\begin{tabular}{l|ccc|ccc|ccc|c}
\toprule
\multirow{2}{*}{\textbf{Method}} 
& \multicolumn{3}{c|}{\textbf{RefCOCO}} 
& \multicolumn{3}{c|}{\textbf{RefCOCO+}} 
& \multicolumn{3}{c|}{\textbf{G-Ref}} 
& \multirow{2}{*}{\textbf{Avg}} \\
\cmidrule(lr){2-4}\cmidrule(lr){5-7}\cmidrule(lr){8-10}
& \textbf{val} & \textbf{testA} & \textbf{testB}
& \textbf{val} & \textbf{testA} & \textbf{testB}
& \textbf{val(u)} & \textbf{test(u)} & \textbf{val(g)}
& \\
\midrule

\multicolumn{11}{l}{\textbf{Traditional Full Fine-tuning}} \\
\midrule

ReSTR~\cite{kim2022restr} 
& 67.2 & 69.3 & 64.5 
& 55.8 & 60.4 & 48.3
& 54.5 & - & 54.5 
& 58.8 \\

CRIS~\cite{wang2022cris} 
& 70.5 & 73.2 & 66.1 
& 62.3 & 68.1 & 53.7
& 59.9 & 60.4 & - 
& 63.8 \\

LAVT~\cite{yang2022lavt} 
& 72.7 & 75.8 & 68.8 
& 62.1 & 68.4 & 55.1
& - & - & 60.5 
& 64.9 \\

SEEM~\cite{zou2023seem}
& - & - & - 
& - & - & - 
& 65.6 & - & - 
& - \\

VPD~\cite{10377753} 
& 73.5 & - & - 
& 63.9 & - & -
& 63.1 & - & - 
& 66.8 \\

DMMI~\cite{hu2023dmm} 
& 74.1 & 77.1 & 70.2 
& 64.0 & 69.7 & 57.0
& 63.5 & 64.2 & 62.0 
& 66.9 \\

ReLA~\cite{liu2023rela} 
& 73.8 & 76.5 & 70.2 
& 66.0 & 71.0 & 57.7
& 65.0 & 66.0 & 62.7 
& 67.7 \\

ReLA~\cite{liu2023rela} 
& 73.8 & 76.5 & 70.18 
& 66.0 & 71.0 & 57.7
& 65.0 & 66.0 & 62.7 
& 67.5 \\

CGFormer~\cite{tang2023cgformer} 
& 74.8 & 77.3 & 70.6 
& 64.5 & 71.0 & 57.1
& 64.7 & 65.1 & 62.5 
& 67.7 \\

Lisa-7B~\cite{lai2024lisa}
& 74.1 & 76.5 & 71.1 
& 62.4 & 67.4 & 56.5 
& 66.4 & 68.5 & - 
& 67.9 \\

MagNet~\cite{cheng2024magnet} 
& 75.2 & 78.2 & 71.1 
& 66.2 & 71.3 & 58.1
& 65.4 & 66.2 & 63.1 
& 68.3 \\

ReMamber~\cite{yang2024remamber}
& 74.5 & 76.7 & 70.9 
& 65.0 & 70.8 & 57.5 
& 63.9 & 64.0 & - 
& 67.9 \\

RISCLIP-B~\cite{kim2024risclip}
& 75.7 & 78.0 & 72.5 
& 69.2 & 73.5 & 60.7 
& 67.6 & 68.0 & - 
& 70.6 \\

VATEX~\cite{Nguyen-Truong_2025_WACV}
& 78.2 & 79.6 & 75.6 
& 70.0 & 74.4 & 62.5 
& -& - & 69.7  
& 72.8 \\

MOAT~\cite{ZHANG2024110719}
& 73.6 & 76.4 & 68.8 
& 63.0 & 68.8 & 56.2
& 62.5 & 63.2 & - 
& 66.6 \\

LGFormer~\cite{WEI2026112306}
& 74.7 & 77.8 & 70.7 
& 65.7 & 71.5 & 57.9
& 63.7 & 65.2 & - 
& 68.4 \\

\midrule
\multicolumn{11}{l}{\textbf{Parameter Efficient Tuning}} \\
\midrule

ETRIS~\cite{xu2023bridging} 
& 70.5 & 73.5 & 66.6 
& 60.1 & 66.9 & 50.2
& 59.8 & 59.9 & 57.9 
& 62.8 \\

DETRIS-B\cite{10.1609/aaai.v39i4.32380}
& 76.0 & 78.2 & 73.5
& 68.9 & 74.0 & 61.5
& 67.9 & 68.1 & 65.9
& 70.4 \\

\textbf{SERA (Ours)*}
& \textbf{76.5} & \textbf{78.2} & \textbf{73.7}
& \textbf{70.4} & \textbf{74.4} & \textbf{62.8}
& \textbf{68.8} & \textbf{68.9} & \textbf{66.6}
& \textbf{71.1} \\

\midrule

\end{tabular}
\end{adjustbox}

\vspace{2pt}
{ $^{*}$ Indicates that the backbone is kept frozen while only bias and LayerNorm parameters are updated during training. This configuration differs from conventional PET approaches.}

\end{table*}
\subsection{Implementation Details}
We implement SERA on top of DETRIS \cite{10.1609/aaai.v39i4.32380} for referring image segmentation. We use a distilled DINOv2 \cite{oquab2024dinov2learningrobustvisual} transformer with register tokens as the visual backbone and a CLIP \cite{radford2021learning} text transformer to encode the referring expression. Unless stated otherwise, both pretrained encoders are frozen, and we train only the proposed modules and the task-specific segmentation layers. The model is trained using the Adam optimizer with an initial learning rate of $1\times10^{-4}$, which is reduced by a factor of 0.1 at later stages. All experiments are conducted on a single NVIDIA A6000 GPU with a batch size of 16. We compare the performance of SERA mainly using mean Intersection over Union (mIoU) and overall Intersection over Union (oIoU). between the predicted masks and the ground truth.

\newcommand{\promptcell}[1]{%
\parbox[c][1.85cm][c]{1.9cm}{\centering\footnotesize #1}
}

\begin{figure}
\centering
\setlength{\tabcolsep}{2pt}
\resizebox{\textwidth}{!}{%
\begin{tabular}{c c c c c}
 & Original Image & Ground Truth & DETRIS & SERA (ours) \\

\shortstack{ the \\banana in the\\ middle\\ with 2\\ black dots\\ on the right} &
\includegraphics[width=0.23\linewidth]{results/1460-img.jpg} &
\includegraphics[width=0.23\linewidth]{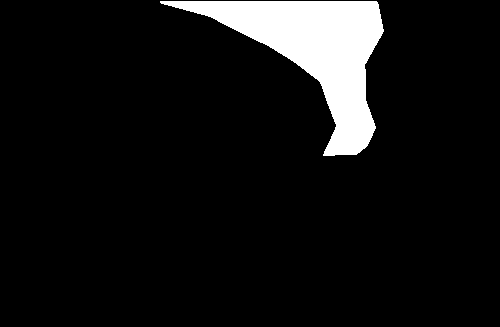} &
\includegraphics[width=0.23\linewidth]{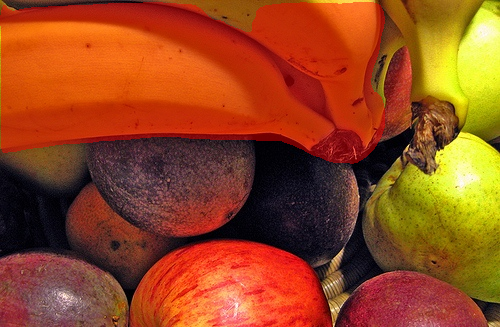} &
\includegraphics[width=0.23\linewidth]{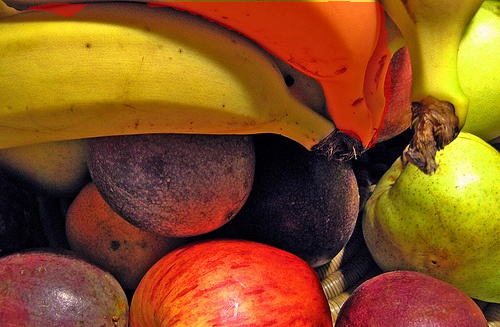} \\

\shortstack{ green\\and purple\\book} &
\includegraphics[width=0.23\linewidth]{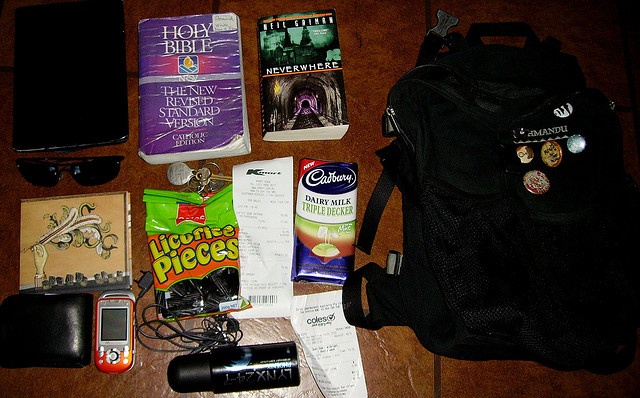} &
\includegraphics[width=0.23\linewidth]{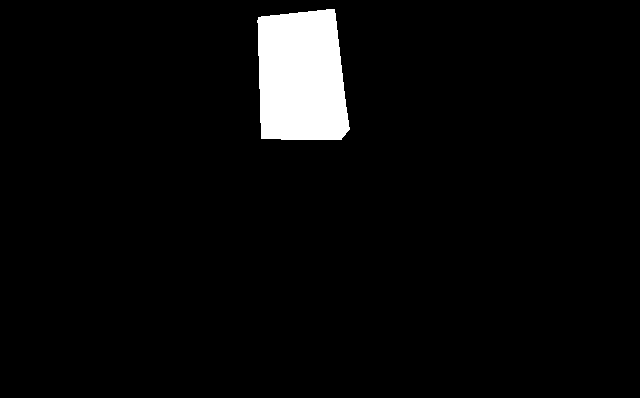} &
\includegraphics[width=0.23\linewidth]{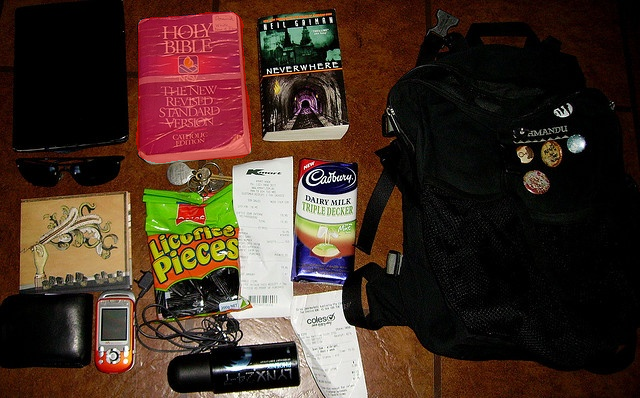} &
\includegraphics[width=0.23\linewidth]{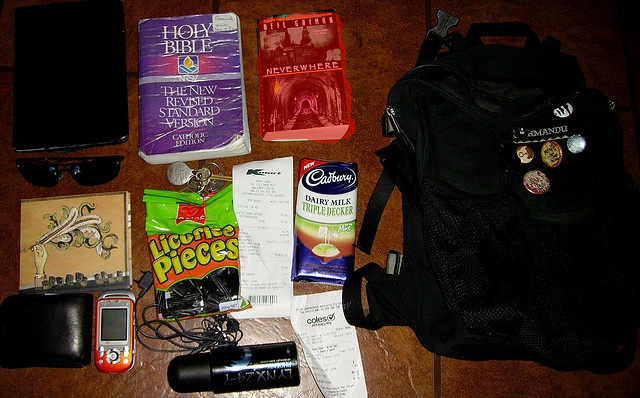} \\

\shortstack{ woman\\on the\\left} &
\includegraphics[width=0.23\linewidth]{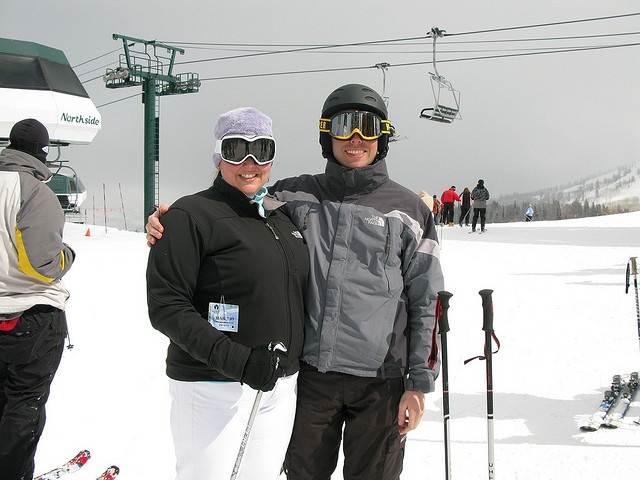} &
\includegraphics[width=0.23\linewidth]{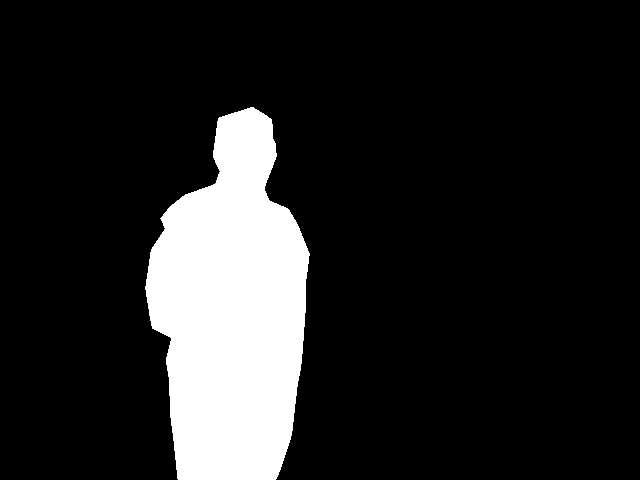} &
\includegraphics[width=0.23\linewidth]{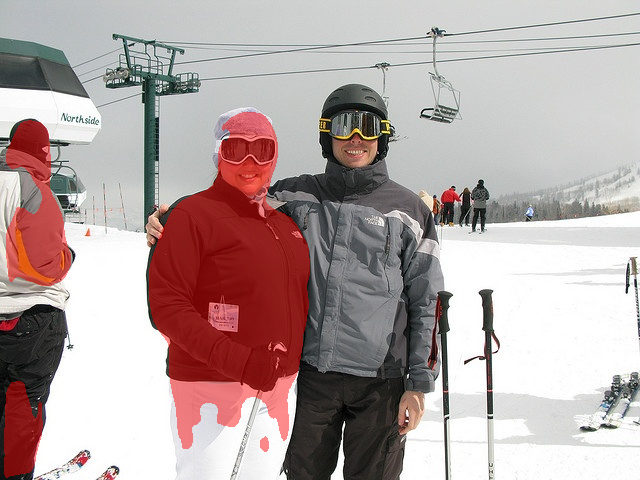} &
\includegraphics[width=0.23\linewidth]{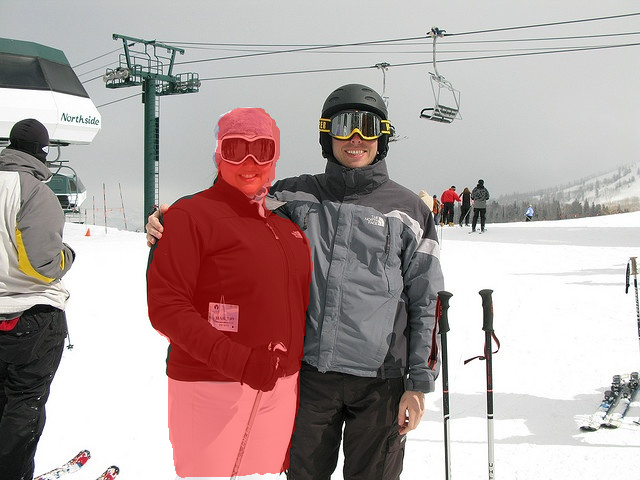} \\

\shortstack{ right\\white} &
\includegraphics[width=0.23\linewidth]{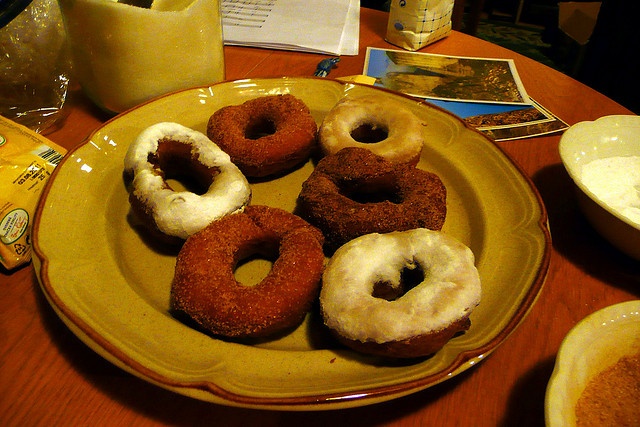} &
\includegraphics[width=0.23\linewidth]{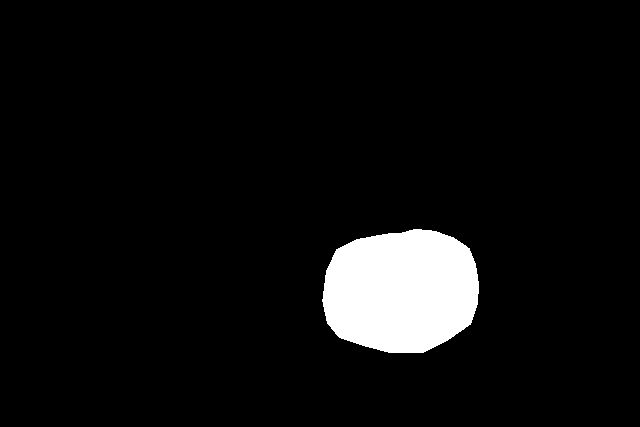} &
\includegraphics[width=0.23\linewidth]{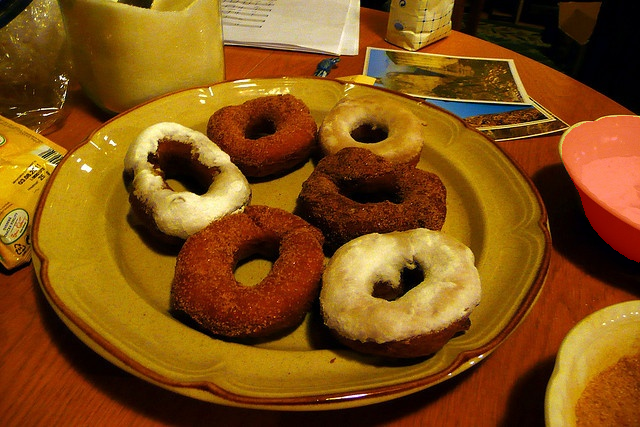} &
\includegraphics[width=0.23\linewidth]{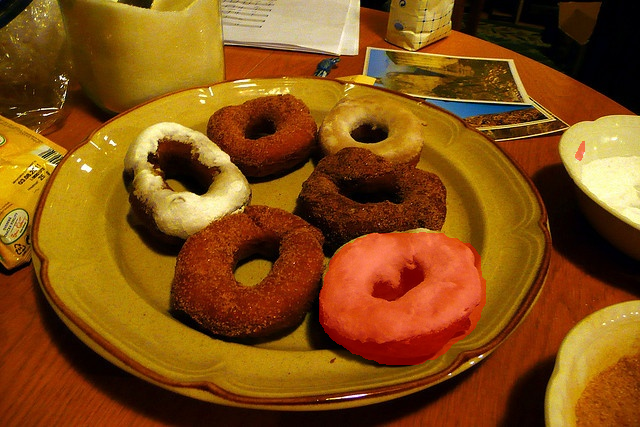} \\

\shortstack{ the\\bear\\hiding behind\\pole} &
\includegraphics[width=0.23\linewidth]{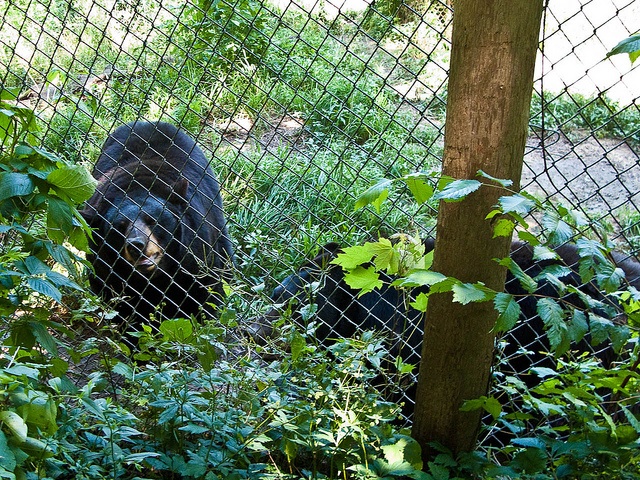} &
\includegraphics[width=0.23\linewidth]{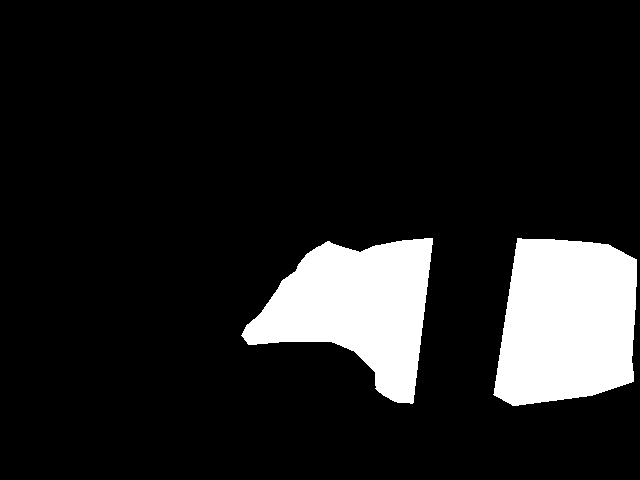} &
\includegraphics[width=0.23\linewidth]{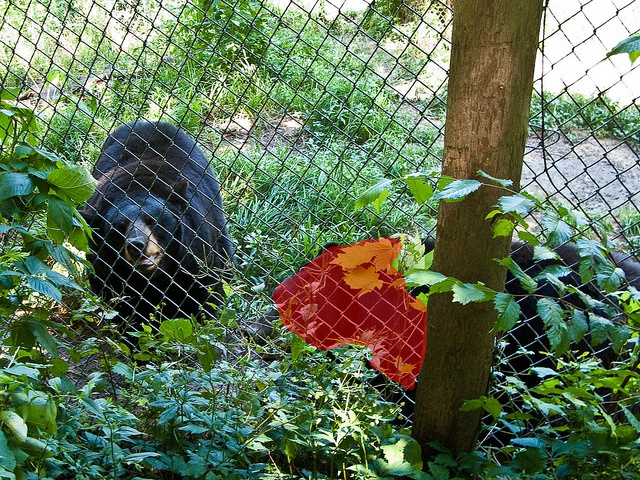} &
\includegraphics[width=0.23\linewidth]{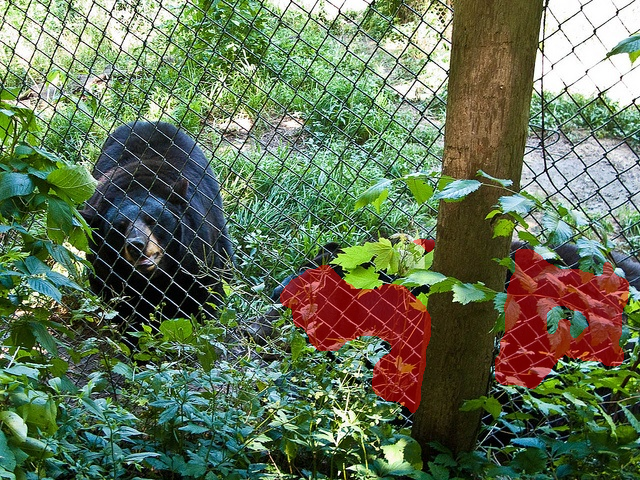} \\

\shortstack{ boy in\\purple wearing\\dark glasses} &
\includegraphics[width=0.23\linewidth]{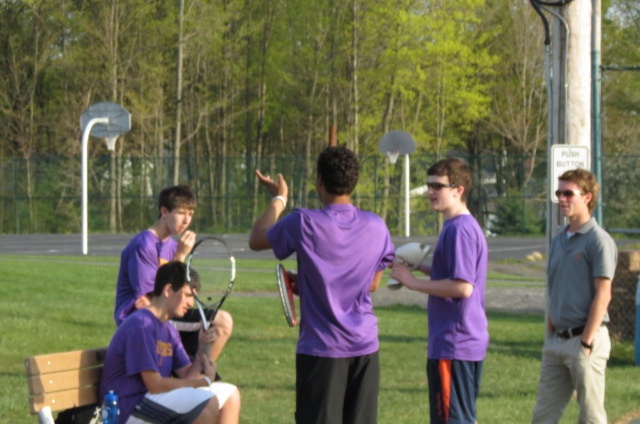} &
\includegraphics[width=0.23\linewidth]{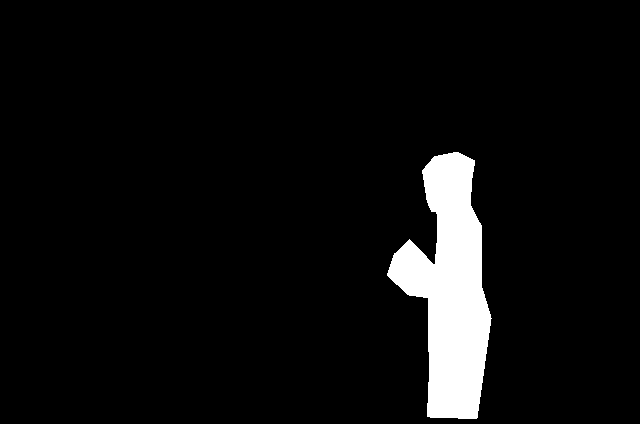} &
\includegraphics[width=0.23\linewidth]{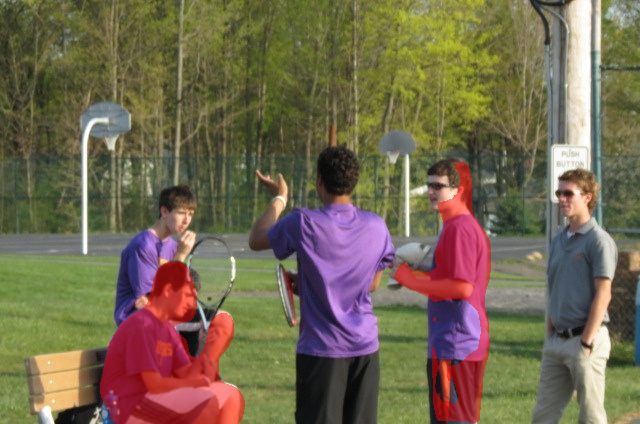} &
\includegraphics[width=0.23\linewidth]{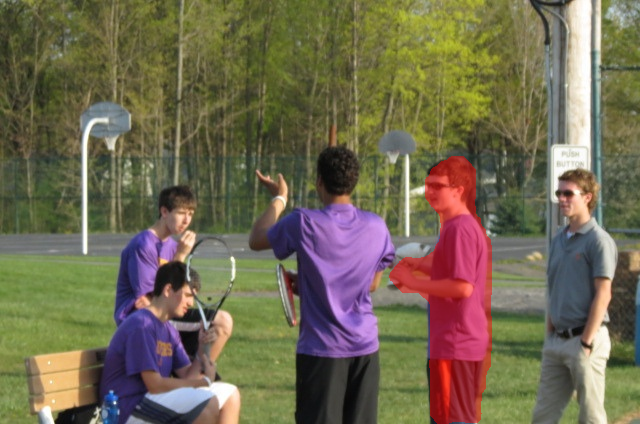} \\
\end{tabular}}

\caption{ Qualitative comparison on RefCOCO. Each row corresponds to a referring expression, with columns showing the original image, ground-truth mask, DETRIS prediction, and SERA (ours) prediction.}
\label{fig:qualitative_refcoco}
\end{figure}

\section{Results and Discussions}\label{sec2b}
This section evaluates the proposed SERA framework through quantitative comparisons, ablation studies, and qualitative analyses on standard referring image segmentation benchmarks.

\subsection{Comparison with State of the Art}

Table~\ref{tab:sota_refcoco} reports quantitative comparisons between SERA and recent referring image segmentation methods under both full fine-tuning and parameter-efficient tuning (PET) settings.
We follow the standard evaluation protocol and report mean Intersection-over-Union (mIoU) on official validation and test splits. For fair comparison with parameter-efficient methods, SERA follows the same frozen-backbone training protocol adopted by prior PET-based RIS approaches. As shown in Table~\ref{tab:sota_refcoco}, SERA achieves competitive performance across RefCOCO, RefCOCO+, and G-Ref. Under a frozen-backbone training protocol, SERA consistently outperforms prior parameter-efficient approaches while remaining competitive with several fully fine-tuned models. Notably, larger gains are achieved on RefCOCO+, which excludes absolute spatial terms and emphasizes appearance-driven and context-driven reasoning.
Results on G-Ref further indicate stable performance under longer and more descriptive expressions. A qualitative analysis is provided in the following subsection to examine the nature of these improvements further.

\begin{figure*}
\centering
\setlength{\tabcolsep}{0.8pt} 

\begin{tabular}{c c c c c c c c c c} 

 & \hyperlink{p1}{1}
 & \hyperlink{p2}{2}
 & \hyperlink{p3}{3}
 & \hyperlink{p4}{4}
 & \hyperlink{p5}{5}
 & \hyperlink{p6}{6}
 & \hyperlink{p7}{7}
 & \hyperlink{p8}{8}
 & \hyperlink{p9}{9} \\

\rotatebox{90}{\hspace{3mm}Input} &
\includegraphics[width=0.108\textwidth,height=1.85cm]{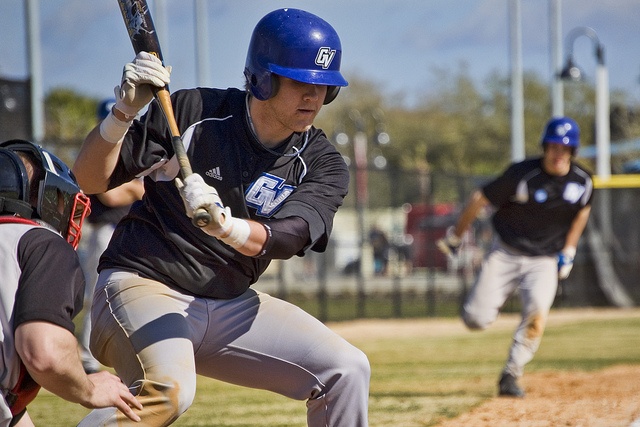} &
\includegraphics[width=0.108\textwidth,height=1.85cm]{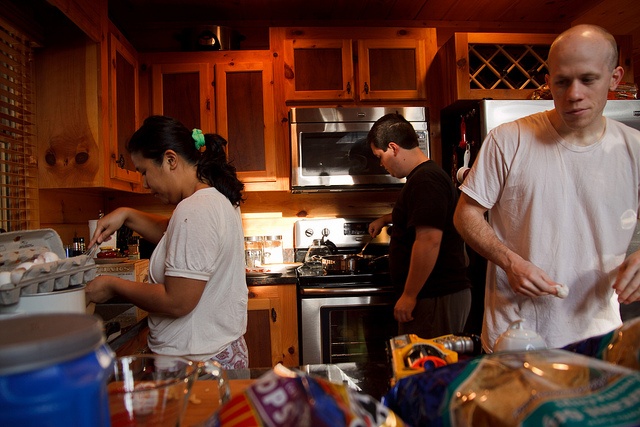} &
\includegraphics[width=0.108\textwidth,height=1.85cm]{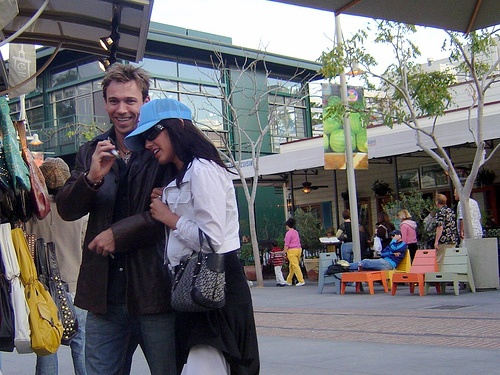} &
\includegraphics[width=0.108\textwidth,height=1.85cm]{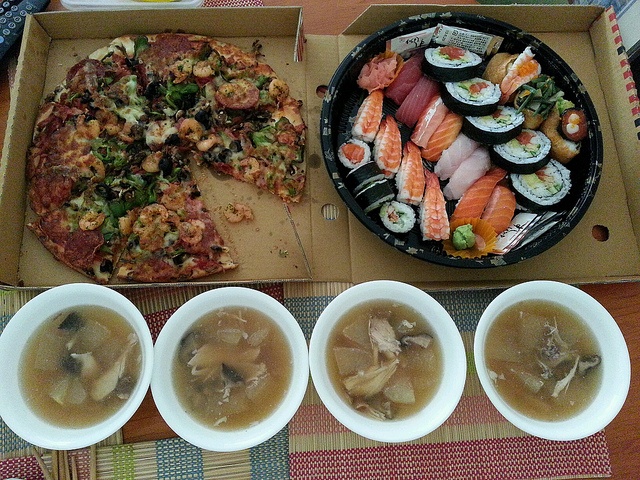} &
\includegraphics[width=0.108\textwidth,height=1.85cm]{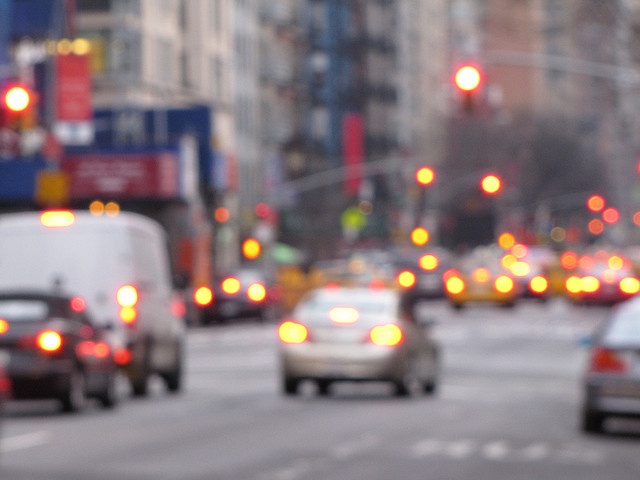} &
\includegraphics[width=0.108\textwidth,height=1.85cm]{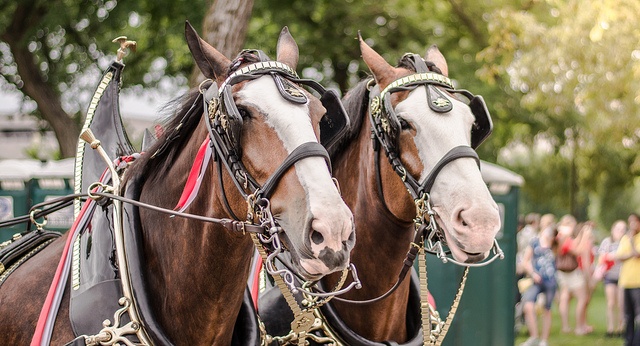} &
\includegraphics[width=0.108\textwidth,height=1.85cm]{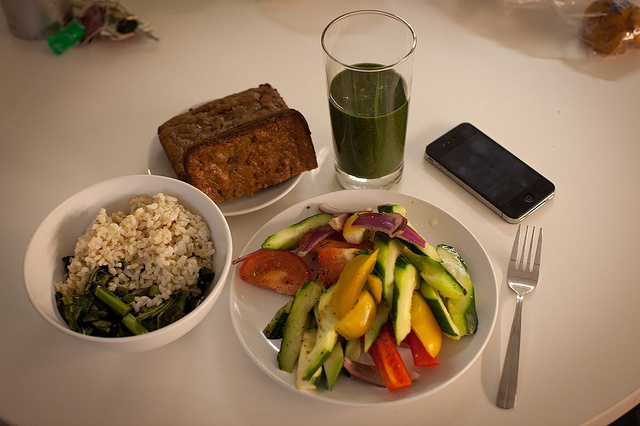} &
\includegraphics[width=0.108\textwidth,height=1.85cm]{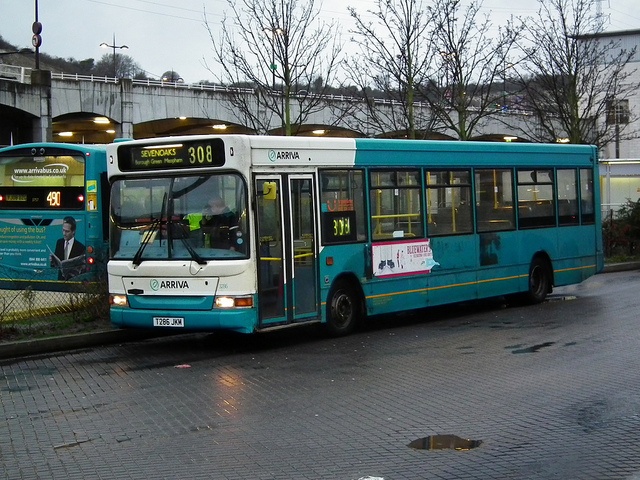} &
\includegraphics[width=0.108\textwidth,height=1.85cm]{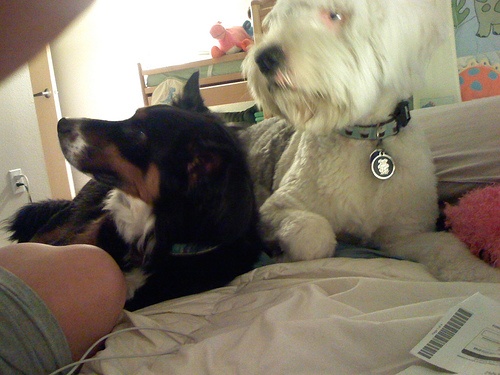}\\[-0.5mm]

\rotatebox{90}{\hspace{1mm}GT Mask} &
\includegraphics[width=0.108\textwidth,height=1.85cm]{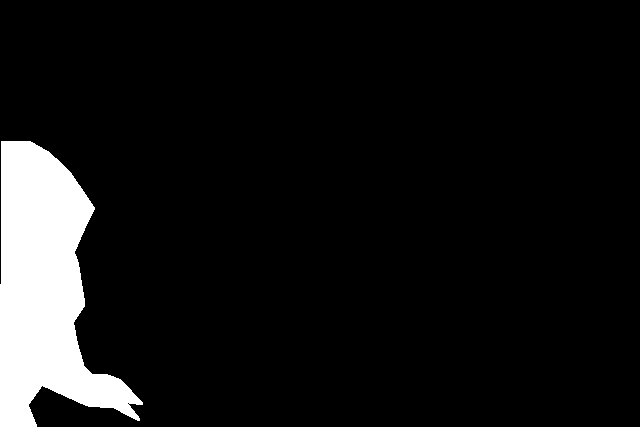} &
\includegraphics[width=0.108\textwidth,height=1.85cm]{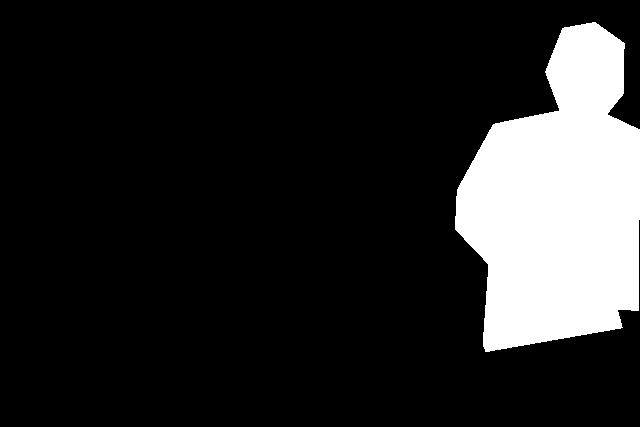} &
\includegraphics[width=0.108\textwidth,height=1.85cm]{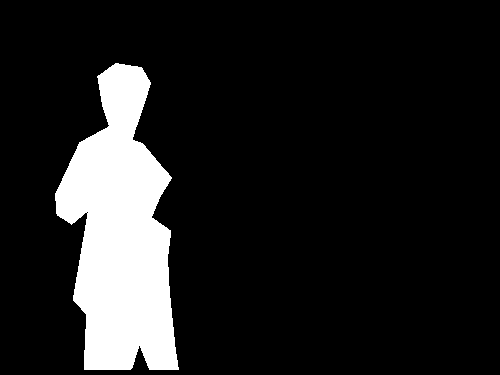} &
\includegraphics[width=0.108\textwidth,height=1.85cm]{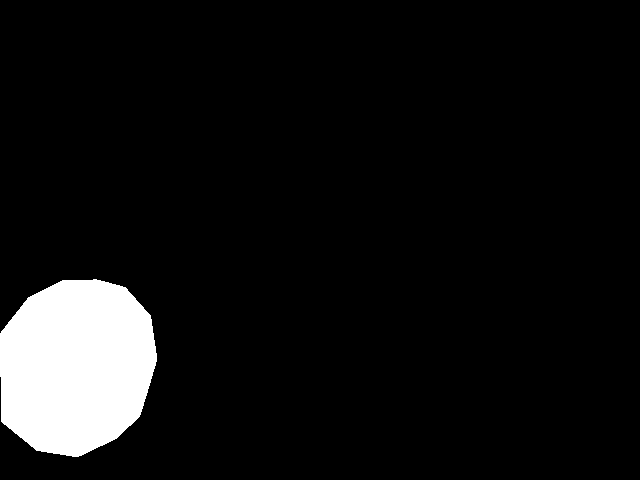} &
\includegraphics[width=0.108\textwidth,height=1.85cm]{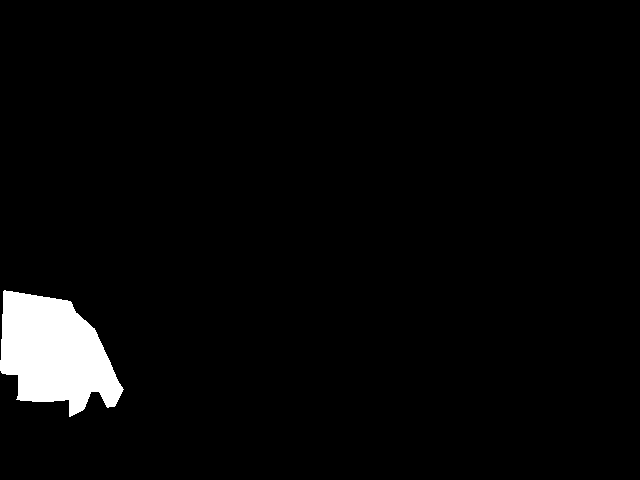} &
\includegraphics[width=0.108\textwidth,height=1.85cm]{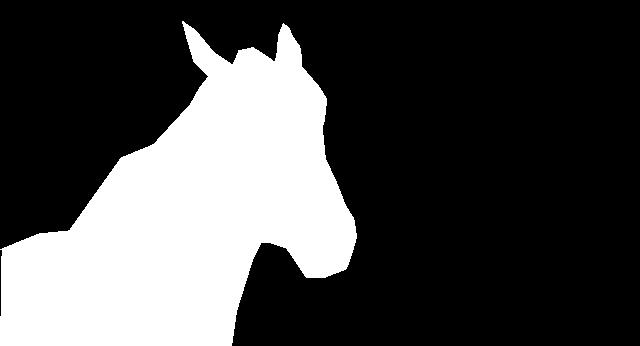} &
\includegraphics[width=0.108\textwidth,height=1.85cm]{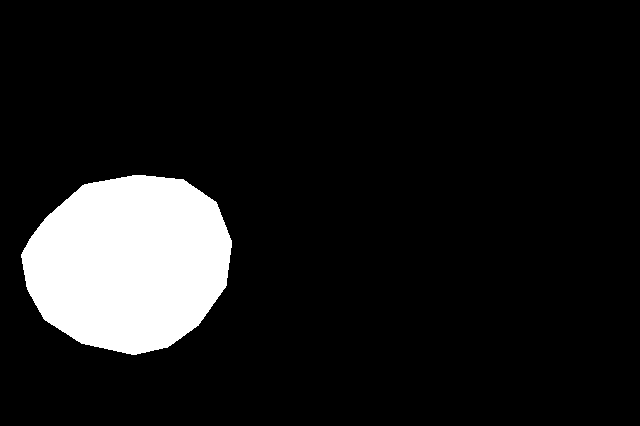} &
\includegraphics[width=0.108\textwidth,height=1.85cm]{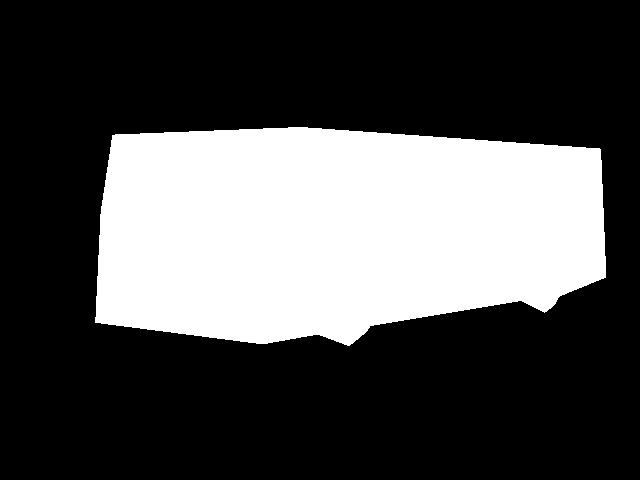} &
\includegraphics[width=0.108\textwidth,height=1.85cm]{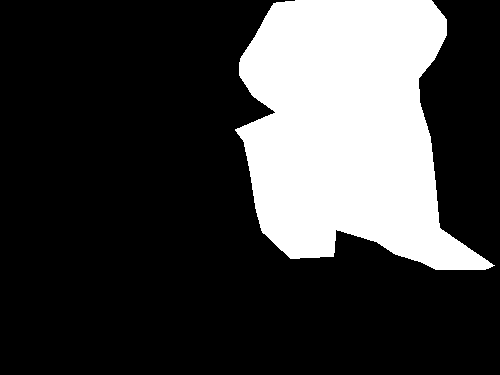} \\[-0.5mm]

\rotatebox{90}{\hspace{5mm}Ours} &
\shortstack{
\includegraphics[width=0.108\textwidth,height=1.85cm]{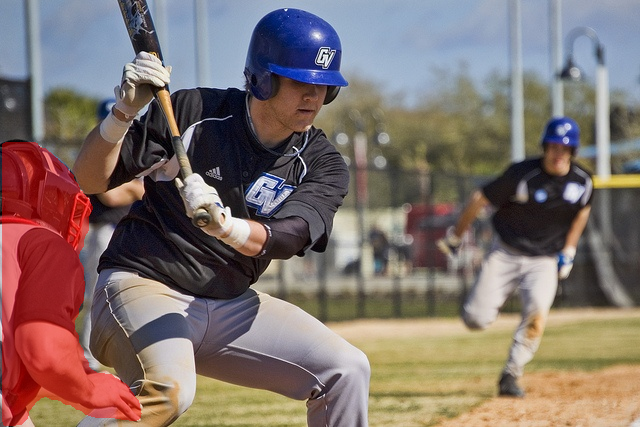} \\
{\scriptsize 91.06}
} &
\shortstack{
\includegraphics[width=0.108\textwidth,height=1.85cm]{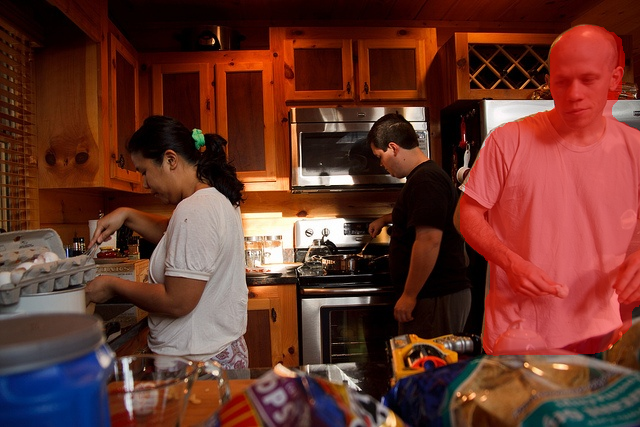} \\
{\scriptsize 92.47}
} &
\shortstack{
\includegraphics[width=0.108\textwidth,height=1.85cm]{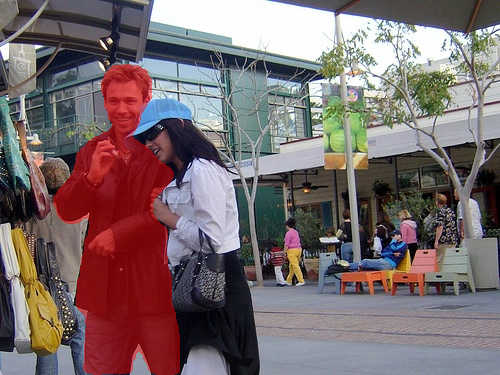} \\
{\scriptsize 91.83}
} &
\shortstack{
\includegraphics[width=0.108\textwidth,height=1.85cm]{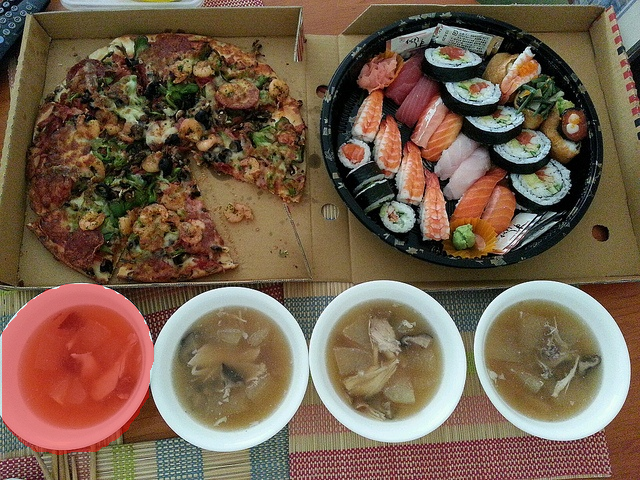} \\
{\scriptsize 93.63}
} &
\shortstack{
\includegraphics[width=0.108\textwidth,height=1.85cm]{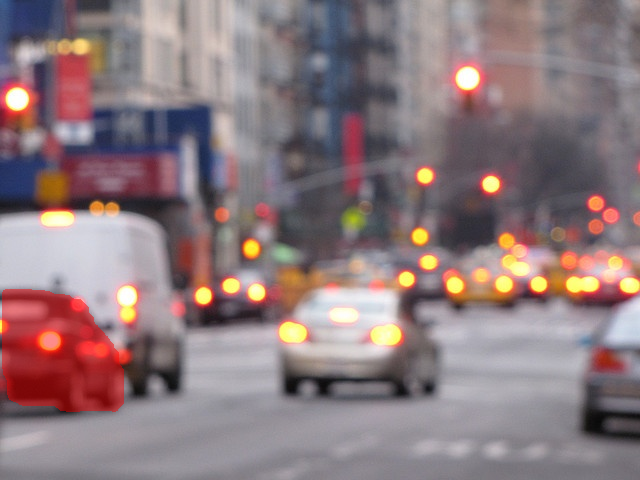} \\
{\scriptsize 81.71}
} &
\shortstack{
\includegraphics[width=0.108\textwidth,height=1.85cm]{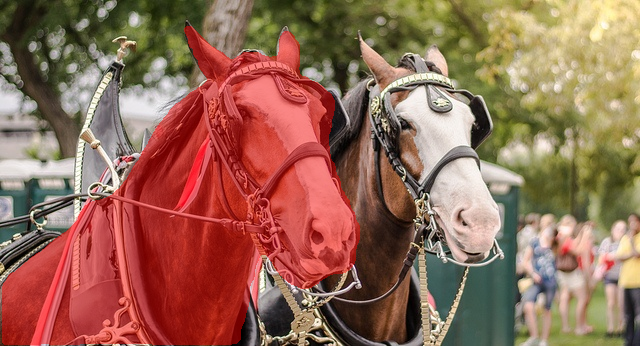} \\
{\scriptsize 90.36}
} &
\shortstack{
\includegraphics[width=0.108\textwidth,height=1.85cm]{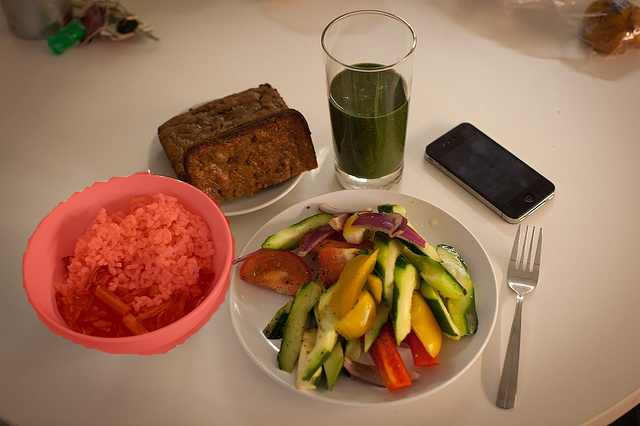} \\
{\scriptsize 94.77}
} &
\shortstack{
\includegraphics[width=0.108\textwidth,height=1.85cm]{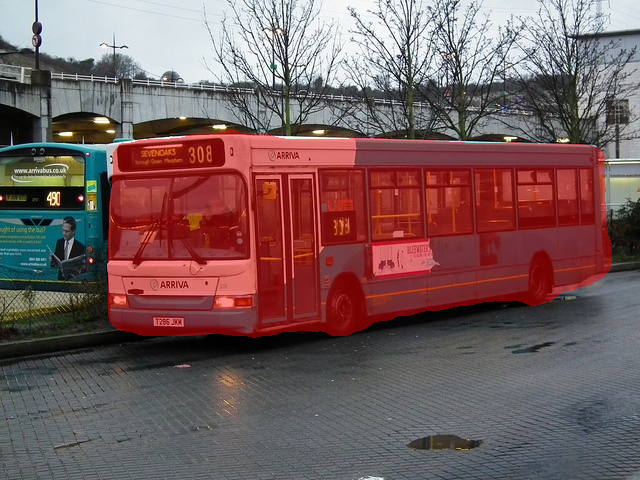} \\
{\scriptsize 93.09}
} &
\shortstack{
\includegraphics[width=0.108\textwidth,height=1.85cm]{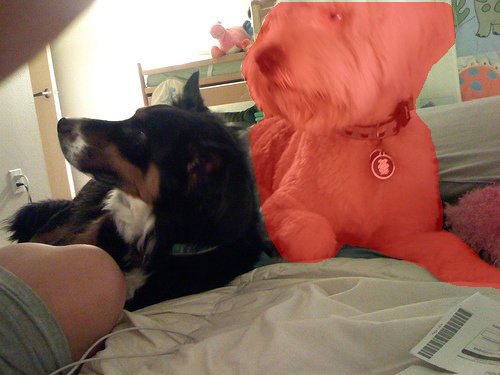} \\
{\scriptsize 90.50}
} \\

\end{tabular}

\caption{  Qualitative results on the RefCOCO dataset. From top to bottom: input images with referring expressions, ground-truth masks, and our predicted segmentation masks. IoU values are shown below each prediction. The clickable sample IDs correspond to those listed in Appendix~\ref{sec:appendix}.}
\label{fig:qual1}
\end{figure*}

\begin{figure*}
\centering
\setlength{\tabcolsep}{0.8pt}

\begin{tabular}{c c c c c c c c c c} 

 & \hyperlink{p10}{10}
 & \hyperlink{p11}{11}
 & \hyperlink{p12}{12}
 & \hyperlink{p13}{13}
 & \hyperlink{p14}{14}
 & \hyperlink{p15}{15}
 & \hyperlink{p16}{16}
 & \hyperlink{p17}{17}
 & \hyperlink{p18}{18} \\

\rotatebox{90}{\hspace{3mm}Input} &
\includegraphics[width=0.108\textwidth,height=1.85cm]{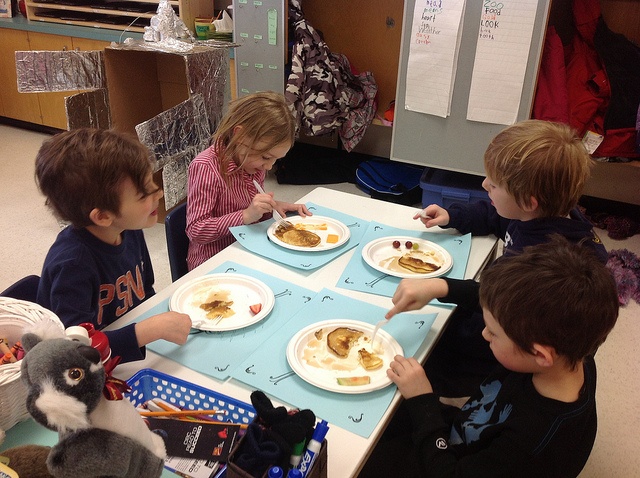} &
\includegraphics[width=0.108\textwidth,height=1.85cm]{refcocoplusQua/409-img.jpg} &
\includegraphics[width=0.108\textwidth,height=1.85cm]{refcocoplusQua/410-img.jpg} &
\includegraphics[width=0.108\textwidth,height=1.85cm]{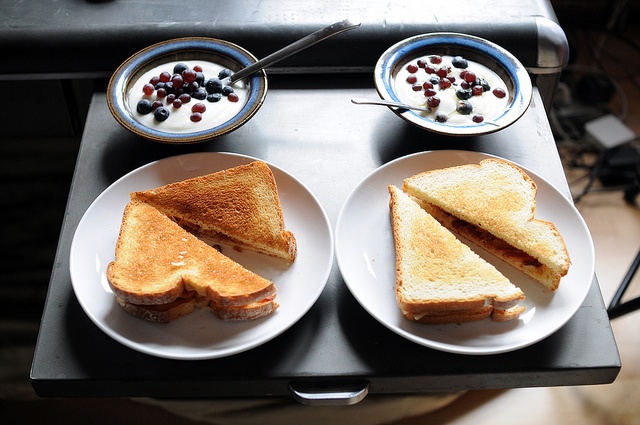} &
\includegraphics[width=0.108\textwidth,height=1.85cm]{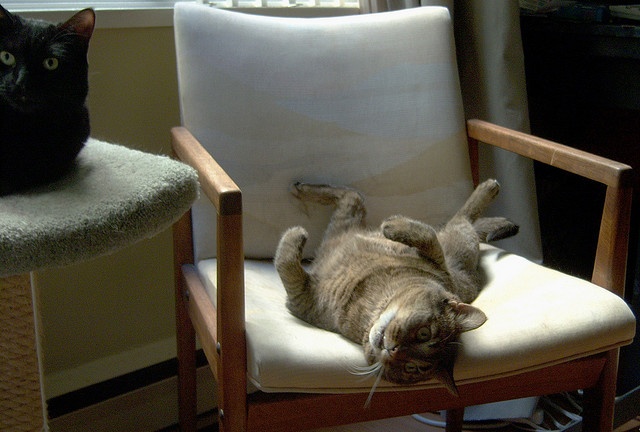} &
\includegraphics[width=0.108\textwidth,height=1.85cm]{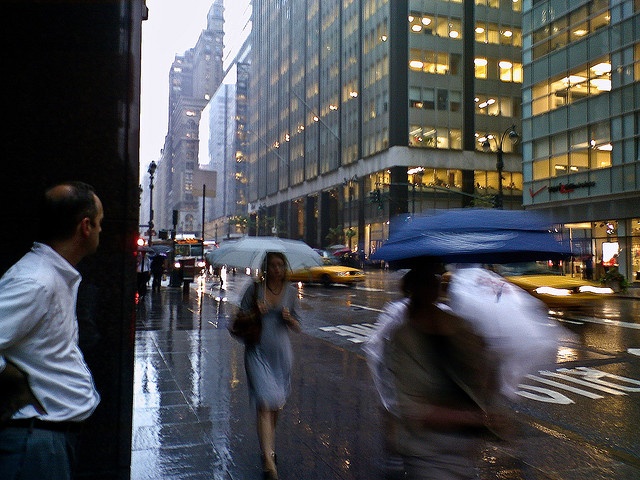} &
\includegraphics[width=0.108\textwidth,height=1.85cm]{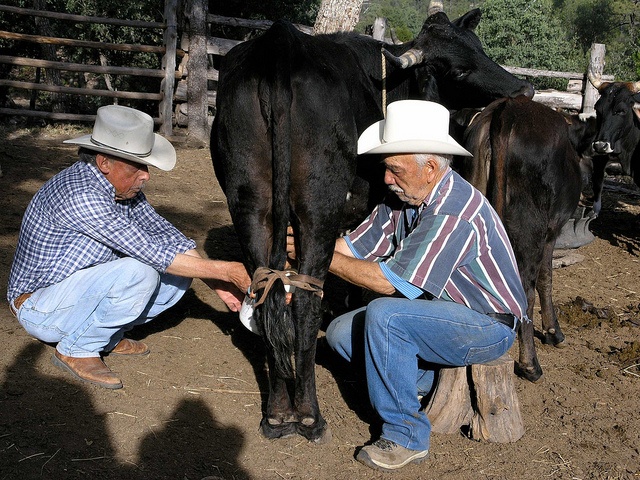} &
\includegraphics[width=0.108\textwidth,height=1.85cm]{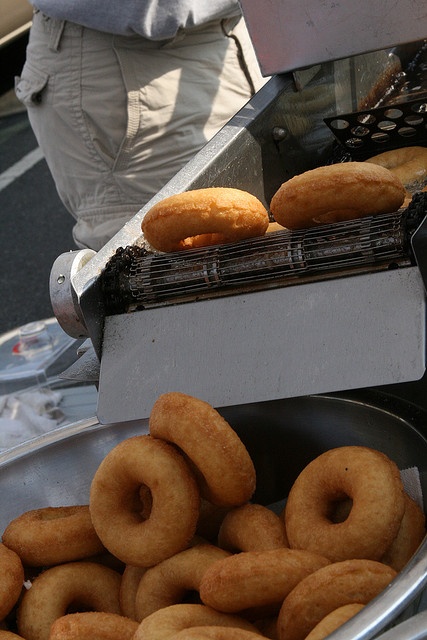} &
\includegraphics[width=0.108\textwidth,height=1.85cm]{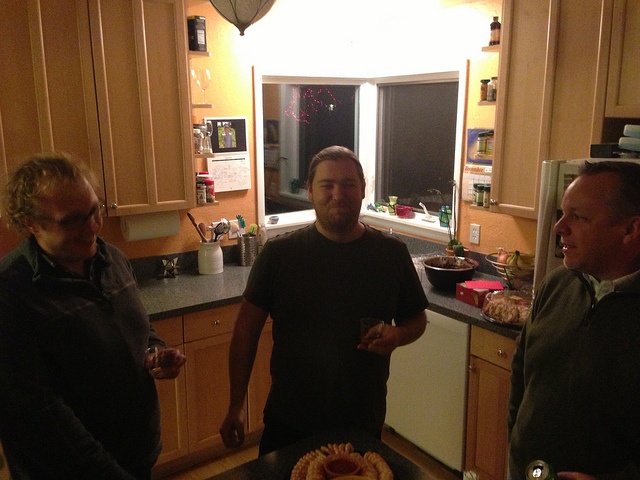} \\[-1mm]


\rotatebox{90}{\hspace{1mm}GT Mask} &
\includegraphics[width=0.108\textwidth,height=1.85cm]{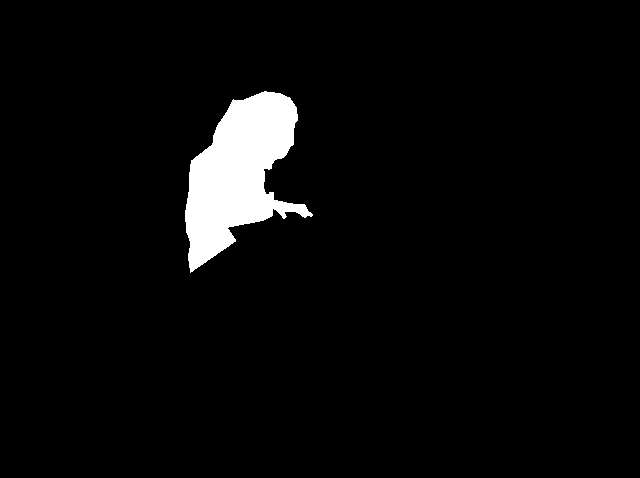} &
\includegraphics[width=0.108\textwidth,height=1.85cm]{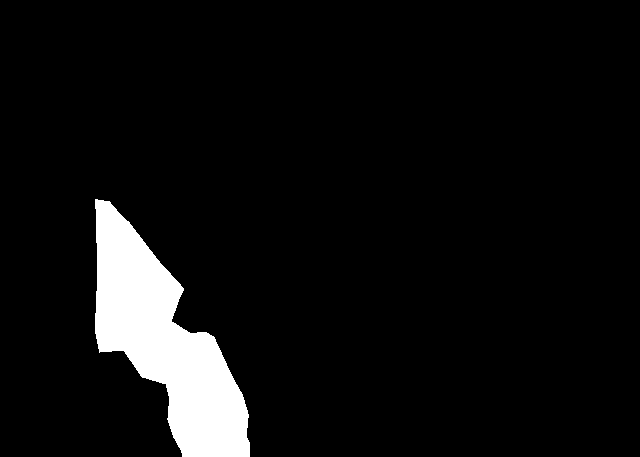} &
\includegraphics[width=0.108\textwidth,height=1.85cm]{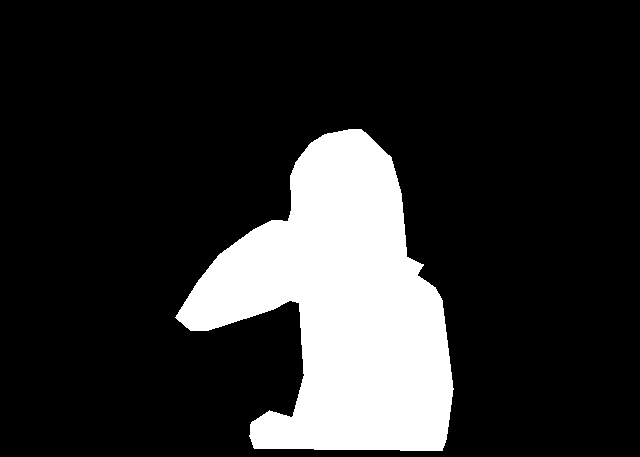} &
\includegraphics[width=0.108\textwidth,height=1.85cm]{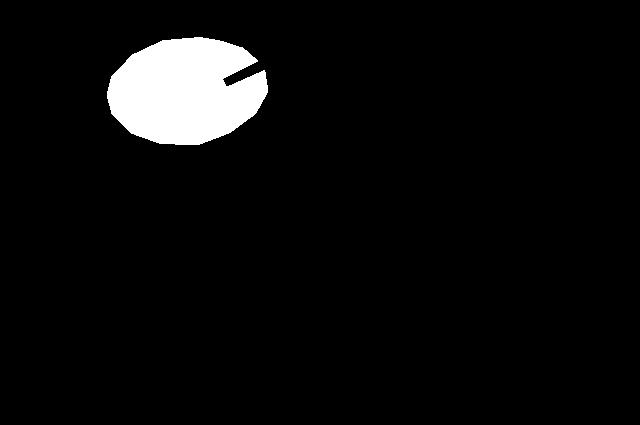} &
\includegraphics[width=0.108\textwidth,height=1.85cm]{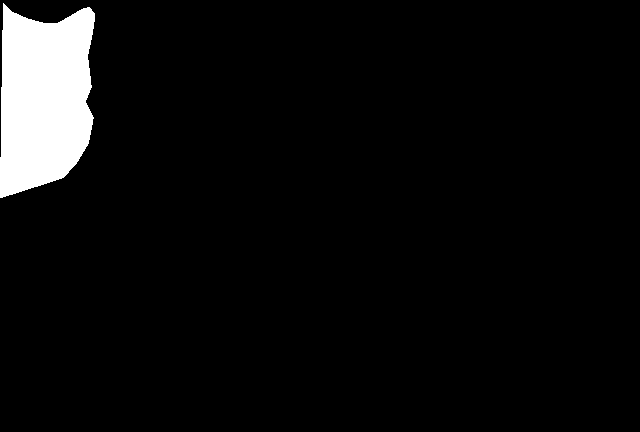} &
\includegraphics[width=0.108\textwidth,height=1.85cm]{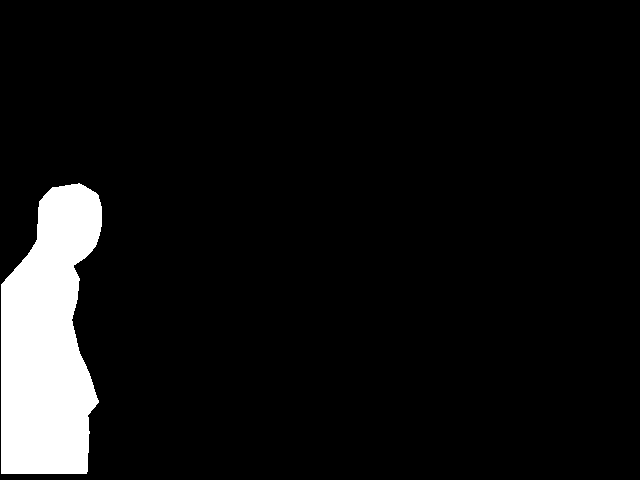} &
\includegraphics[width=0.108\textwidth,height=1.85cm]{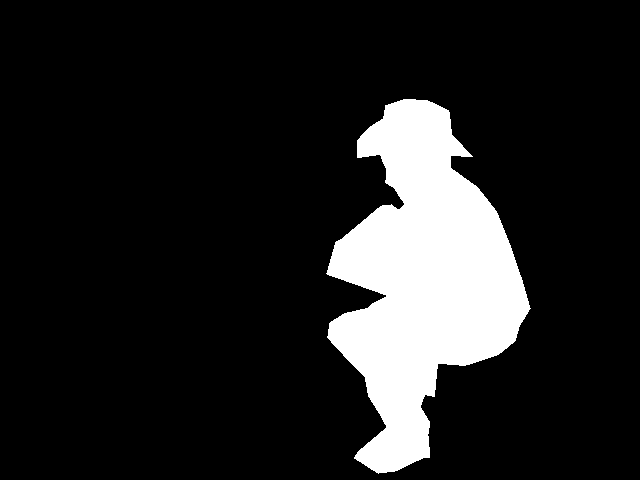} &
\includegraphics[width=0.108\textwidth,height=1.85cm]{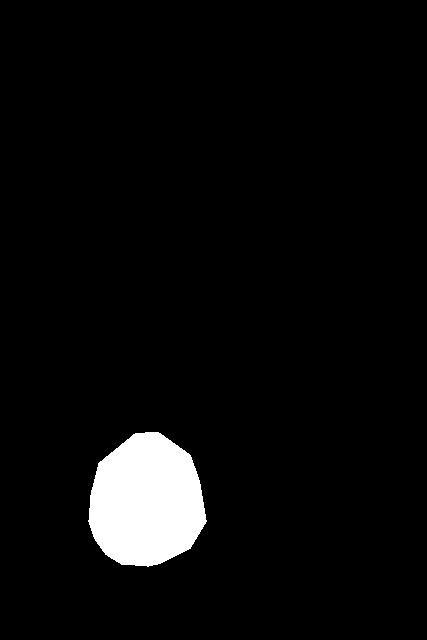} &
\includegraphics[width=0.108\textwidth,height=1.85cm]{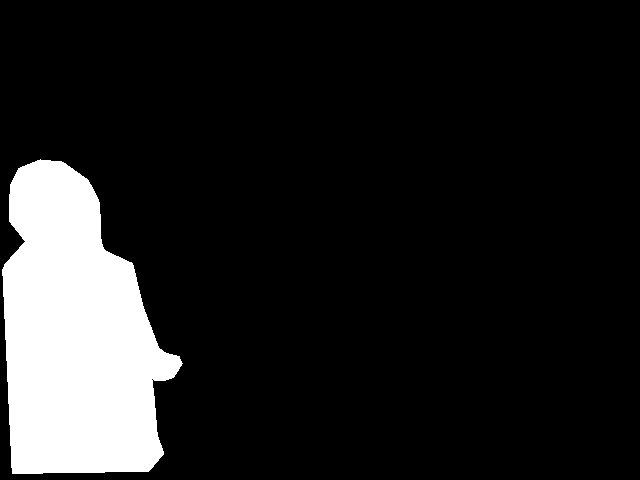} \\[-1mm]

\rotatebox{90}{\hspace{5mm}Ours} &
\shortstack{
\includegraphics[width=0.108\textwidth,height=1.85cm]{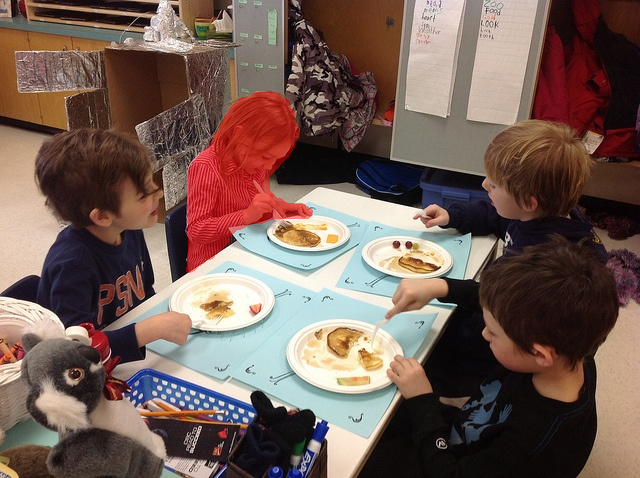} \\
{\scriptsize 90.21}
} &
\shortstack{
\includegraphics[width=0.108\textwidth,height=1.85cm]{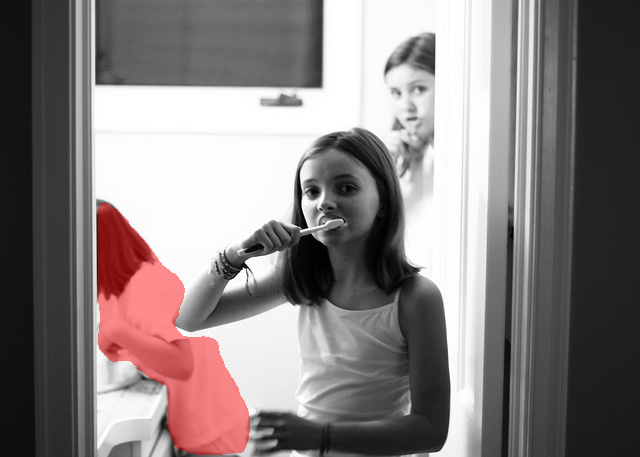} \\
{\scriptsize 91.87}
} &
\shortstack{
\includegraphics[width=0.108\textwidth,height=1.85cm]{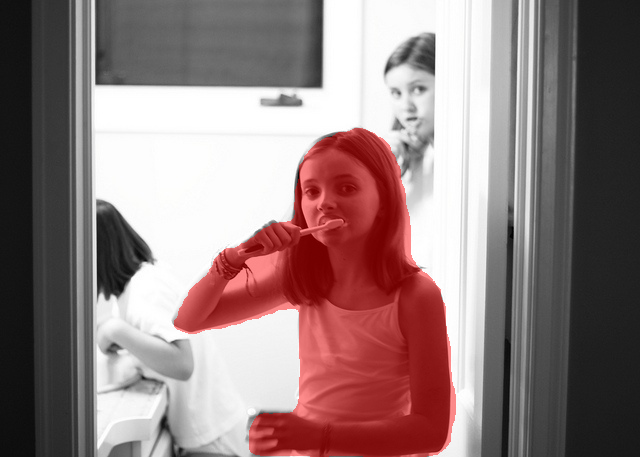} \\
{\scriptsize 92.33}
} &
\shortstack{
\includegraphics[width=0.108\textwidth,height=1.85cm]{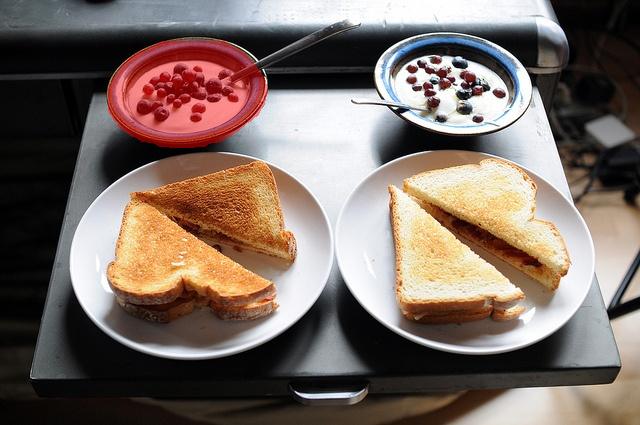} \\
{\scriptsize 90.93}
} &
\shortstack{
\includegraphics[width=0.108\textwidth,height=1.85cm]{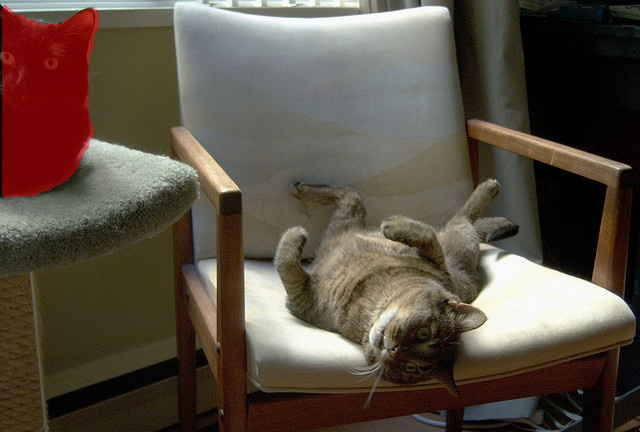} \\
{\scriptsize 93.07}
} &
\shortstack{
\includegraphics[width=0.108\textwidth,height=1.85cm]{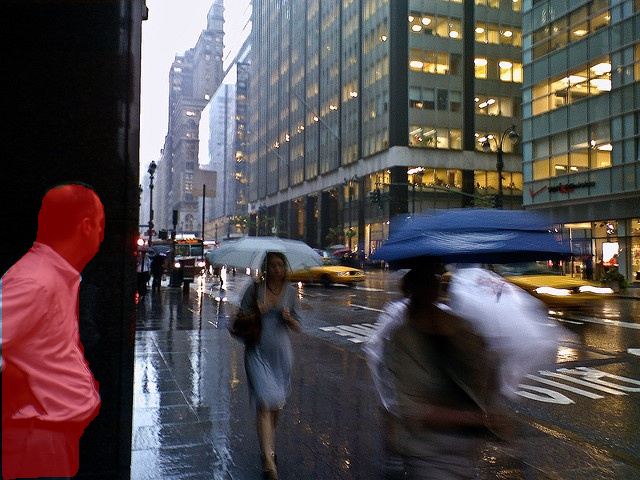} \\
{\scriptsize 91.43}
} &
\shortstack{
\includegraphics[width=0.108\textwidth,height=1.85cm]{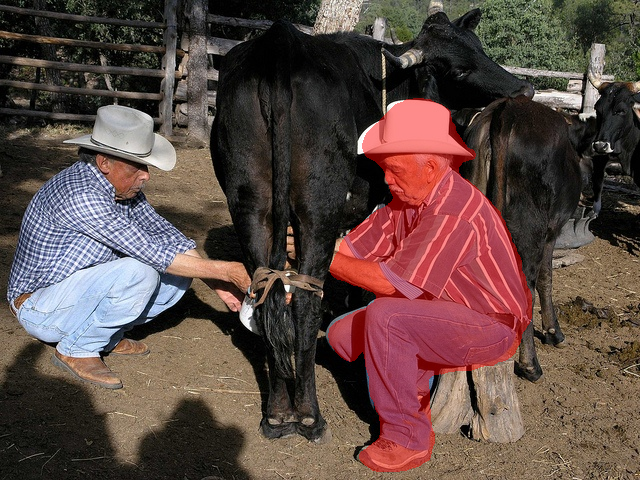} \\
{\scriptsize 90.85}
} &
\shortstack{
\includegraphics[width=0.108\textwidth,height=1.85cm]{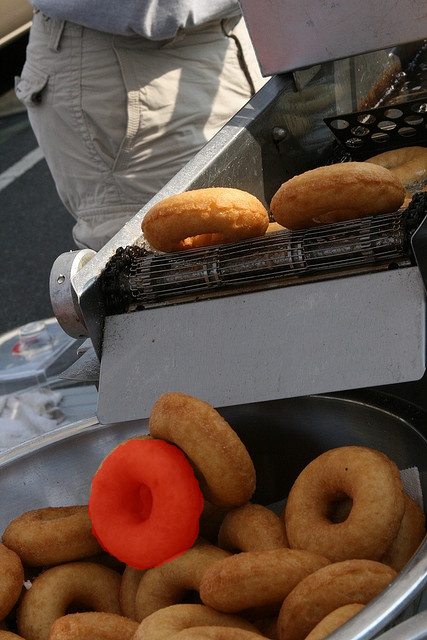} \\
{\scriptsize 92.29}
} &
\shortstack{
\includegraphics[width=0.108\textwidth,height=1.85cm]{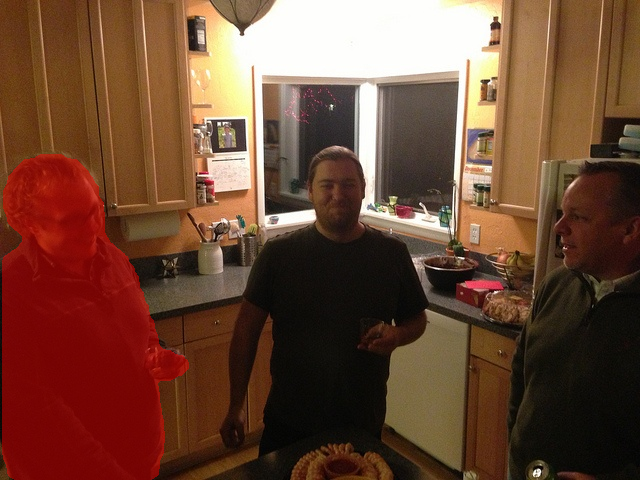} \\
{\scriptsize 90.11}
} \\

\end{tabular}

\caption{ Qualitative results on the RefCOCO+ dataset. From top to bottom: input images with referring expressions, ground-truth masks, and our predicted segmentation masks. IoU scores are reported below each prediction. The clickable sample IDs correspond to Appendix~\ref{sec:appendix}.}
\label{fig:qual2}
\end{figure*}

\begin{figure*} 
\centering
\setlength{\tabcolsep}{0.8pt}

\begin{tabular}{c c c c c c c c c c}

 & \hyperlink{p19}{19}
 & \hyperlink{p20}{20}
 & \hyperlink{p21}{21}
 & \hyperlink{p22}{22}
 & \hyperlink{p23}{23}
 & \hyperlink{p24}{24}
 & \hyperlink{p25}{25}
 & \hyperlink{p26}{26}
 & \hyperlink{p27}{27} \\

\rotatebox{90}{\hspace{3mm}Input} &
\includegraphics[width=0.108\textwidth,height=1.86cm]{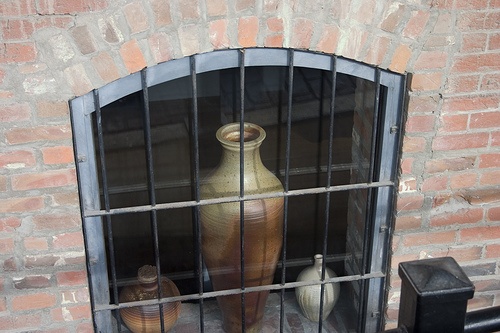} &
\includegraphics[width=0.108\textwidth,height=1.86cm]{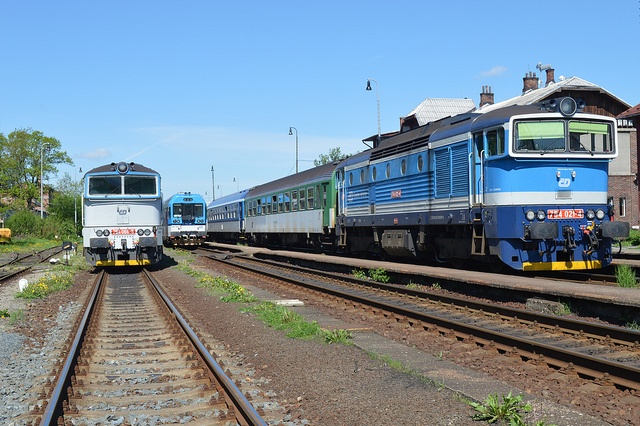} &
\includegraphics[width=0.108\textwidth,height=1.86cm]{refcocoGQua/13-img.jpg} &
\includegraphics[width=0.108\textwidth,height=1.86cm]{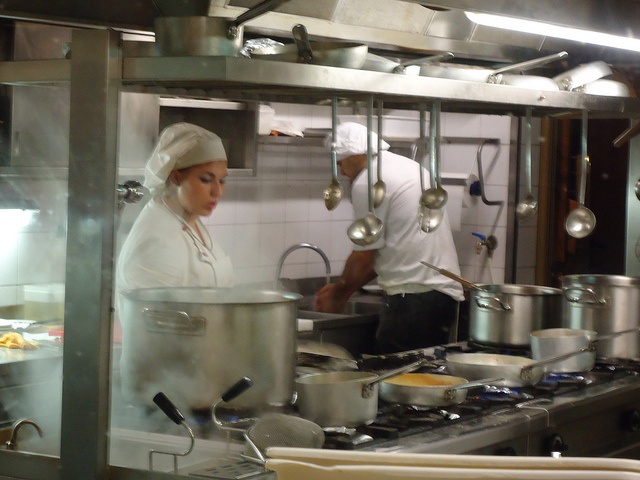} &
\includegraphics[width=0.108\textwidth,height=1.86cm]{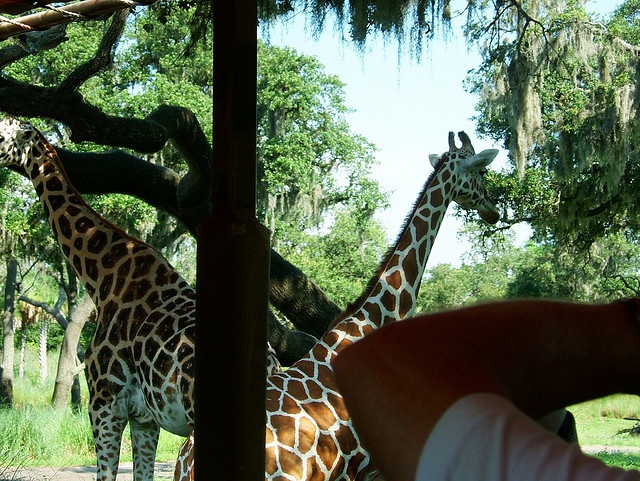} &
\includegraphics[width=0.108\textwidth,height=1.86cm]{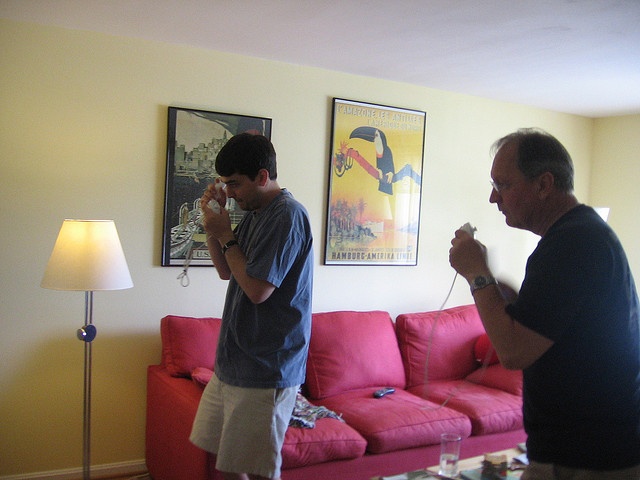} &
\includegraphics[width=0.108\textwidth,height=1.86cm]{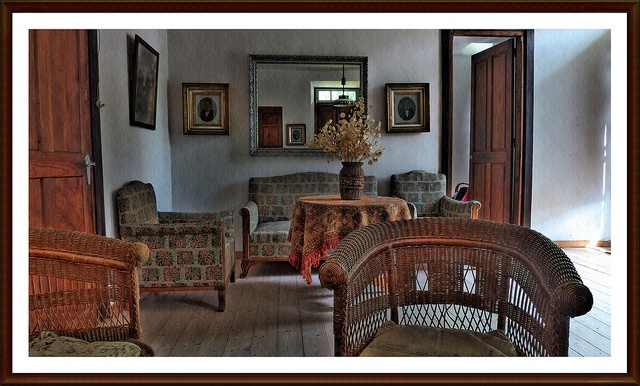} &
\includegraphics[width=0.108\textwidth,height=1.86cm]{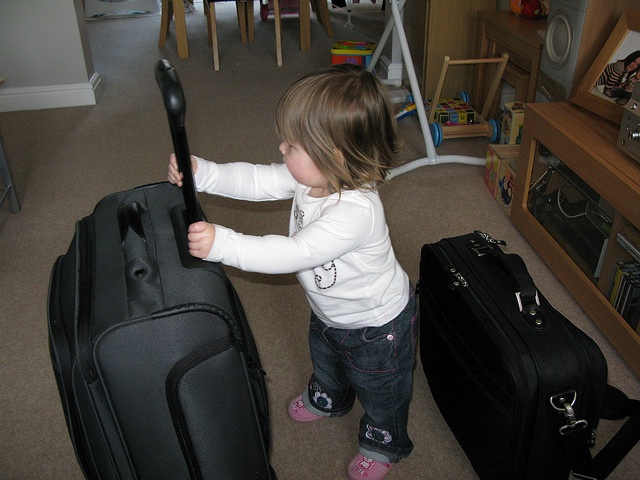} &
\includegraphics[width=0.108\textwidth,height=1.86cm]{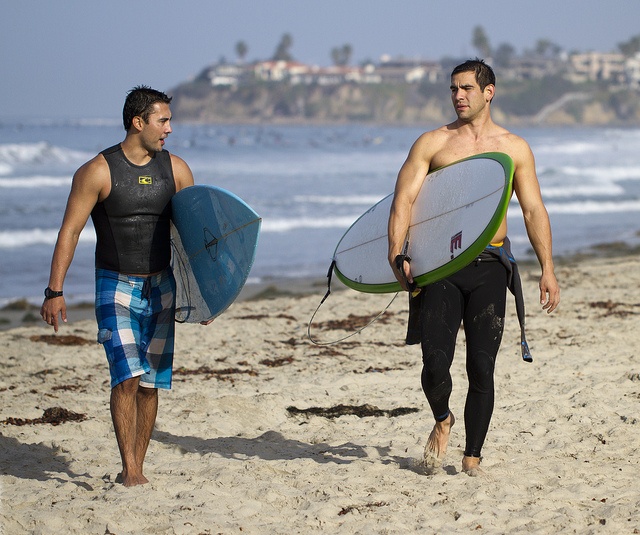} \\[-1mm]

\rotatebox{90}{\hspace{1mm}GT Mask} &
\includegraphics[width=0.108\textwidth,height=1.86cm]{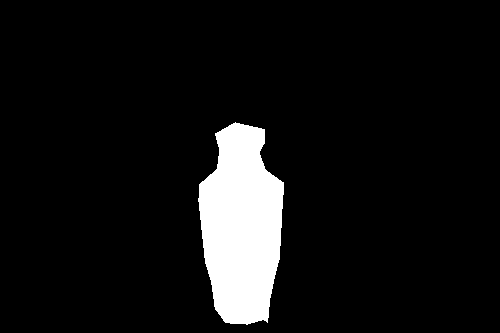} &
\includegraphics[width=0.108\textwidth,height=1.86cm]{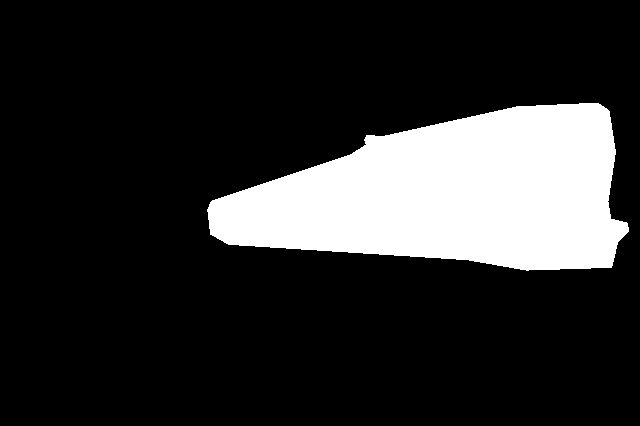} &
\includegraphics[width=0.108\textwidth,height=1.86cm]{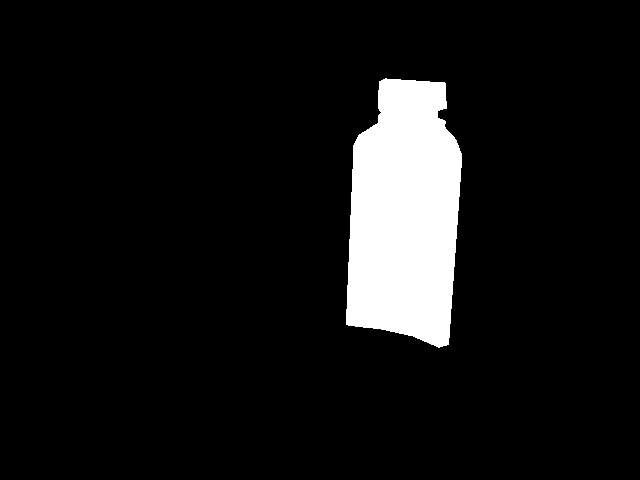} &
\includegraphics[width=0.108\textwidth,height=1.86cm]{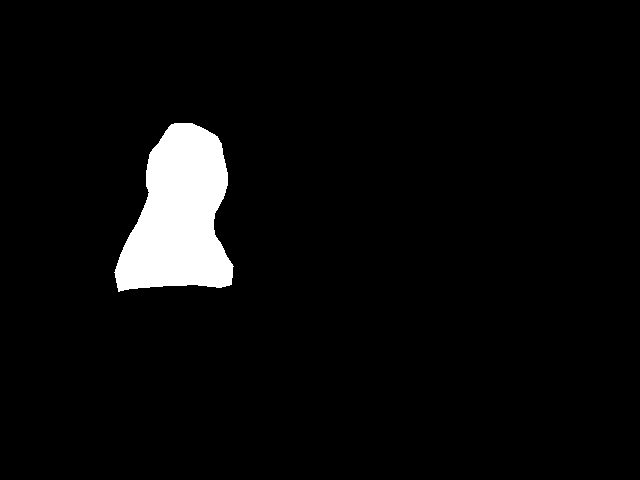} &
\includegraphics[width=0.108\textwidth,height=1.86cm]{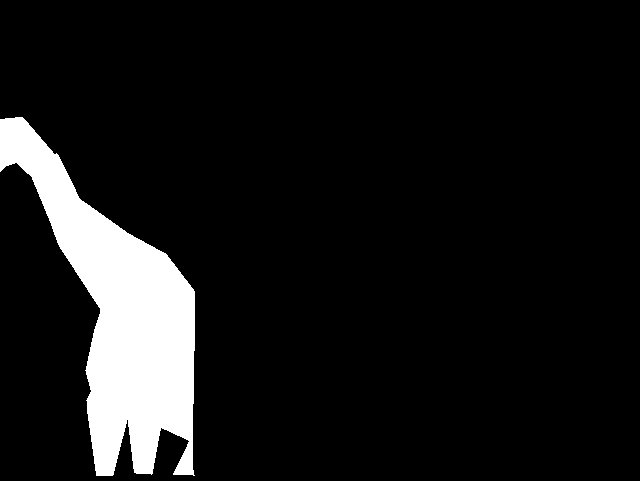} &
\includegraphics[width=0.108\textwidth,height=1.86cm]{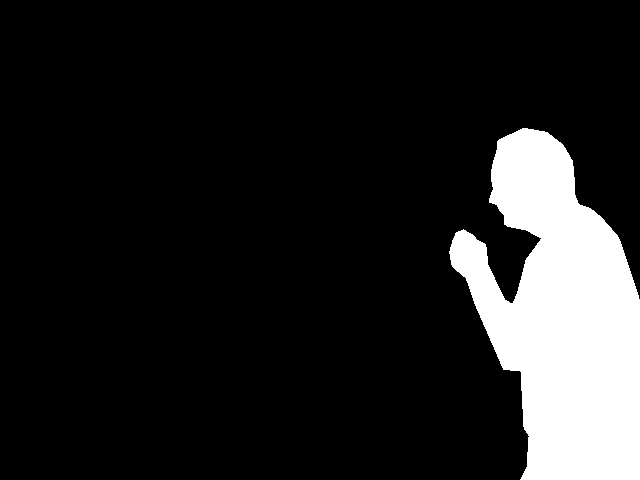} &
\includegraphics[width=0.108\textwidth,height=1.86cm]{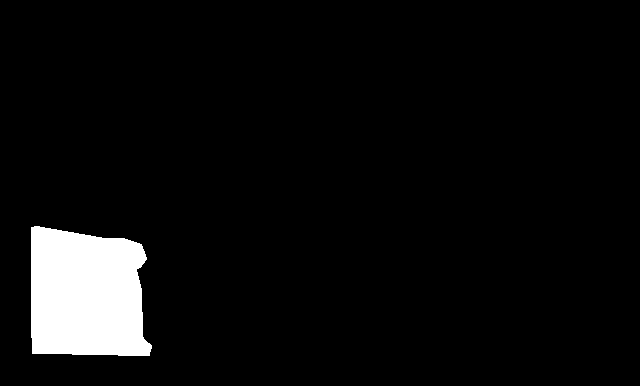} &
\includegraphics[width=0.108\textwidth,height=1.86cm]{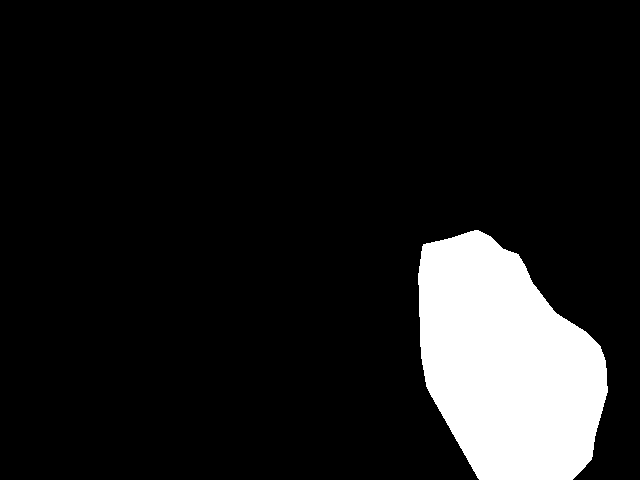} &
\includegraphics[width=0.108\textwidth,height=1.86cm]{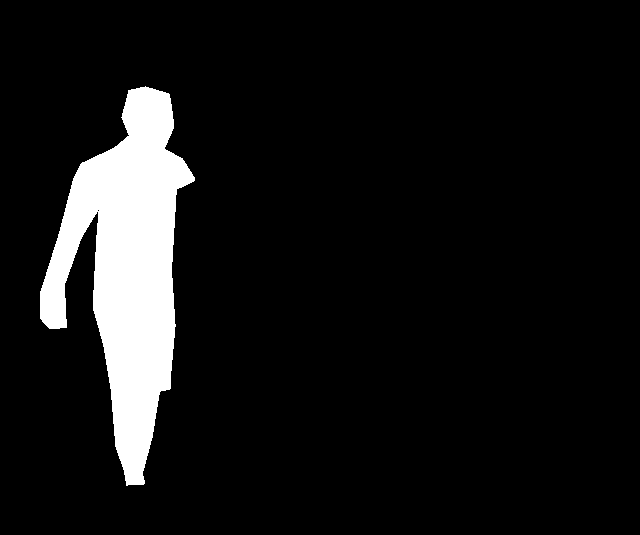} \\[-1mm]

\rotatebox{90}{\hspace{5mm}Ours} &
\shortstack{
\includegraphics[width=0.108\textwidth,height=1.86cm]{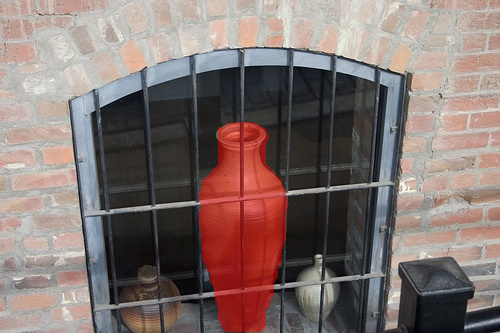} \\
{\scriptsize 90.51}
} &
\shortstack{
\includegraphics[width=0.108\textwidth,height=1.86cm]{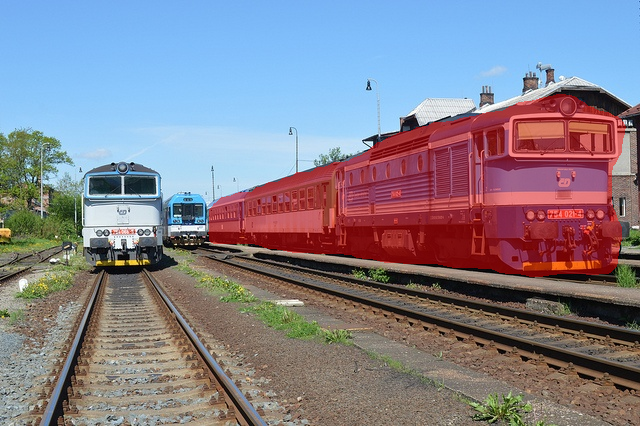} \\
{\scriptsize 92.03}
} &
\shortstack{
\includegraphics[width=0.108\textwidth,height=1.86cm]{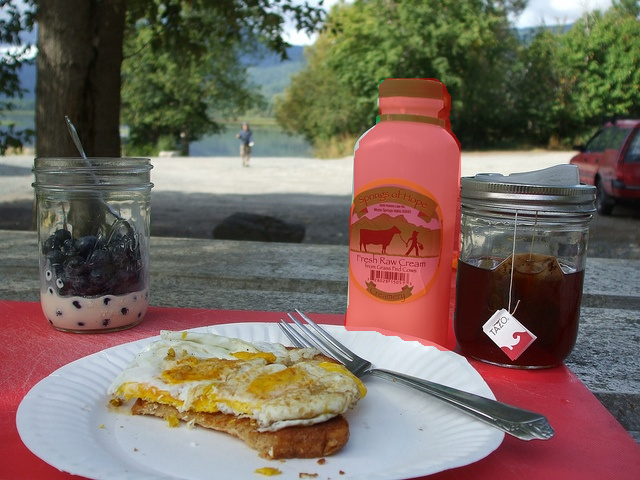} \\
{\scriptsize 93.26}
} &
\shortstack{
\includegraphics[width=0.108\textwidth,height=1.86cm]{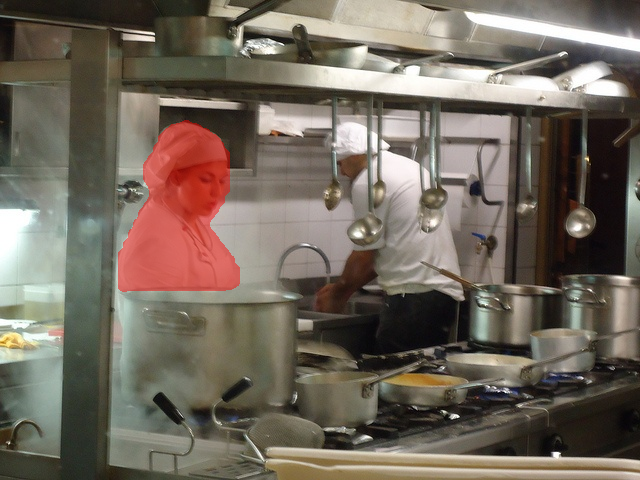} \\
{\scriptsize 90.84}
} &
\shortstack{
\includegraphics[width=0.108\textwidth,height=1.86cm]{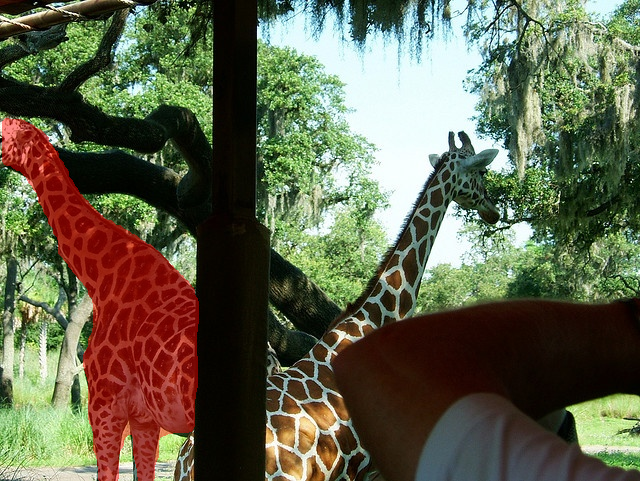} \\
{\scriptsize 90.37}
} &
\shortstack{
\includegraphics[width=0.108\textwidth,height=1.86cm]{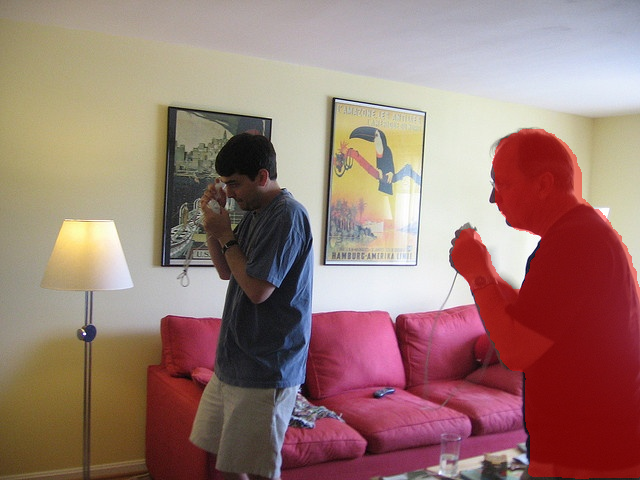} \\
{\scriptsize 93.84}
} &
\shortstack{
\includegraphics[width=0.108\textwidth,height=1.86cm]{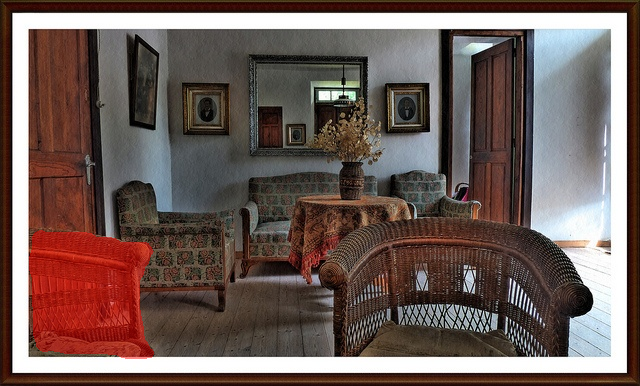} \\
{\scriptsize 93.09}
} &
\shortstack{
\includegraphics[width=0.108\textwidth,height=1.86cm]{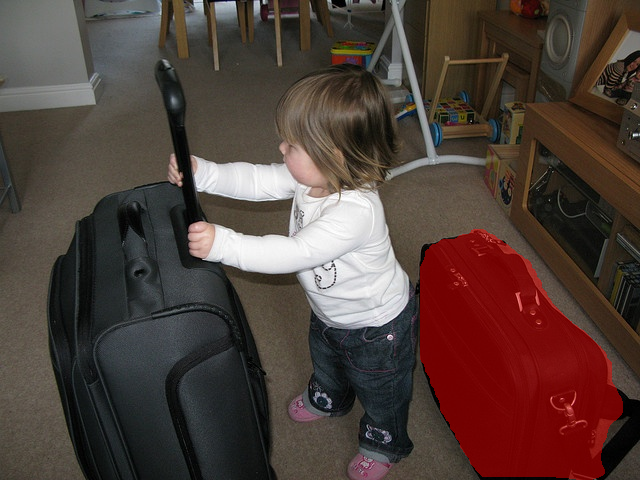} \\
{\scriptsize 92.07}
} &
\shortstack{
\includegraphics[width=0.108\textwidth,height=1.86cm]{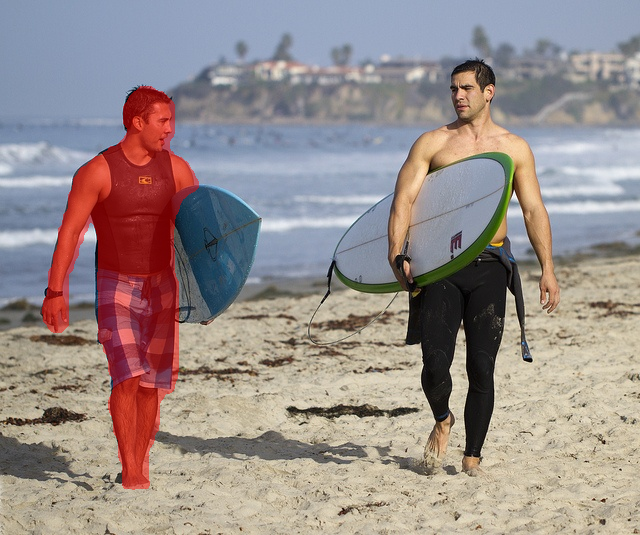} \\
{\scriptsize 90.03}
} \\

\end{tabular}

\caption{ Qualitative results on the RefCOCOg dataset. From top to bottom: input images with referring expressions, ground-truth masks, and our predicted segmentation masks. IoU scores are shown below each prediction. The clickable sample IDs correspond to Appendix~\ref{sec:appendix}.}
\label{fig:qual3}
\end{figure*}

\subsection{Qualitative Results}
We provide qualitative comparisons to further examine the behavior of SERA alongside the quantitative results. Figure~\ref{fig:qualitative_refcoco} illustrates segmentation results from DETRIS and SERA together with the input images and ground-truth masks. In challenging cases such as cluttered scenes, small objects, or visually similar distractors, the baseline often produces incomplete masks or activates nearby instances. In contrast, SERA generates more coherent predictions that align better with the ground-truth regions, particularly in cases with ambiguous boundaries or complex backgrounds. This indicates that expert-guided refinement improves both localization accuracy and boundary quality.

Additional examples demonstrating the generalization ability of SERA are presented in Figures~\ref{fig:qual1}-\ref{fig:qual3}. In Figure~\ref{fig:qual1}, SERA consistently localizes small or partially occluded objects and preserves stable object shapes in cluttered scenes on RefCOCO. In Figure~\ref{fig:qual2}, which corresponds to RefCOCO+, where referring expressions rely mainly on appearance cues, SERA continues to produce accurate segmentations even without location-based information. Finally, Figure~\ref{fig:qual3} presents results on RefCOCOg, which contains longer and more descriptive expressions. Even in these linguistically complex cases, SERA generates coherent masks by effectively using contextual information from the expressions. Together, these examples highlight the robustness of SERA across datasets with different linguistic characteristics, complementing the quantitative improvements reported in Table~\ref{tab:sota_refcoco}.

\subsection{Ablation Study}
\label{sec:ablation}

We conduct ablation experiments to examine the contribution of each SERA component and the effect of routing sparsity. Unless noted otherwise, all settings follow the same training and evaluation protocol used in the main experiments. We report both mean Intersection over Union (mIoU) and overall Intersection over Union (oIoU).

\subsubsection{Impact of SERA-Adapter and SERA-Fusion}

We first study the effect of SERA-Adapter and SERA-Fusion by enabling them incrementally on top of the baseline. The corresponding mIoU results on RefCOCO, RefCOCO+, and G-Ref are listed in Table~\ref{tab:ablation_components}. Adding SERA-Adapter alone improves performance on all three datasets, which indicates that backbone-level refinement is already beneficial for expression-conditioned segmentation. The gains are positive but modest. Once SERA-Fusion is introduced together with SERA-Adapter, the improvement becomes more pronounced across all benchmarks. This points to an additional benefit from expert-based spatial refinement before multimodal interaction. The best performance is achieved when both components are combined, suggesting that they provide complementary benefits. In particular, refining visual features at the fusion stage appears to provide a stronger representation for the subsequent backbone-level adaptation, leading to larger and more consistent gains.

\subsubsection{Effect of Top-$K$ Routing in SERA-Fusion}
\paragraph{Quantitative Analysis}
We next examine the role of routing sparsity by varying the Top-$K$ parameter in SERA-Fusion, with $K \in {1,2,3,4}$. For each configuration, the same value of $K$ is used during both training and inference so that the routing behavior remains consistent. This ablation focuses only on SERA-Fusion. SERA-Adapter does not introduce a sparsity hyperparameter because it relies on soft routing and keeps all experts active through continuous weighting. In contrast, SERA-Fusion uses sparse Top-$K$ routing and explicitly selects a subset of experts for each sample, making $K$ a direct control over routing capacity.

The RefCOCO results in Table~\ref{tab:topk_refcoco} show a clear pattern. When only a single expert is used ($K=1$), performance is lowest across all splits. Allowing more experts to participate improves both mIoU and oIoU. The gains, however, become smaller beyond $K=2$, suggesting that only a small number of experts is needed to capture most of the complementary refinement.

A similar trend appears on RefCOCO+, as reported in Table~\ref{tab:topk_refcocoplus}. Increasing the number of active experts improves performance compared to the $K=1$ setting, and the best validation and testA results occur at $K=4$. The testB split behaves slightly differently, where the differences between configurations are smaller and $K=2$ achieves the highest mIoU. Even so, the overall behavior remains similar: allowing multiple experts improves performance, while the differences among larger $K$ values are relatively small.

The G-Ref results in Table~\ref{tab:topk_gref} vary somewhat across evaluation splits. On the Google validation split, the highest mIoU is achieved with $K=2$, while the UMD validation and test splits perform best at $K=4$. This suggests that the benefit of additional routing capacity depends on the evaluation setting, with the UMD splits benefiting more from larger values of $K$.

Overall, increasing $K$ beyond 1 consistently improves performance, although the optimal value is not identical across all datasets and splits. Considering the full set of results, we use $K=4$ in the remaining experiments, as it provides the most stable trade-off between accuracy and computational cost.

\begin{table*}[b!]
\centering
\small
\setlength{\tabcolsep}{6pt}
\renewcommand{\arraystretch}{1.15}
\caption{ Ablation study of SERA components in terms of mIoU. Arrows indicate changes in performance relative to the baseline.}
\label{tab:ablation_components}
\begin{tabular}{l|ccc}
\toprule
\textbf{Method} & \textbf{RefCOCO} & \textbf{RefCOCO+} & \textbf{G-Ref(g)} \\
\midrule

(Baseline) 
& 74.90 & 68.70 & 65.10 \\

+ SERA-Adapter 
& 75.35\,\up{0.45} 
& 69.42\,\up{0.72} 
& 65.74\,\up{0.64} \\

\textbf{+ SERA-Adapter + SERA-Fusion (SERA)} 
& \textbf{76.50}\,\up{1.60} 
& \textbf{70.40}\,\up{1.70} 
& \textbf{66.62}\,\up{1.52} \\

\bottomrule
\end{tabular}
\end{table*}

\begin{table*}[!t]
\centering
\small
\setlength{\tabcolsep}{4pt}
\renewcommand{\arraystretch}{1.15}

\caption{ Top-K routing ablation on RefCOCO. Results are reported in terms of mIoU and oIoU on the validation, testA, and testB splits. Arrows indicate performance change relative to $K=1$.}
\label{tab:topk_refcoco}
\begin{tabular}{c|cc|cc|cc}
\toprule
\multirow{2}{*}{\textbf{Top-$K$}} 
& \multicolumn{2}{c|}{\textbf{val}} 
& \multicolumn{2}{c|}{\textbf{testA}} 
& \multicolumn{2}{c}{\textbf{testB}} \\
\cmidrule(lr){2-3}\cmidrule(lr){4-5}\cmidrule(lr){6-7}
& \textbf{mIoU} & \textbf{oIoU}
& \textbf{mIoU} & \textbf{oIoU}
& \textbf{mIoU} & \textbf{oIoU} \\
\midrule
$K=1$  & 75.46 & 73.32  & 77.72 & 76.31 & 72.80 & 69.15 \\
$K=2$  & 76.47\,\up{1.01} & 74.65\,\up{1.33} & 78.17\,\up{0.45} & 76.91\,\up{0.60} & 73.55\,\up{0.75} & 71.23\,\up{2.08} \\
$K=3$  & 76.20\,\up{0.74} & 74.10\,\up{0.78} & 78.10\,\up{0.38} & 77.09\,\up{0.78} & 73.73\,\up{0.93} & 71.13\,\up{1.98} \\
\textbf{$K=4$} & \textbf{76.50}\,\up{1.04} & \textbf{74.74}\,\up{1.42} & \textbf{78.20}\,\up{0.48} & \textbf{77.09}\,\up{0.78} & \textbf{73.73}\,\up{0.93} & \textbf{71.32}\,\up{2.17} \\
\bottomrule
\end{tabular}

\vspace{8pt}

\caption{ Top-K routing ablation on RefCOCO+. Results are reported in terms of mIoU and oIoU on the validation, testA, and testB splits. Arrows indicate performance change relative to $K=1$.}
\label{tab:topk_refcocoplus}
\begin{tabular}{c|cc|cc|cc}
\toprule
\multirow{2}{*}{\textbf{Top-$K$}} 
& \multicolumn{2}{c|}{\textbf{val}} 
& \multicolumn{2}{c|}{\textbf{testA}} 
& \multicolumn{2}{c}{\textbf{testB}} \\
\cmidrule(lr){2-3}\cmidrule(lr){4-5}\cmidrule(lr){6-7}
& \textbf{mIoU} & \textbf{oIoU}
& \textbf{mIoU} & \textbf{oIoU}
& \textbf{mIoU} & \textbf{oIoU} \\
\midrule
$K=1$  & 69.60 & 66.31  & 73.39 & 70.44 & 62.20 & 57.30 \\
$K=2$  & 70.16\,\up{0.56} & 66.64\,\up{0.33} & 74.16\,\up{0.77} & 70.80\,\up{0.36} & 63.77\,\up{1.57} & 58.92\,\up{1.62} \\
$K=3$  & 69.84\,\up{0.24} & 66.41\,\up{0.10} & 74.40\,\up{1.01} & 71.70\,\up{1.26} & 62.76\,\up{0.56} & 57.43\,\up{0.13} \\
\textbf{$K=4$} & \textbf{70.40}\,\up{0.80} & \textbf{66.91}\,\up{0.60} & \textbf{74.43}\,\up{1.04} & \textbf{72.28}\,\up{1.84} & \textbf{62.84}\,\up{0.64} & \textbf{58.20}\,\up{0.90} \\
\bottomrule
\end{tabular}

\vspace{8pt}

\caption{ Top-K routing ablation on RefCOCOg. Results are reported in terms of mIoU and oIoU on the validation, testA, and testB splits. Arrows indicate performance change relative to $K=1$.}
\label{tab:topk_gref}
\begin{tabular}{c|cc|cc|cc}
\toprule
\multirow{2}{*}{\textbf{Top-$K$}} 
& \multicolumn{2}{c|}{\textbf{val(g)}} 
& \multicolumn{2}{c|}{\textbf{val(u)}} 
& \multicolumn{2}{c}{\textbf{test(u)}} \\
\cmidrule(lr){2-3}\cmidrule(lr){4-5}\cmidrule(lr){6-7}
& \textbf{mIoU} & \textbf{oIoU}
& \textbf{mIoU} & \textbf{oIoU}
& \textbf{mIoU} & \textbf{oIoU} \\
\midrule
$K=1$  & 66.22 & 61.80 & 66.61 & 63.61 & 67.58 & 65.72 \\
$K=2$  & 67.09\,\up{0.87} & 63.84\,\up{2.04} & 68.24\,\up{1.63} & 64.91\,\up{1.30} & 68.58\,\up{1.00} & 66.26\,\up{0.54} \\
\textbf{$K=3$} & 67.00\,\up{0.78} & 63.35\,\up{1.55} & 66.95\,\up{0.34} & 64.91\,\up{1.30} & 67.55\,\down{0.03} & 65.64\,\down{0.08} \\
$K=4$  & \textbf{66.62}\,\up{0.40} & \textbf{63.30}\,\up{1.50} & \textbf{68.79}\,\up{2.18} & \textbf{65.58}\,\up{1.97} & \textbf{68.94}\,\up{1.36} & \textbf{66.34}\,\up{0.62} \\
\bottomrule
\end{tabular}

\end{table*}

\begin{figure}
\centering
\begin{subfigure}{\linewidth}
    \centering
    \includegraphics[width=\linewidth]{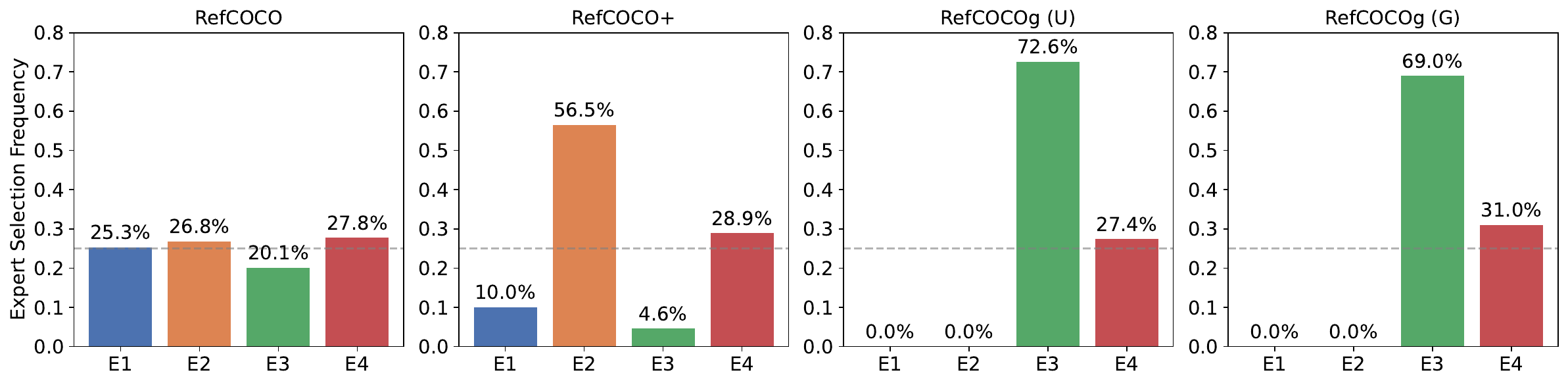}
    \caption{$K=1$ routing}
\end{subfigure}
\begin{subfigure}{\linewidth}
    \centering
    \includegraphics[width=\linewidth]{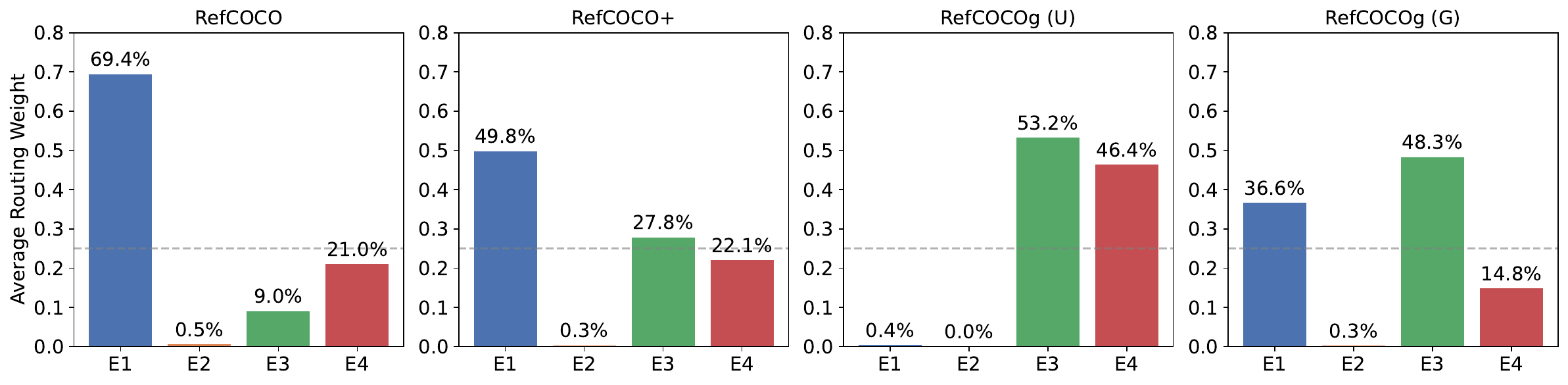}
    \caption{$K=2$ routing}
\end{subfigure}
\begin{subfigure}{\linewidth}
    \centering
    \includegraphics[width=\linewidth]{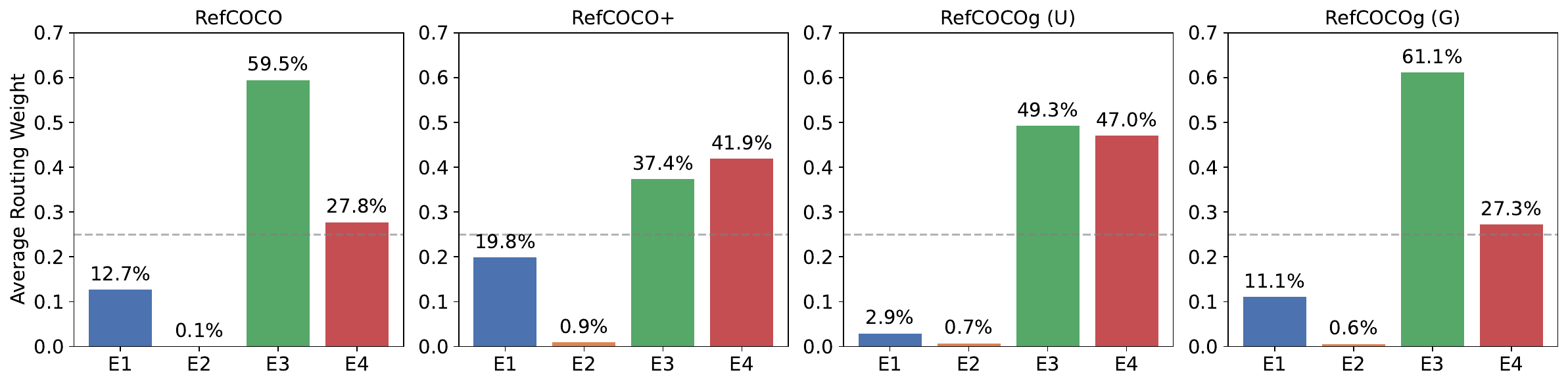}
    \caption{$K=3$ routing}
\end{subfigure}
\begin{subfigure}{\linewidth}
    \centering
    \includegraphics[width=\linewidth]{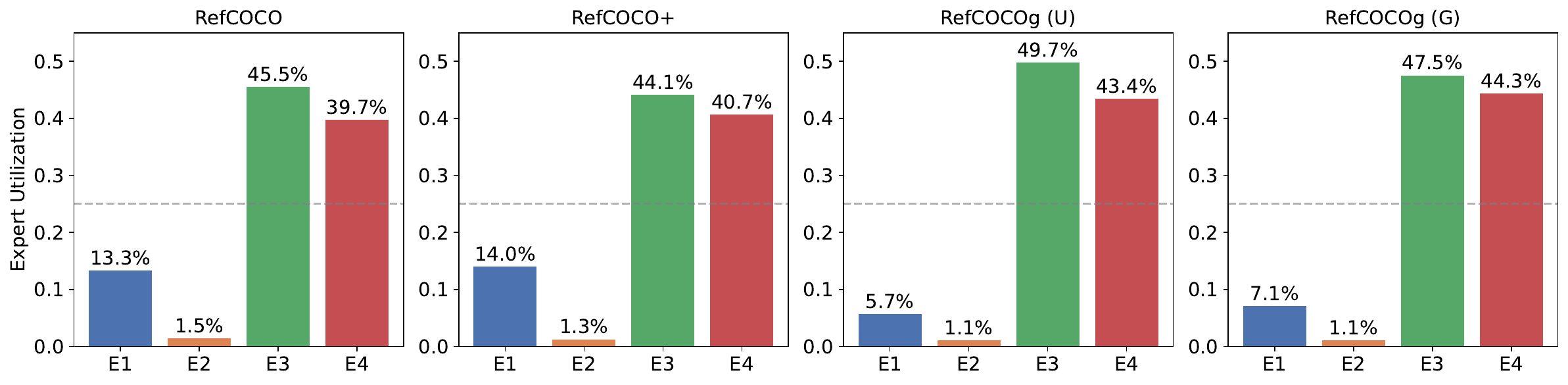}
    \caption{$K=4$ routing}
\end{subfigure}

\caption{ Average routing weight assigned to each expert in SERA-Fusion across routing configurations ($K=1, 2, 3, 4$) for RefCOCO, RefCOCO+, RefCOCOg (U), and RefCOCOg (G). Boundary and Shape experts consistently receive the highest routing mass, while the Context expert is minimally utilized, indicating stable expert specialization across datasets.}

\label{fig:sera_routing}
\end{figure}

\begin{figure}
\centering
\setlength{\tabcolsep}{2pt}
\renewcommand{\arraystretch}{1.0}
\newcommand{\imgh}{2.2cm}   
\begin{tabular}{c c c c c}
& \shortstack{\scriptsize third\\ from left}
& \shortstack{\scriptsize yogurt\\ on left}
& \shortstack{\scriptsize guy in the back\\ with arm\\ on chair}
& \shortstack{\scriptsize man not\\ in purple} \\
\scriptsize Image &
\includegraphics[height=\imgh]{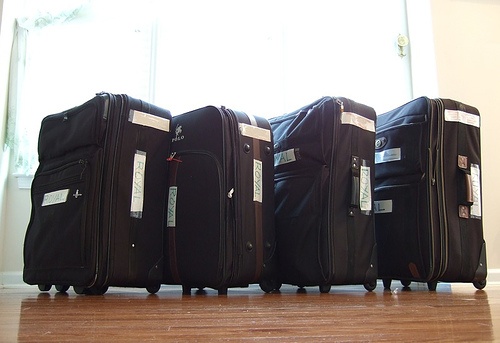} &
\includegraphics[height=\imgh]{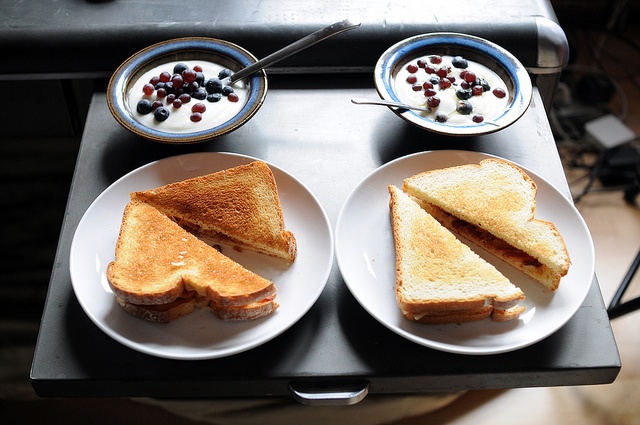} &
\includegraphics[height=\imgh]{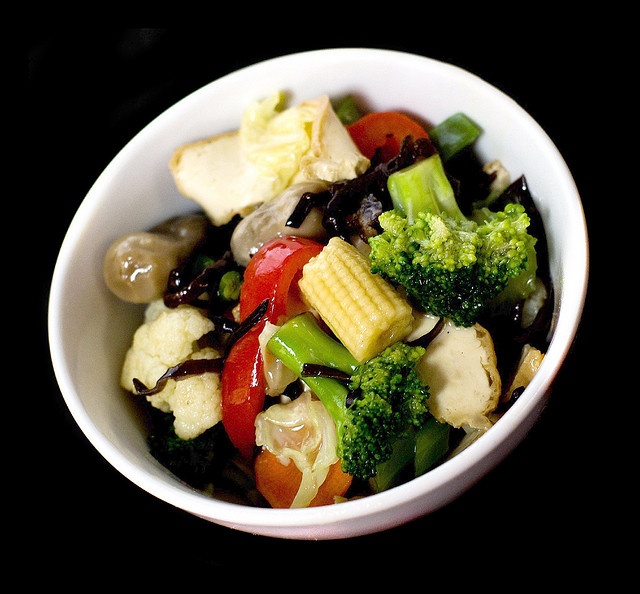} &
\includegraphics[height=\imgh]{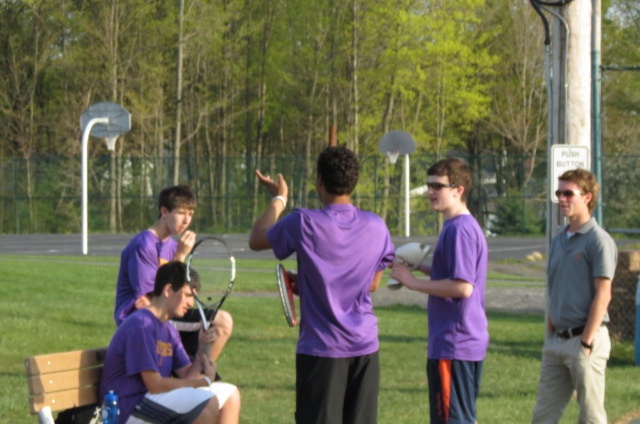} \\
\scriptsize GT &
\includegraphics[height=\imgh,trim=12 12 12 12,clip]{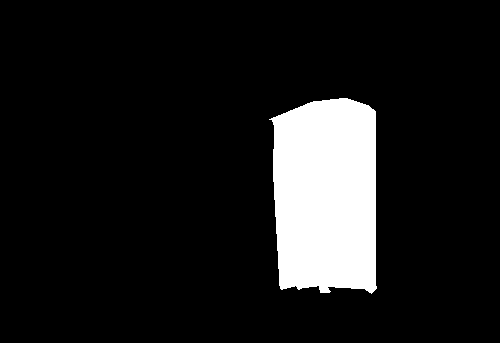} &
\includegraphics[height=\imgh,trim=12 12 12 12,clip]{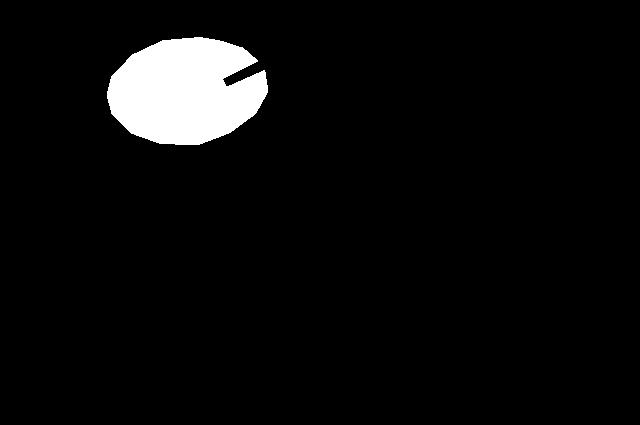} &
\includegraphics[height=\imgh,trim=12 12 12 12,clip]{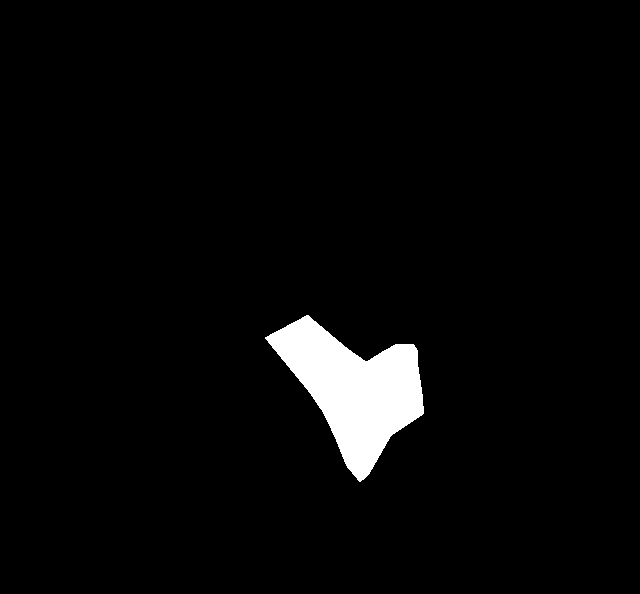} &
\includegraphics[height=\imgh,trim=12 12 12 12,clip]{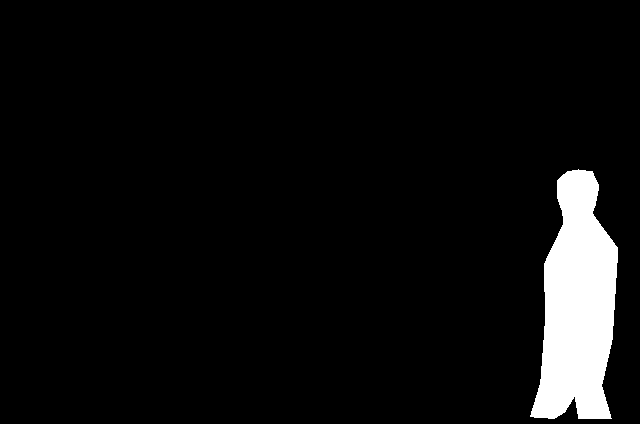} \\
\rotatebox{90}{\scriptsize Base} &
\includegraphics[height=\imgh]{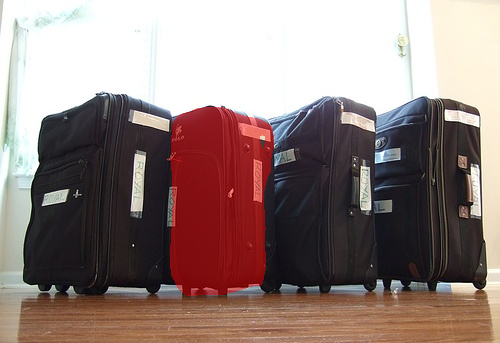} &
\includegraphics[height=\imgh]{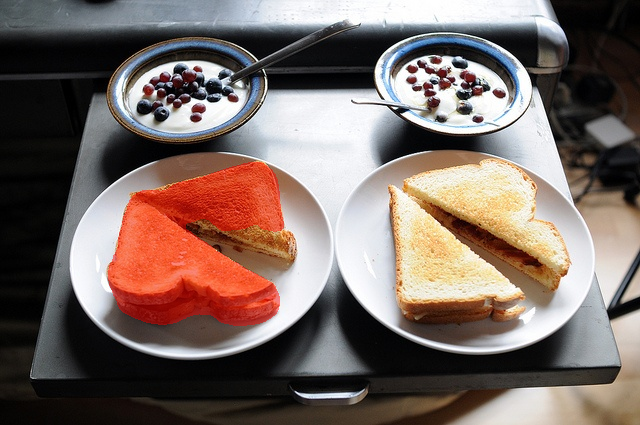} &
\includegraphics[height=\imgh]{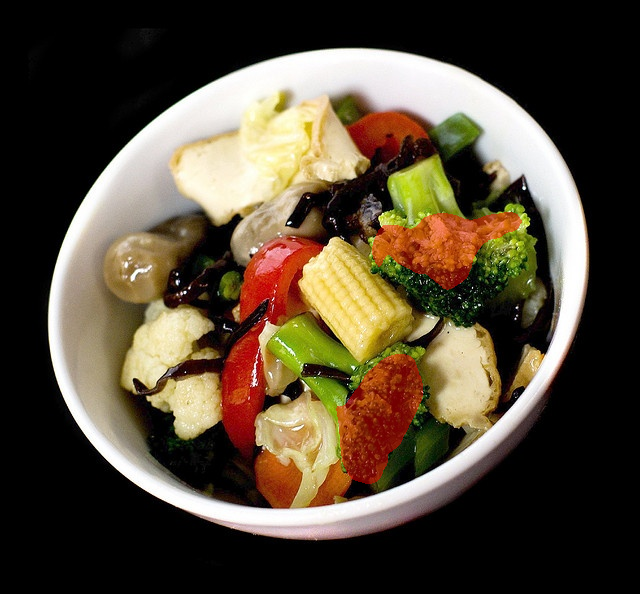} &
\includegraphics[height=\imgh]{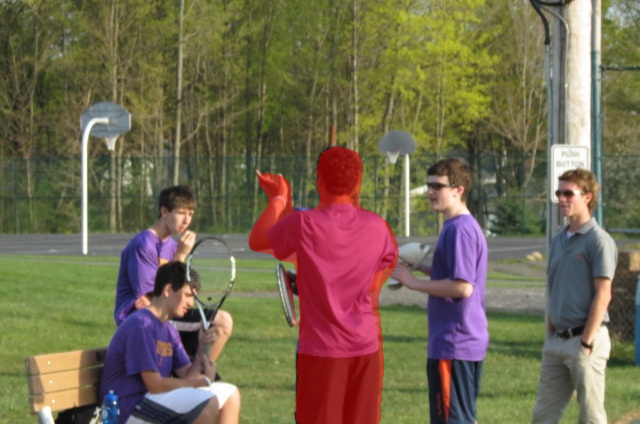} \\
\rotatebox{90}{\scriptsize $K=1$} &
\includegraphics[height=\imgh]{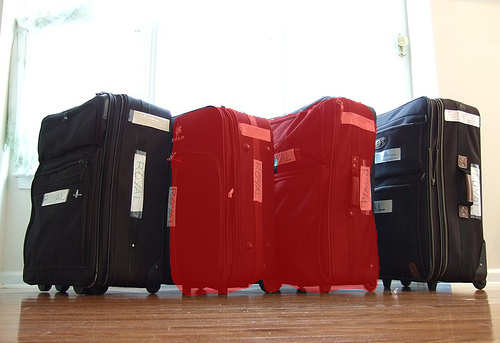} &
\includegraphics[height=\imgh]{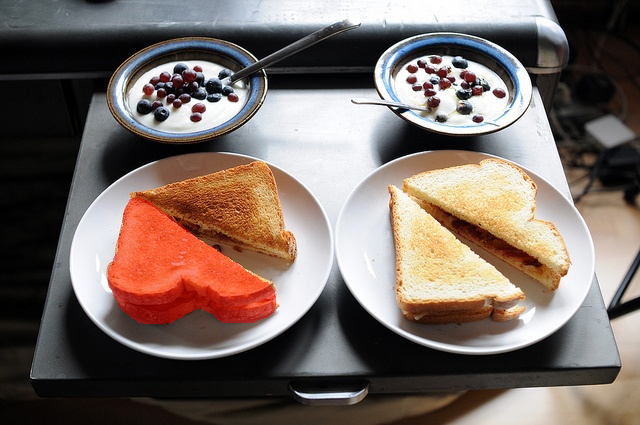} &
\includegraphics[height=\imgh]{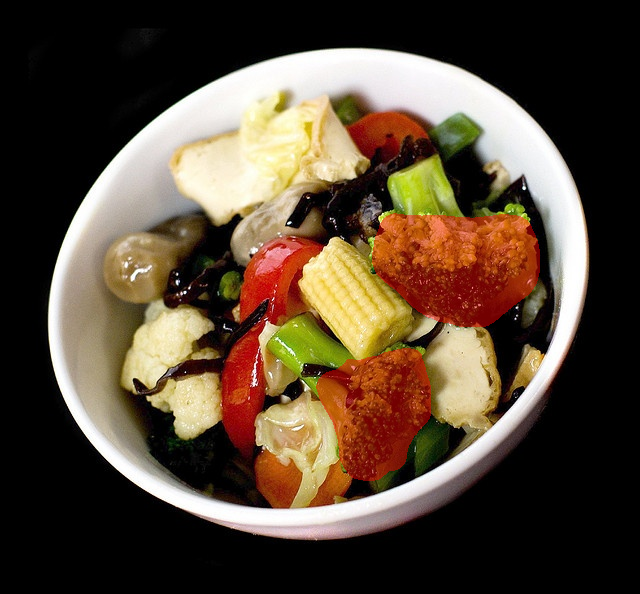} &
\includegraphics[height=\imgh]{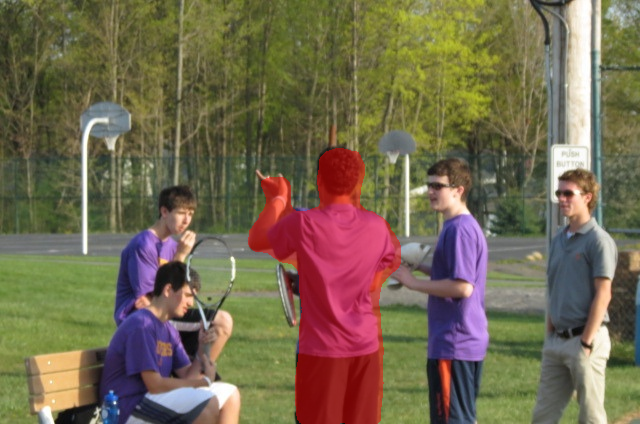} \\
\rotatebox{90}{\scriptsize $K=2$} &
\includegraphics[height=\imgh]{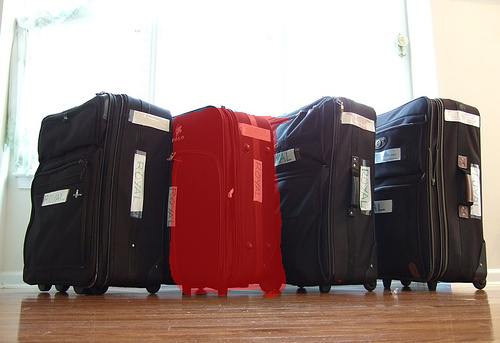} &
\includegraphics[height=\imgh]{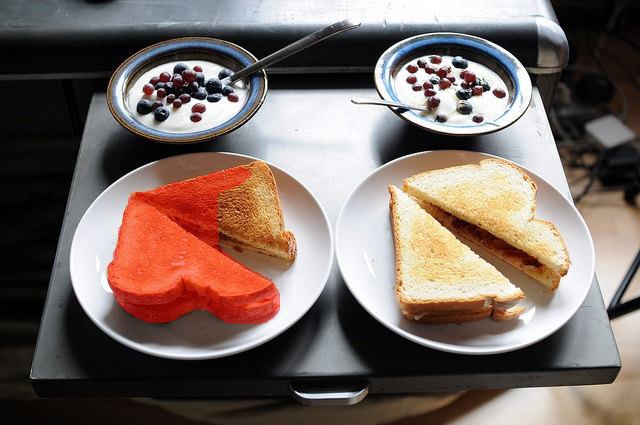} &
\includegraphics[height=\imgh]{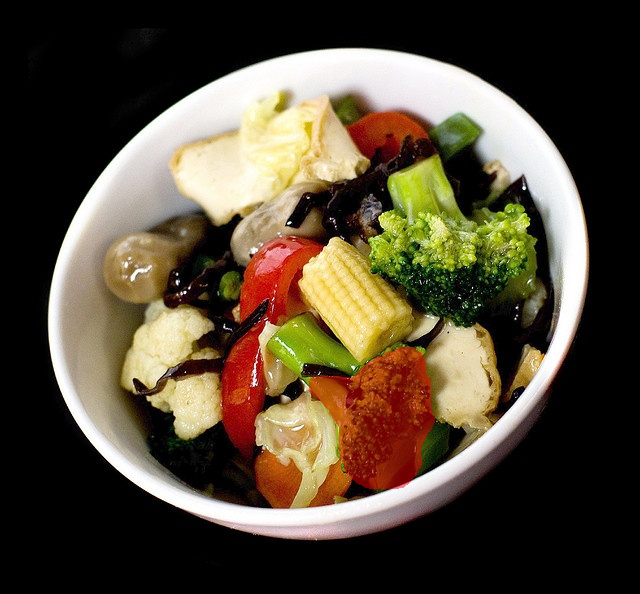} &
\includegraphics[height=\imgh]{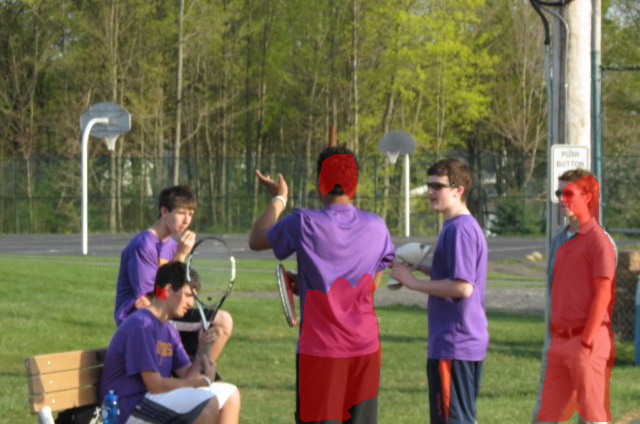} \\
\rotatebox{90}{\scriptsize $K=3$} &
\includegraphics[height=\imgh]{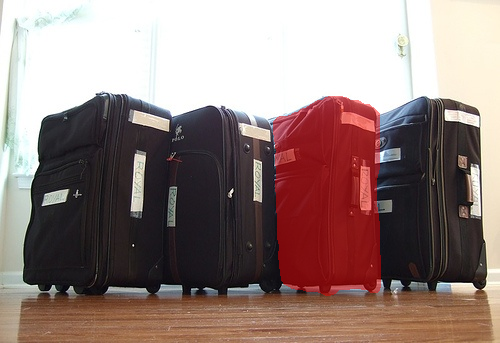} &
\includegraphics[height=\imgh]{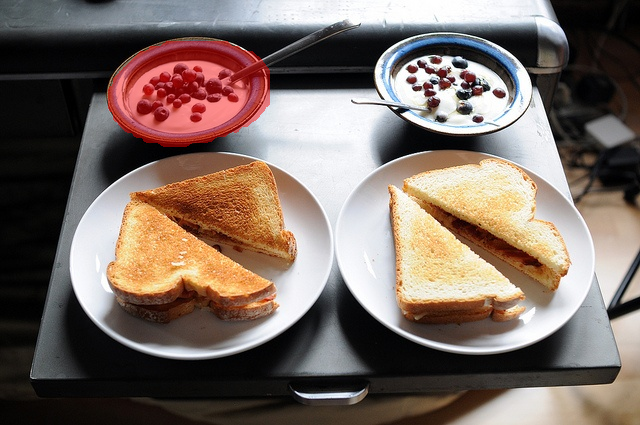} &
\includegraphics[height=\imgh]{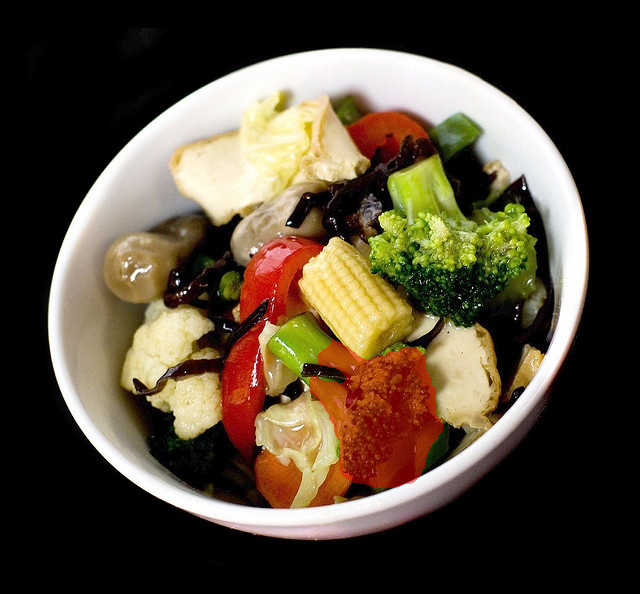} &
\includegraphics[height=\imgh]{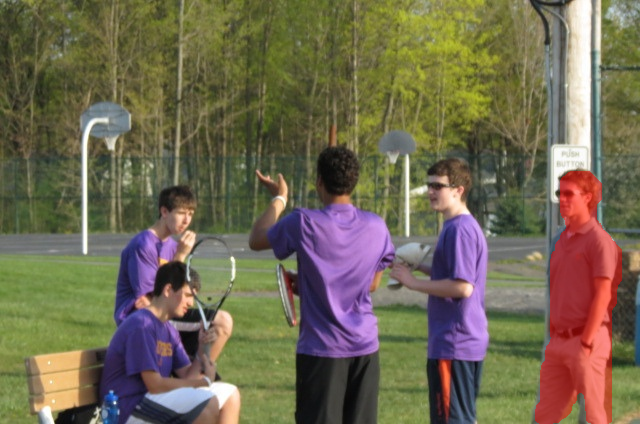} \\
\rotatebox{90}{\scriptsize $K=4$} &
\includegraphics[height=\imgh]{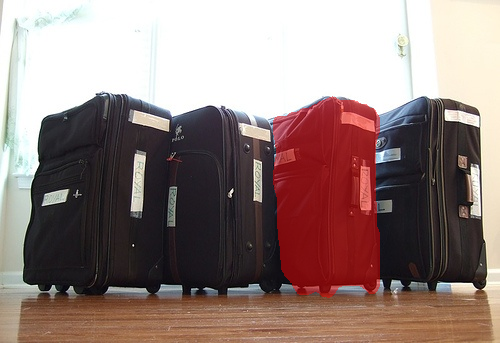} &
\includegraphics[height=\imgh]{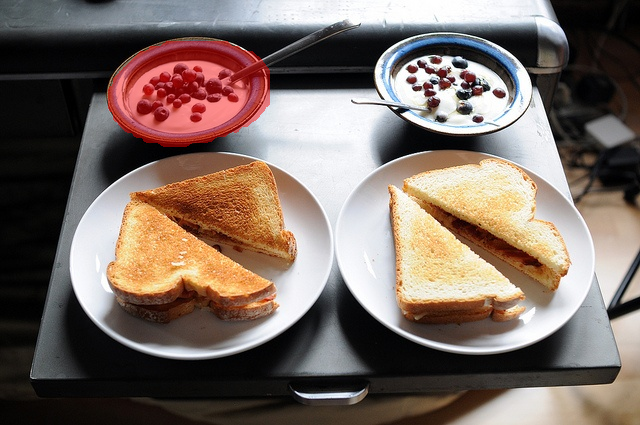} &
\includegraphics[height=\imgh]{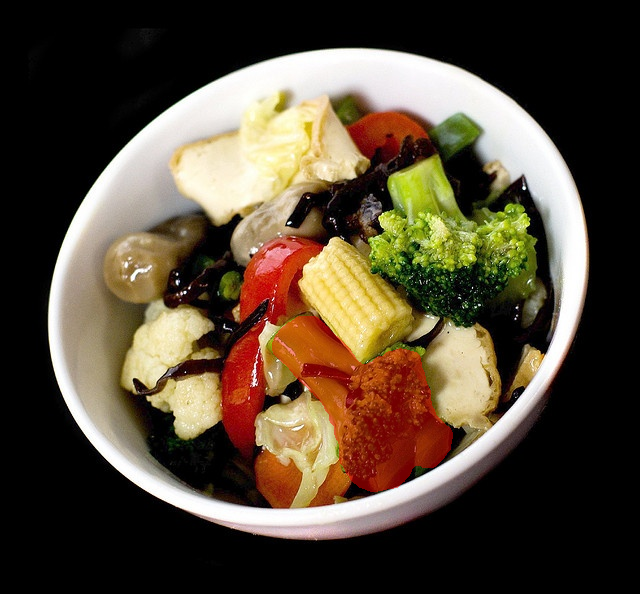} &
\includegraphics[height=\imgh]{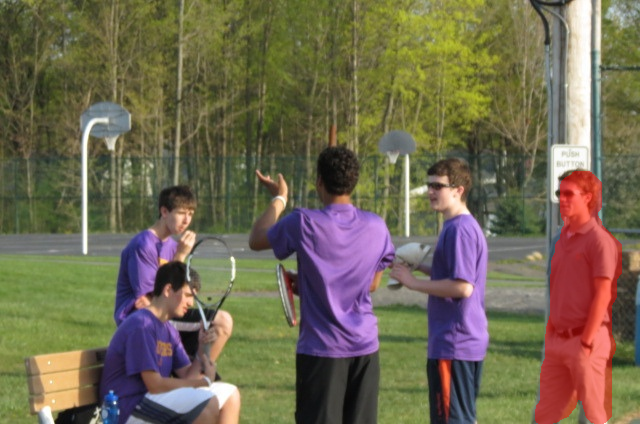} \\
\end{tabular}
\caption{  Qualitative comparison between the base model and SERA under different Top-$K$ expert routing settings. Columns correspond to different referring expressions, while rows illustrate the effect of increasing expert capacity. Compared with the base model, SERA produces more complete masks, sharper boundaries, and better foreground-background separation as $K$ increases.}
\label{fig:qual_topk}
\end{figure}

\paragraph{Qualitative Analysis.}
We examine qualitative examples to understand how Top-$K$ routing influences the behavior of SERA-Fusion as more experts participate in the prediction. Unlike SERA-Adapter, which uses soft routing within frozen backbone layers, SERA-Fusion performs explicit Top-$K$ selection, allowing direct control over how many experts contribute while keeping the routing input-dependent. Representative examples from RefCOCO are shown in Figure~\ref{fig:qual_topk}. When $K=1$, predictions rely on a single expert. This can produce reasonable masks in simple cases, but it often struggles when errors arise from multiple factors such as weak boundaries, distracting regions, or incomplete object structure. Increasing the routing capacity improves prediction quality. With $K=2$, masks generally become cleaner and better localized, suggesting that combining experts provides more balanced refinement. At $K=3$, the model becomes more reliable in complex scenes, particularly when several similar objects appear together or when contextual cues are needed to identify the correct referent. The strongest qualitative results are typically observed at $K=4$, where the predicted masks appear more complete and spatially coherent with fewer boundary artifacts. Overall, these observations are consistent with the quantitative results: combining multiple experts provides complementary refinements and leads to more reliable segmentation.

\section{Expert Routing Analysis}
\label{sec4}
In order to provide better insight into how well the SERA-Fusion architecture works, we analyze in Figure~\ref{fig:sera_routing} the expert routing distributions learned by the Top-$K$ router on four different routing settings, $K=1, 2, 3, 4$. For each expert level, statistics are computed on the validation set at the expert level. When $K=1$, the reported values are frequencies of selected experts. For $K> 1$, we show the average routing weight for each expert. For example, for $4$ experts, a routing weight of $0.25$ per expert corresponds to uniform routing.

These results show that expert usage is not uniform and differs across datasets. 
For RefCOCO with $K=1$, the selection frequencies are relatively balanced, 
indicating that no single expert consistently dominates. In contrast, RefCOCO+ 
exhibits a more uneven distribution, while RefCOCOg places stronger emphasis on 
two experts. This difference likely reflects the nature of the referring 
expressions in each dataset. RefCOCOg contains longer and more relational 
descriptions, which may encourage the router to rely on a smaller set of 
specialized experts, whereas RefCOCO+ generally involves shorter expressions.

As $K$ increases, routing becomes less sparse since multiple experts can 
contribute to each prediction. However, the distributions remain structured 
rather than uniform. At $K=2$, most of the routing weight is still concentrated 
on two experts, indicating a continued preference for a limited set of 
specialists. Even at larger values ($K=3$ and $K=4$), the distribution does not 
approach uniformity. One expert remains consistently active across datasets, 
while the remaining contributions are shared among a small subset of experts. 
These observations indicate that expert specialization is largely preserved 
even when more experts are allowed to participate.

This is particularly true in the case of RefCOCOg, in which the same two experts consistently rank as the top two across all $K$ values, from sparse top-1 selection to denser top-4 routing. This is unlikely to be due to chance. Rather, it suggests that there is some form of stable specialization. The case of RefCOCO+ is similar, yet distinct. The routing is heavily concentrated at $K=1$, yet as larger $K$ values are added, it becomes more collaborative. This suggests that there is some form of selectivity, yet flexibility in the routing.

In summary, this analysis of expert routing demonstrates that SERA-Fusion can learn expert specialization. There are three key takeaways. First, the routing changes depending on the dataset. Second, expert collaboration naturally occurs when multiple contributing branches are present. Third, expert preference is stable even when there is non-sparse routing. One might also notice that expert usage is highly unbalanced, with one branch underutilized in most cases. This is not necessarily a negative, however. Rather, it is likely that some experts are simply better than others at contributing to RIS, given its strong visual and linguistic requirements.

\section{Zero-Shot Cross-Dataset Generalization}
\label{sec5}

We also examine the generalization capabilities of our model by performing a series of qualitative experiments across the RefCOCO dataset family. For these experiments, our model is trained only on one dataset, yet evaluated directly on the remaining datasets without fine-tuning. For visualization, we overlay our predicted segmentation masks onto the images. This allows us to visually inspect how well our model is able to maintain its vision-language alignment when moving across different datasets, as well as changes in expression, scene, and annotation protocols.

\begin{figure}
\centering
\setlength{\tabcolsep}{2pt}
\renewcommand{\arraystretch}{1.05}
\begin{tabular}{c c c c c c c c}
 & \hyperlink{p28}{28}
 & \hyperlink{p29}{29}
 & \hyperlink{p30}{30}
 & \hyperlink{p31}{31}
 & \hyperlink{p32}{32}
 & \hyperlink{p33}{33}
 & \hyperlink{p34}{34} \\
\rotatebox{90}{RefCOCOg} &
\includegraphics[width=0.13\textwidth,height=1.86cm]{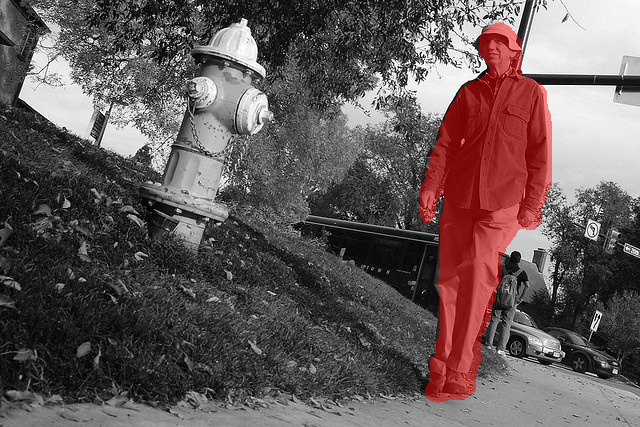} &
\includegraphics[width=0.13\textwidth,height=1.86cm]{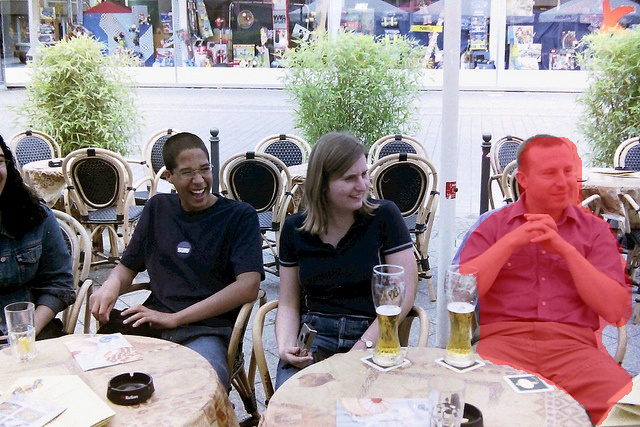} &
\includegraphics[width=0.13\textwidth,height=1.86cm]{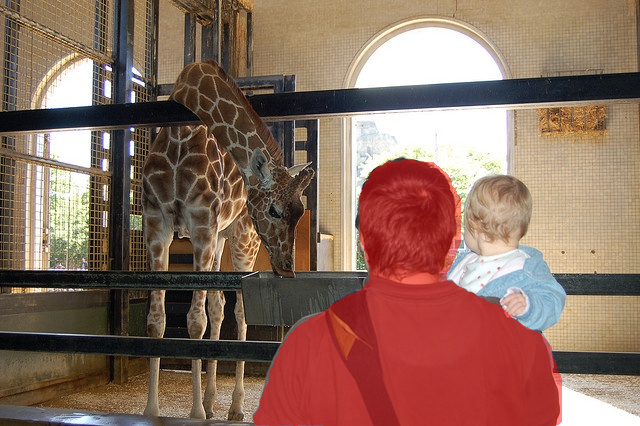} &
\includegraphics[width=0.13\textwidth,height=1.86cm]{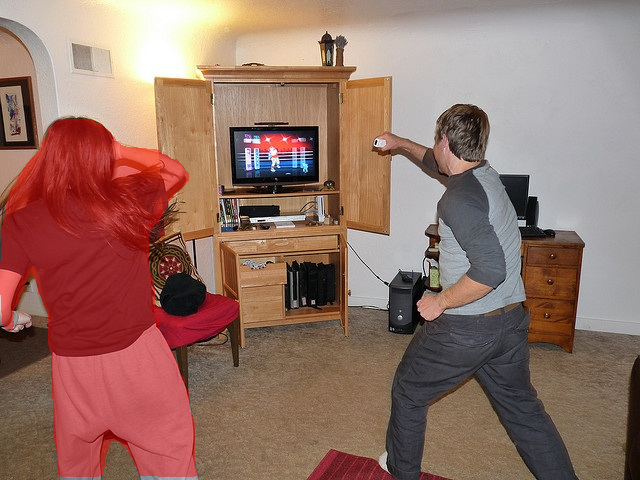} &
\includegraphics[width=0.13\textwidth,height=1.86cm]{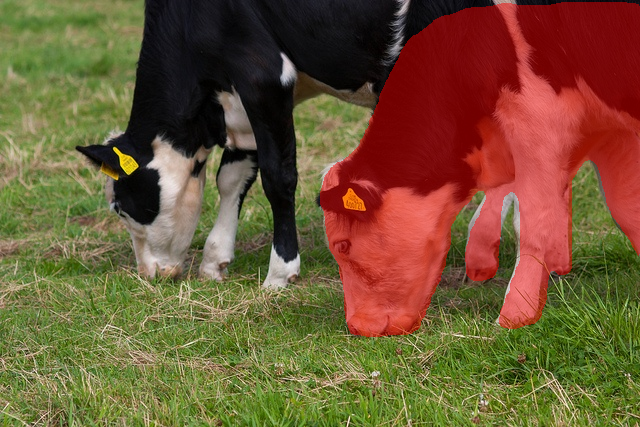} &
\includegraphics[width=0.13\textwidth,height=1.86cm]{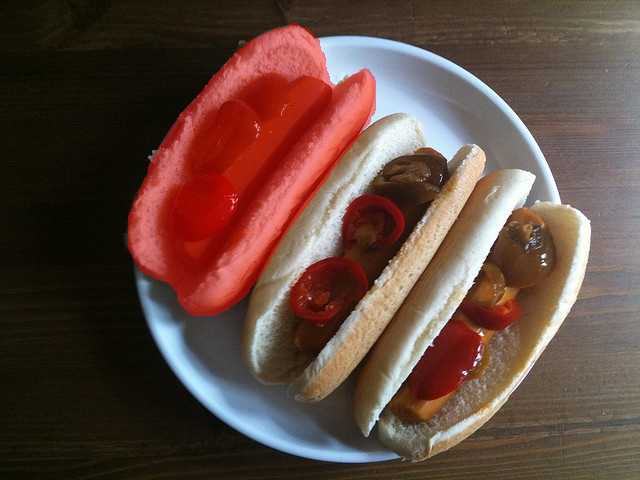} &
\includegraphics[width=0.13\textwidth,height=1.86cm]{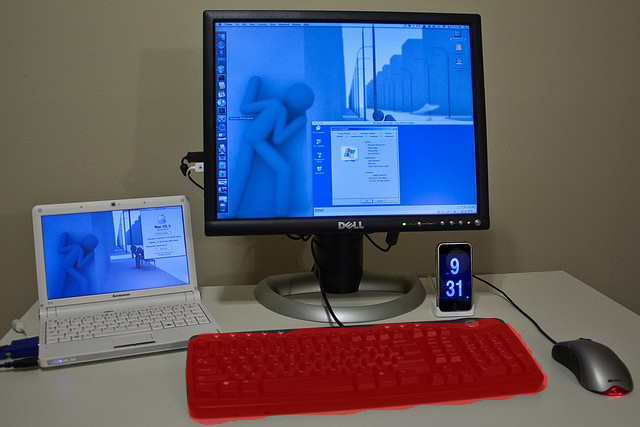} \\
 & \scriptsize 87.50
 & \scriptsize 88.47
 & \scriptsize 90.20
 & \scriptsize 90.67
 & \scriptsize 92.04
 & \scriptsize 92.09
 & \scriptsize 91.02 \\[2mm]
 & \hyperlink{p35}{35}
 & \hyperlink{p36}{36}
 & \hyperlink{p37}{37}
 & \hyperlink{p38}{38}
 & \hyperlink{p39}{39}
 & \hyperlink{p40}{40}
 & \hyperlink{p41}{41} \\
\rotatebox{90}{RefCOCO+} &
\includegraphics[width=0.13\textwidth,height=1.86cm]{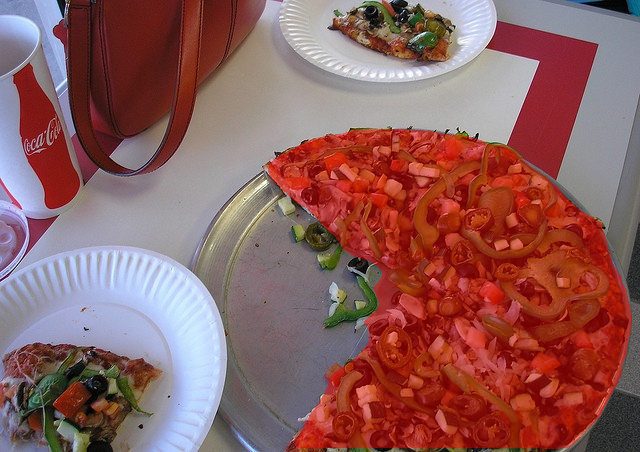} &
\includegraphics[width=0.13\textwidth,height=1.86cm]{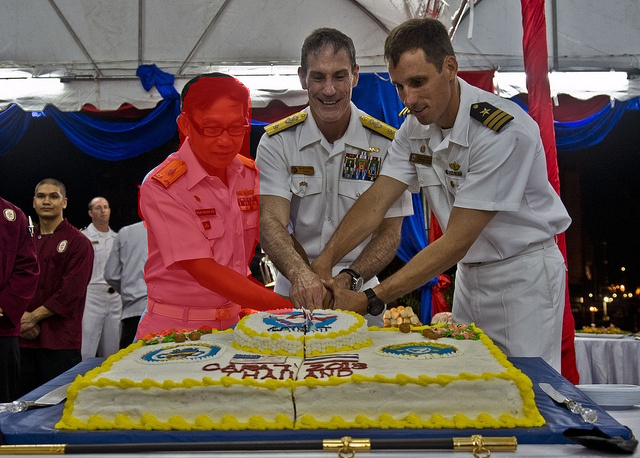} &
\includegraphics[width=0.13\textwidth,height=1.86cm]{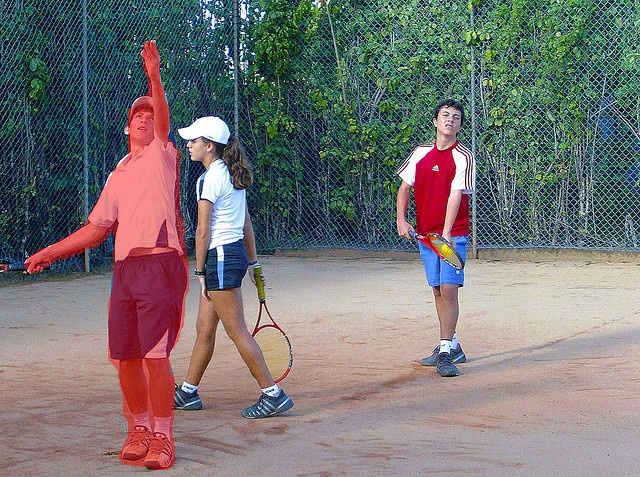} &
\includegraphics[width=0.13\textwidth,height=1.86cm]{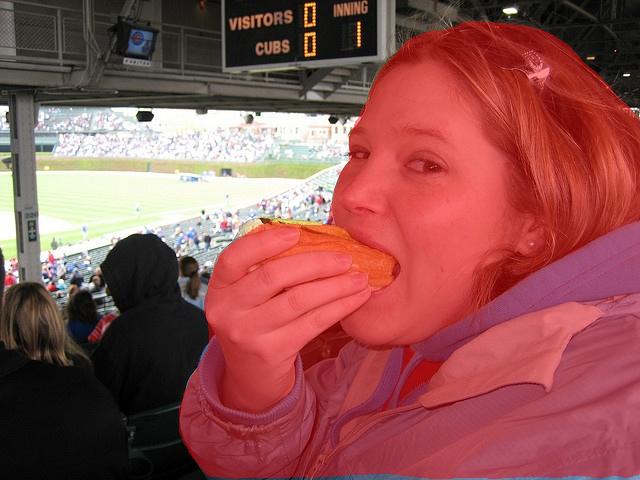} &
\includegraphics[width=0.13\textwidth,height=1.86cm]{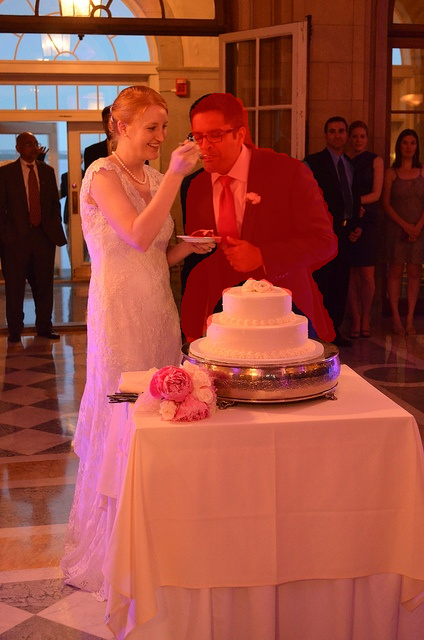} &
\includegraphics[width=0.13\textwidth,height=1.86cm]{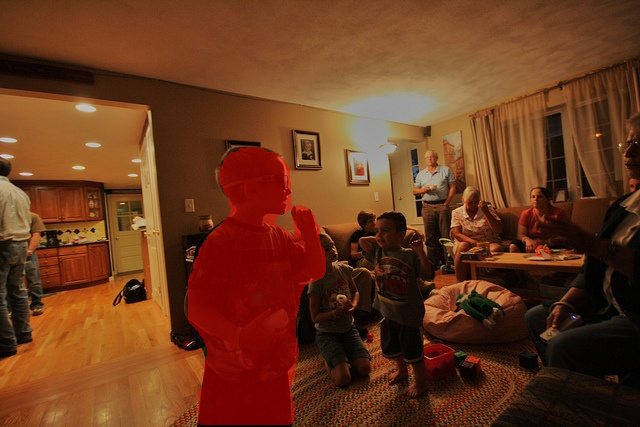} &
\includegraphics[width=0.13\textwidth,height=1.86cm]{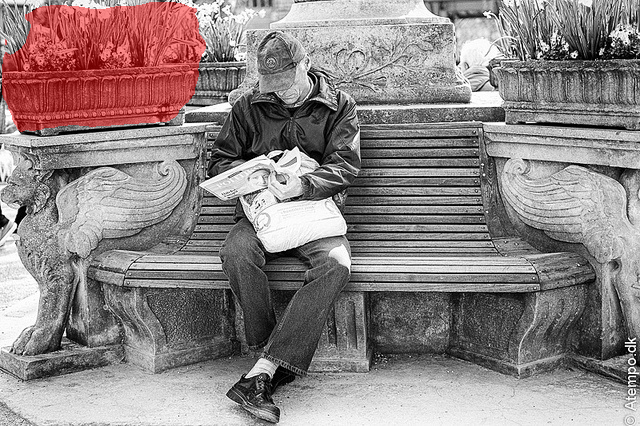} \\
 & \scriptsize 92.05
 & \scriptsize 91.13
 & \scriptsize 89.50
 & \scriptsize 96.29
 & \scriptsize 84.88
 & \scriptsize 91.07
 & \scriptsize 83.58 \\
\end{tabular}
\caption{ Zero-shot qualitative overlay results on RefCOCOg and RefCOCO+. Numbers above images indicate clickable sample IDs referenced in Appendix~\ref{sec:appendix}. Values below each image denote IoU scores.}
\label{fig:qual4}
\end{figure}

Figure~\ref{fig:qual4} shows an example of the transfer setting where the model is fine-tuned on RefCOCO and evaluated on RefCOCO+ and RefCOCOg. Although these datasets differ in their expression styles and annotation protocols, the model still produces reasonable segmentation masks that align well with the referring expressions. The predicted regions remain well localized despite the dataset shift, suggesting that the model does not rely heavily on dataset-specific cues. Instead, it appears to learn cross-modal representations that generalize across datasets.

\begin{figure}
\centering
\setlength{\tabcolsep}{2pt}
\renewcommand{\arraystretch}{1.05}
\begin{tabular}{c c c c c c c c}
 & \hyperlink{p42}{42}
 & \hyperlink{p43}{43}
 & \hyperlink{p44}{44}
 & \hyperlink{p45}{45}
 & \hyperlink{p46}{46}
 & \hyperlink{p47}{47}
 & \hyperlink{p48}{48} \\
\rotatebox{90}{RefCOCO} &
\includegraphics[width=0.13\textwidth,height=1.86cm]{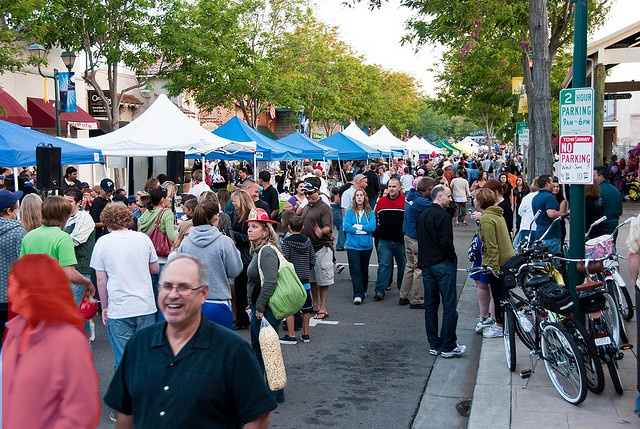} &
\includegraphics[width=0.13\textwidth,height=1.86cm]{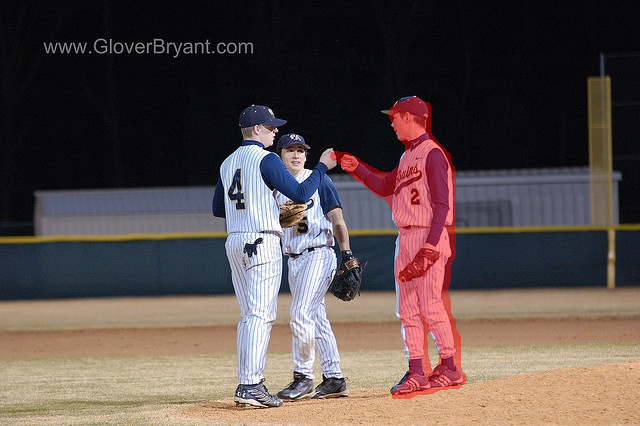} &
\includegraphics[width=0.13\textwidth,height=1.86cm]{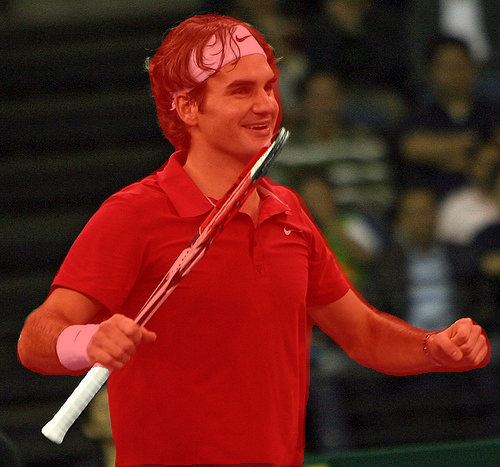} &
\includegraphics[width=0.13\textwidth,height=1.86cm]{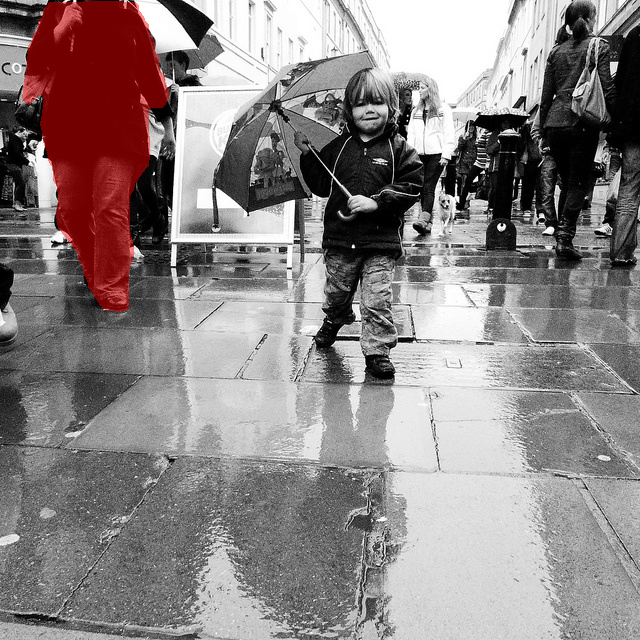} &
\includegraphics[width=0.13\textwidth,height=1.86cm]{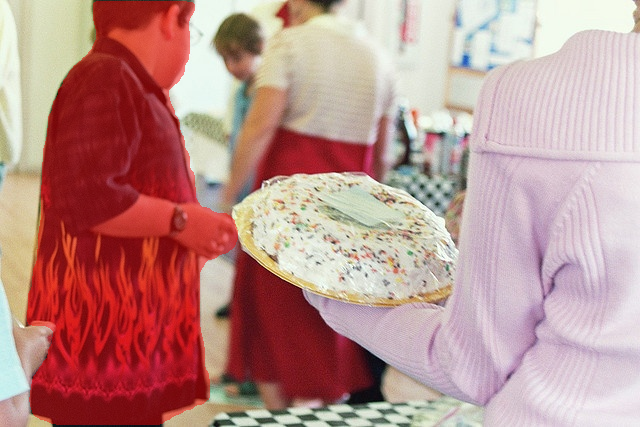} &
\includegraphics[width=0.13\textwidth,height=1.86cm]{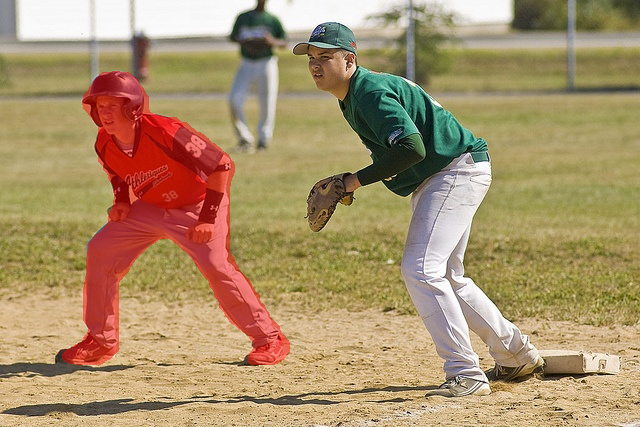} &
\includegraphics[width=0.13\textwidth,height=1.86cm]{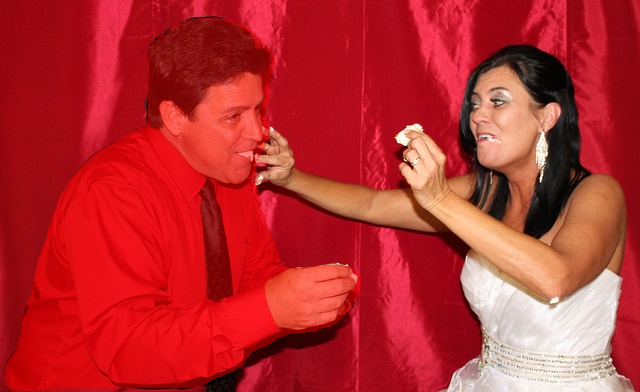} \\
 & \scriptsize 95.15
 & \scriptsize 83.14
 & \scriptsize 89.04
 & \scriptsize 89.85
 & \scriptsize 95.28
 & \scriptsize 91.12
 & \scriptsize 93.02 \\[2mm]
 & \hyperlink{p49}{49}
 & \hyperlink{p50}{50}
 & \hyperlink{p51}{51}
 & \hyperlink{p52}{52}
 & \hyperlink{p53}{53}
 & \hyperlink{p54}{54}
 & \hyperlink{p55}{55} \\
\rotatebox{90}{RefCOCOg} &
\includegraphics[width=0.13\textwidth,height=1.86cm]{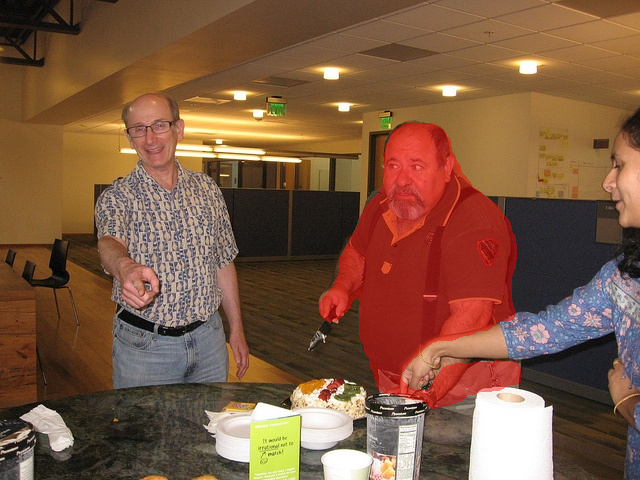} &
\includegraphics[width=0.13\textwidth,height=1.86cm]{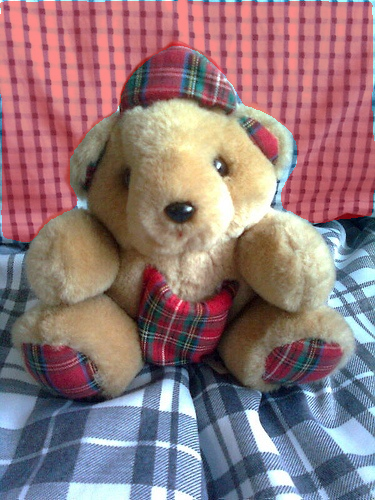} &
\includegraphics[width=0.13\textwidth,height=1.86cm]{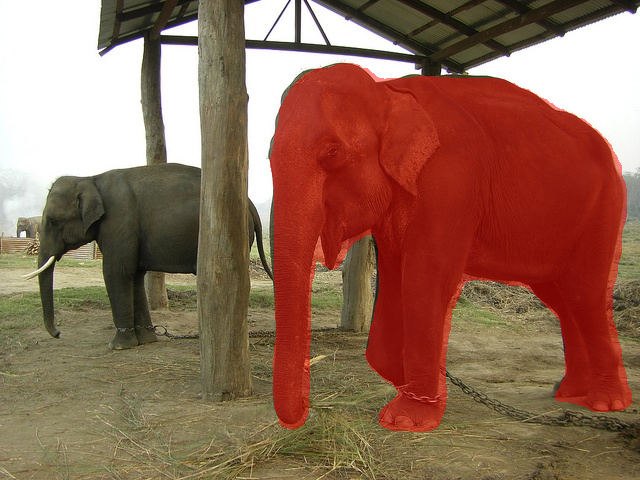} &
\includegraphics[width=0.13\textwidth,height=1.86cm]{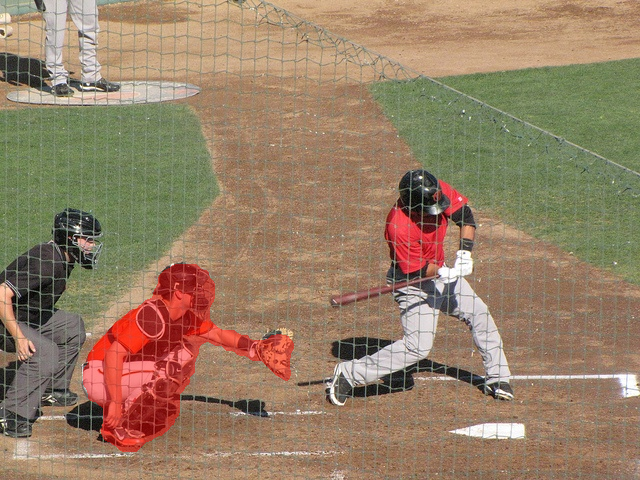} &
\includegraphics[width=0.13\textwidth,height=1.86cm]{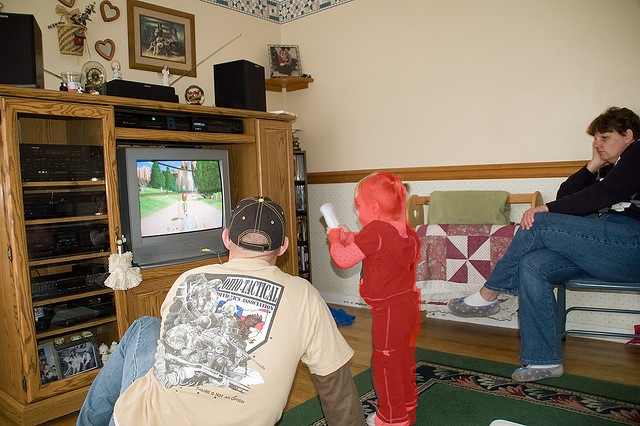} &
\includegraphics[width=0.13\textwidth,height=1.86cm]{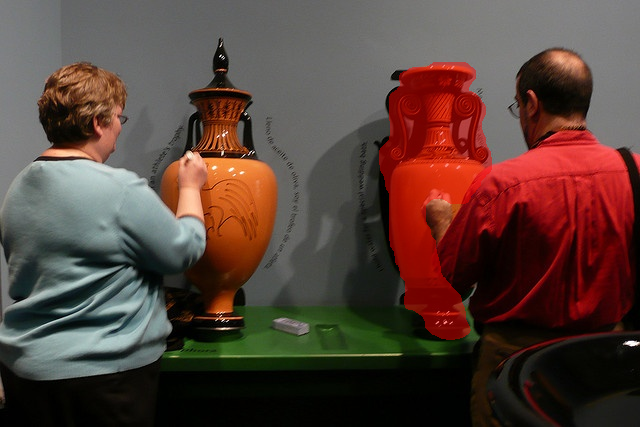} &
\includegraphics[width=0.13\textwidth,height=1.86cm]{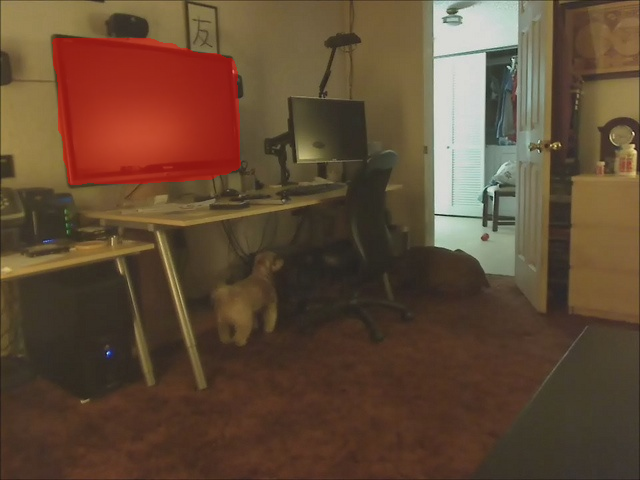} \\
 & \scriptsize 87.17
 & \scriptsize 93.20
 & \scriptsize 84.11
 & \scriptsize 92.51
 & \scriptsize 94.35
 & \scriptsize 87.85
 & \scriptsize 86.49 \\
\end{tabular}
\caption{ Zero-shot qualitative overlay results on RefCOCO and RefCOCOg. Numbers above images indicate clickable sample IDs referenced in Appendix~\ref{sec:appendix}. Values below each image denote IoU scores.}
\label{fig:qual5}
\end{figure}

Figure~\ref{fig:qual5} presents another transfer scenario where the model is trained on RefCOCO+ and tested on RefCOCO and RefCOCOg. Even when the model is not explicitly trained on location-based expressions, it is still able to localize the target objects with reasonable reliability. The predicted segmentation masks also remain stable in cluttered scenes and in situations where objects have more complex relationships with each other. This observation suggests that the model relies mainly on visual appearance cues learned during training, rather than depending only on spatial language information.

\begin{figure}
\centering
\setlength{\tabcolsep}{2pt}
\renewcommand{\arraystretch}{1.05}
\begin{tabular}{c c c c c c c c}
 & \hyperlink{p56}{56}
 & \hyperlink{p57}{57}
 & \hyperlink{p58}{58}
 & \hyperlink{p59}{59}
 & \hyperlink{p60}{60}
 & \hyperlink{p61}{61}
 & \hyperlink{p62}{62} \\
\rotatebox{90}{RefCOCO} &
\includegraphics[width=0.13\textwidth,height=1.86cm]{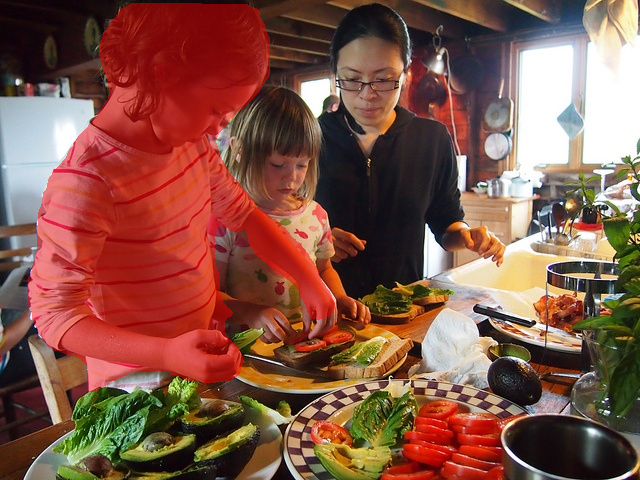} &
\includegraphics[width=0.13\textwidth,height=1.86cm]{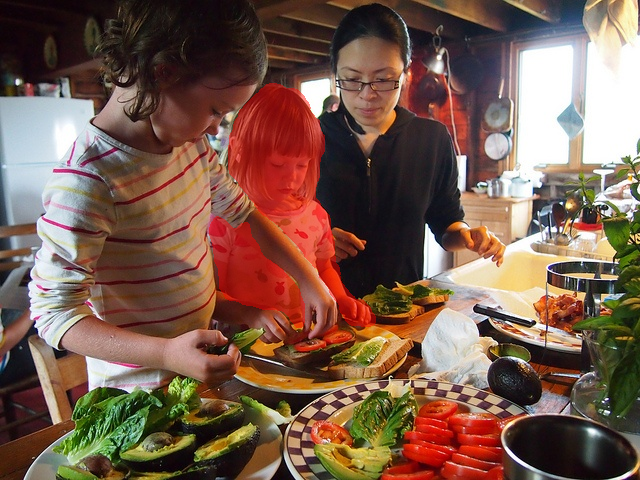} &
\includegraphics[width=0.13\textwidth,height=1.86cm]{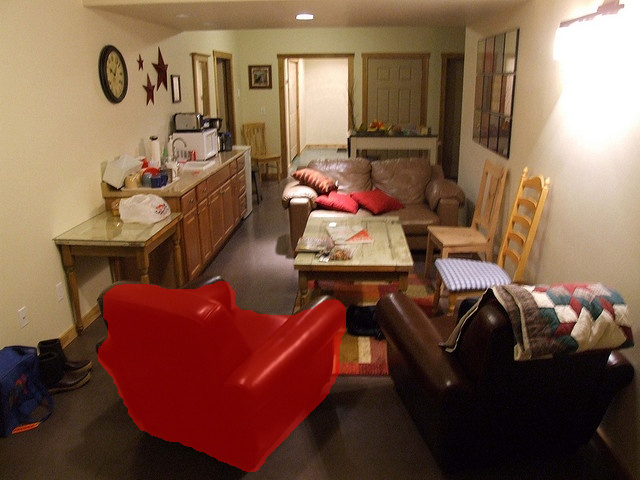} &
\includegraphics[width=0.13\textwidth,height=1.86cm]{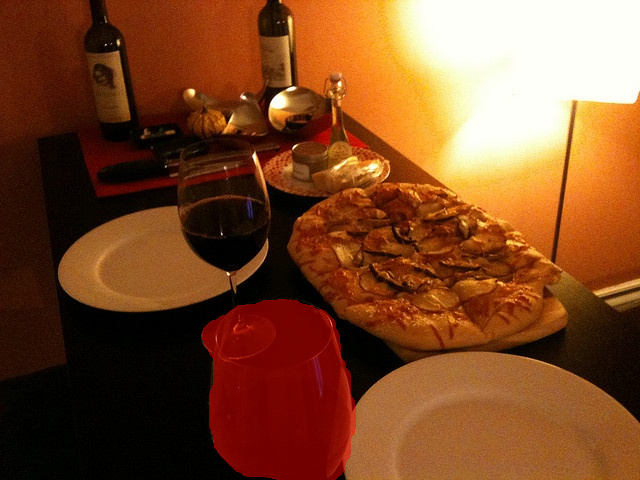} &
\includegraphics[width=0.13\textwidth,height=1.86cm]{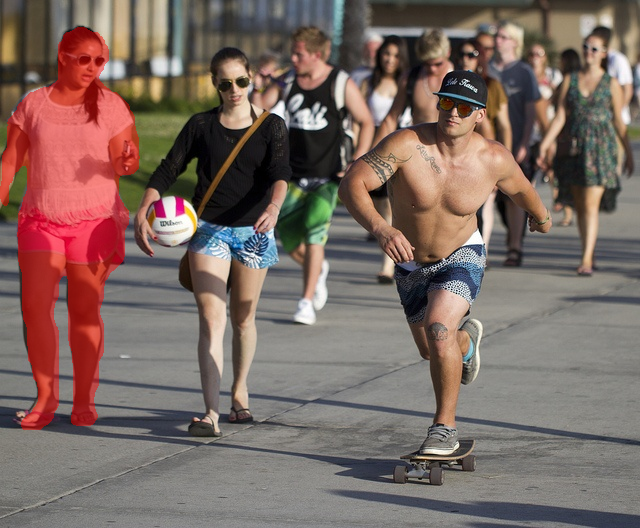} &
\includegraphics[width=0.13\textwidth,height=1.86cm]{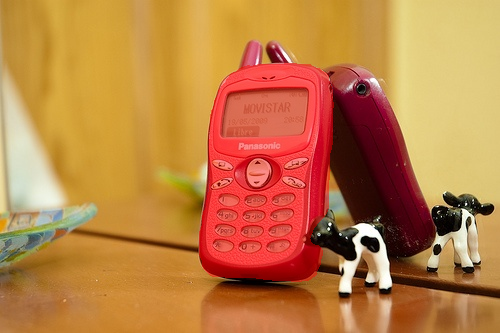} &
\includegraphics[width=0.13\textwidth,height=1.86cm]{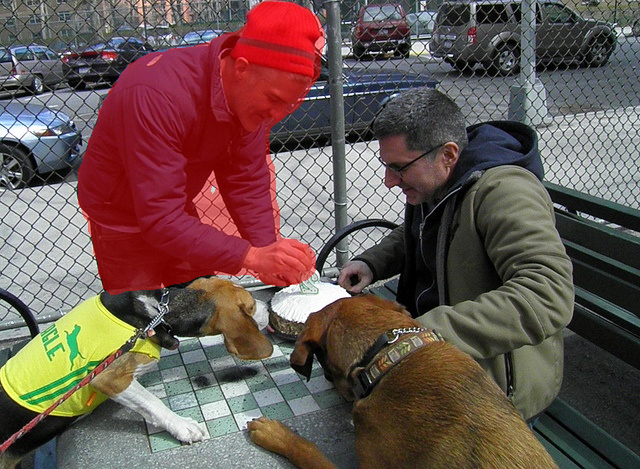} \\
 & \scriptsize 90.57
 & \scriptsize 81.21
 & \scriptsize 91.13
 & \scriptsize 91.23
 & \scriptsize 91.21
 & \scriptsize 93.92
 & \scriptsize 94.60 \\[2mm]
 & \hyperlink{p63}{63}
 & \hyperlink{p64}{64}
 & \hyperlink{p65}{65}
 & \hyperlink{p66}{66}
 & \hyperlink{p67}{67}
 & \hyperlink{p68}{68}
 & \hyperlink{p69}{69} \\
\rotatebox{90}{RefCOCO+} &
\includegraphics[width=0.13\textwidth,height=1.86cm]{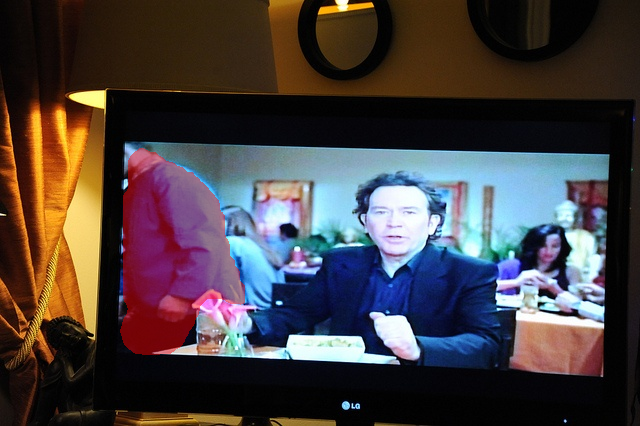} &
\includegraphics[width=0.13\textwidth,height=1.86cm]{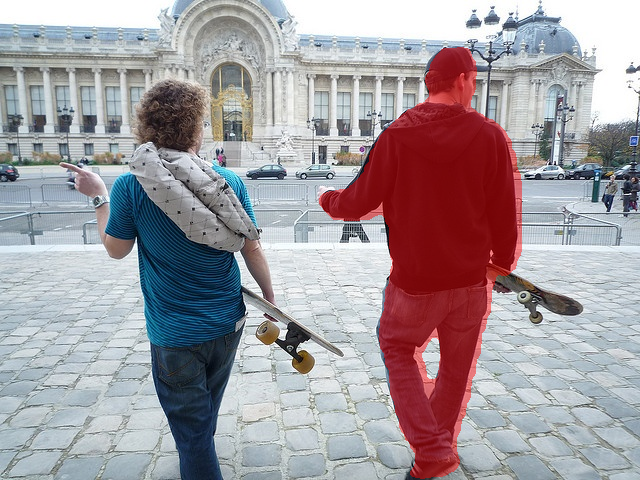} &
\includegraphics[width=0.13\textwidth,height=1.86cm]{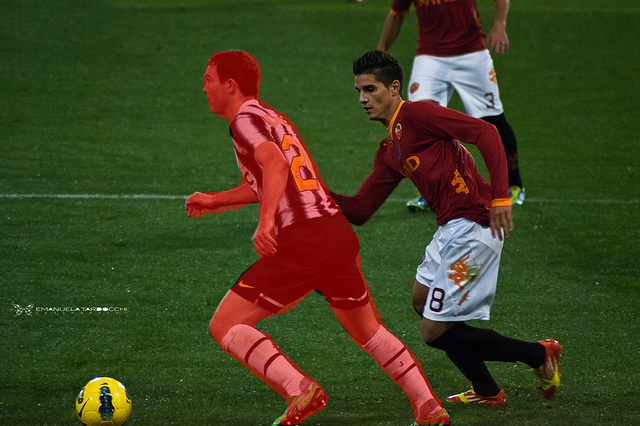} &
\includegraphics[width=0.13\textwidth,height=1.86cm]{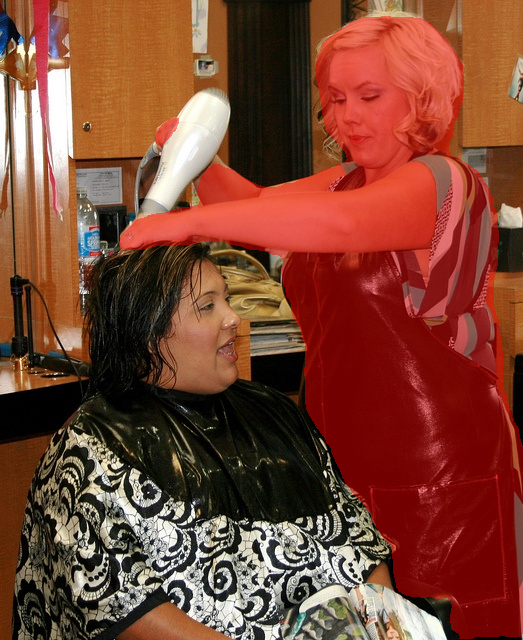} &
\includegraphics[width=0.13\textwidth,height=1.86cm]{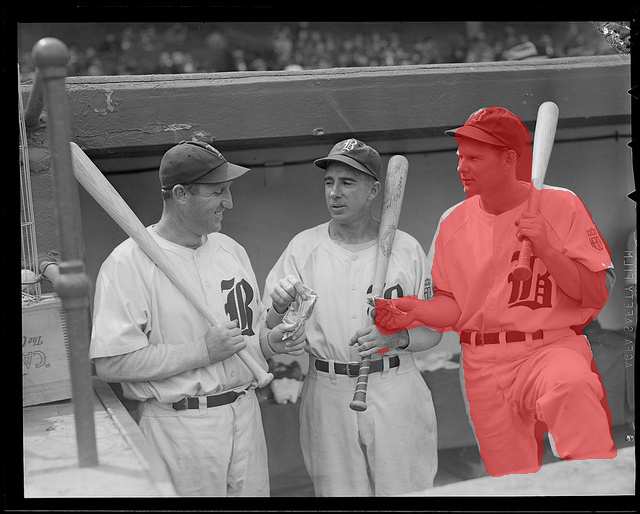} &
\includegraphics[width=0.13\textwidth,height=1.86cm]{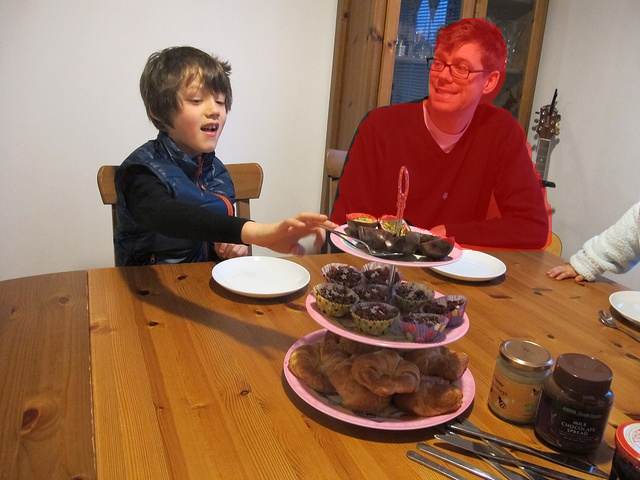} &
\includegraphics[width=0.13\textwidth,height=1.86cm]{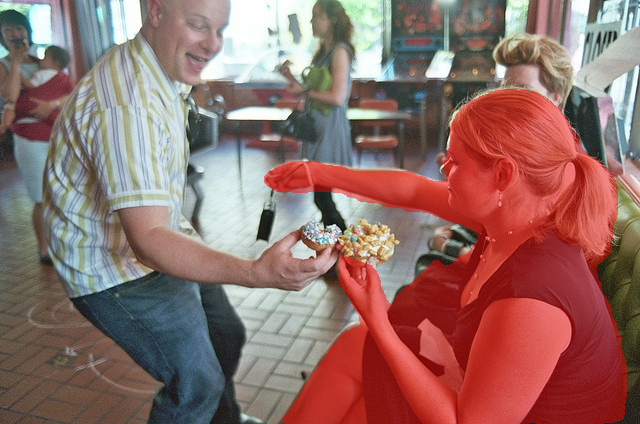} \\
 & \scriptsize 90.93
 & \scriptsize 91.19
 & \scriptsize 88.39
 & \scriptsize 91.40
 & \scriptsize 92.04
 & \scriptsize 89.93
 & \scriptsize 87.51 \\
\end{tabular}
\caption{ Zero-shot qualitative overlay results on RefCOCO and RefCOCOg. Numbers above images indicate clickable sample IDs referenced in Appendix~\ref{sec:appendix}. Values below each image denote IoU scores.}
\label{fig:qual6}
\end{figure}

A similar behavior is observed when training on the RefCOCOg dataset, as shown in Figure~\ref{fig:qual6}. The model generalizes well to both RefCOCO and RefCOCO+, successfully handling diverse object categories and long, descriptive expressions. The consistent transfer performance across datasets highlights the robustness of the learned representations to linguistic complexity and variation in annotation styles.

Overall, these qualitative examples indicate that our model learns transferable vision-language representations that generalize effectively across RefCOCO-family datasets under zero-shot transfer, without requiring target-domain adaptation.


\section{Conclusion and Future Work}
\label{sec6}
In this paper, we introduced SERA, a vision-language framework for referring image segmentation to further improve spatial visual representations with structured expert refinement. Rather than a single homogeneous refinement stage, SERA constructs a set of complementary expert modules that specialize in different visual understanding aspects (spatial layout, contextual relations, and boundary structure), allowing them to be combined in a modular fashion. We use Top-$k$ selection for routing sparsely, on which experts contribute to each prediction, making the mechanism adaptive to refine visual features and well aligned to the pretrained vision–language backbones.
We conduct experiments on the RefCOCO, RefCOCO+, and RefCOCOg benchmarks to show a consistent improvement in the segmentation performance. The improvements are most significant for harder queries, which include ambiguous descriptions, fine-grained attributes, or complex spatial relations. These improvements are achieved using lightweight refinement modules placed on top of frozen pretrained encoders, allowing the model to improve cross-modal grounding without retraining large foundation models. However, some limitations remain. The current framework relies on manually designed experts and performs refinement only in the visual stream, which may limit deeper multimodal specialization. In addition, introducing multiple expert branches increases the model’s complexity, especially when scaling to larger backbones or higher-resolution inputs. Future work will explore data-driven expert discovery, language-aware routing, and hierarchical mixtures of vision–language models to improve multimodal reasoning and generalization.

\FloatBarrier

\bibliographystyle{cas-model2-num}
\bibliography{cas-refs}

\clearpage
\onecolumn
\appendix
\section{Appendix}
\label{sec:appendix}
\begin{longtable}{
>{\centering\arraybackslash}m{1.5cm} | m{11cm}
}
\caption{ Mapping between column indices in different Figures and their corresponding referring expressions. The prompts (with grammatical and spelling mistakes) are taken as it is from the corresponding datasets.}
\label{tab:appendix_prompts} \\
\hline
ID & Referring Expression \\
\hline
\endfirsthead

\hline
ID & Referring Expression \\
\hline
\endhead

\hline
\endfoot

\hline
\endlastfoot
\hline

\hypertarget{p1}{1}  & catcher \\
\hypertarget{p2}{2}  & man on right \\
\hypertarget{p3}{3}  & dude with jacket \\
\hypertarget{p4}{4}  & far left bowl \\
\hypertarget{p5}{5}  & black blurred car on left \\
\hypertarget{p6}{6}  & horse on left \\
\hypertarget{p7}{7}  & bowl of rice \\
\hypertarget{p8}{8}  & bus in front \\
\hypertarget{p9}{9}  & white dog \\
\hypertarget{p10}{10} & girl in pink \\
\hypertarget{p11}{11} & girl without face shown \\
\hypertarget{p12}{12} & girl brushing teeth with elbow out \\
\hypertarget{p13}{13} & bowl of berries more burned toast \\
\hypertarget{p14}{14} & the black cat \\
\hypertarget{p15}{15} & man looking away \\
\hypertarget{p16}{16} & man on stump \\
\hypertarget{p17}{17} & donut standing straight up \\
\hypertarget{p18}{18} & guy with curly hair glasses \\
\hypertarget{p19}{19} & a large vase between two smaller vases \\
\hypertarget{p20}{20} & blue train on the far right rail driving ahead of two other trains \\
\hypertarget{p21}{21} & girl brushing teeth with elbow out \\
\hypertarget{p22}{22} & a female chef wearing a white chef's hat looking into the pan \\
\hypertarget{p23}{23} & the giraffe putting its head on a tree branch which is only partly visible \\
\hypertarget{p24}{24} & an older man wearing a black t-shirt and silver watch \\
\hypertarget{p25}{25} & the wicker chair to the left \\
\hypertarget{p26}{26} & small black suitcase \\
\hypertarget{p27}{27} & surfer wearing blue plaid board shorts \\
\hypertarget{p28}{28} & man walking with hat \\
\hypertarget{p29}{29} & a man with a purple shirt \\
\hypertarget{p30}{30} &  a father with his child watching giraffe drinking water\\
\hypertarget{p31}{31} & a girl playing wii \\
\hypertarget{p32}{32} &  a white headed cow eating grass\\
\hypertarget{p33}{33} &  hot dog on left end \\
\hypertarget{p34}{34} &  black keyboard directly infront of monitor\\
\hypertarget{p35}{35} &  the rest of the pizza\\
\hypertarget{p36}{36} &  guy with glasses\\
\hypertarget{p37}{37} &  man reaching\\
\hypertarget{p38}{38} &  woman eating\\
\hypertarget{p39}{39} &  the groom\\
\hypertarget{p40}{40} &  closest boy\\
\hypertarget{p41}{41} &  box with plants in it\\
\hypertarget{p42}{42} &  blue shirt bottom left\\
\hypertarget{p43}{43} &  person on the right\\
\hypertarget{p44}{44} &  man in red shirt\\
\hypertarget{p45}{45} &  black coat left\\
\hypertarget{p46}{46} &  left guy\\
\hypertarget{p47}{47} &  boy in a red shirt\\
\hypertarget{p48}{48} &  man\\
\hypertarget{p49}{49} &  a man getting ready to cut a cake\\
\hypertarget{p50}{50} &  the patterned fabric hanging behind the teddy bear\\
\hypertarget{p51}{51} & a large elephant standing next to a large wooden pole \\
\hypertarget{p52}{52} & a baseball catcher in the ground \\
\hypertarget{p53}{53} & a little girl dressed in brown playing a video game\\
\hypertarget{p54}{54} & the vase on the right  \\
\hypertarget{p55}{55} & large tv screen mountd on wall\\
\hypertarget{p56}{56} & girl on the left \\
\hypertarget{p57}{57} & little girl in middle \\
\hypertarget{p58}{58} & leather chair on left \\
\hypertarget{p59}{59} &  wine glass in front\\
\hypertarget{p60}{60} &  left\\
\hypertarget{p61}{61} &  ancient cellphone\\
\hypertarget{p62}{62} &  man standing\\
\hypertarget{p63}{63} &  blue shirted arm\\
\hypertarget{p64}{64} &  black shirt\\
\hypertarget{p65}{65} &  the number 2\\
\hypertarget{p66}{66} &  lady standing\\
\hypertarget{p67}{67} &  person holding bat and one knee up\\
\hypertarget{p68}{68} &  the man\\
\hypertarget{p69}{69} &  catwoman rachindg and sitting\\

\hline

\end{longtable}



\end{document}